\documentclass{article}

\PassOptionsToPackage{numbers, sort&compress}{natbib}

\usepackage[preprint]{neurips_2026}

\usepackage[utf8]{inputenc} 
\usepackage[T1]{fontenc}    
\usepackage{hyperref}       
\usepackage{url}            
\providecommand{\doi}[1]{\href{https://doi.org/#1}{doi:\,#1}}  
\usepackage{booktabs}       
\usepackage{amsfonts}       
\usepackage{amsmath}        
\usepackage{nicefrac}       
\usepackage{microtype}      
\usepackage{capt-of}        
\usepackage[ruled]{algorithm}  
\usepackage{algpseudocode}     
\usepackage{xcolor}         
\usepackage{graphicx}       
\usepackage{pifont}         
\usepackage{multirow}       
\usepackage{makecell}       
\usepackage{rotating}       
\usepackage{subcaption}     
\usepackage{wrapfig}        
\newcommand{\cmark}{\textcolor{teal}{\ding{51}}}
\newcommand{\xmark}{\textcolor{red!60!black}{\ding{55}}}

\newcommand{\appref}[1]{\hyperref[#1]{Appendix~\ref*{#1}}}

\title{FLOATBench: A Dataset and Benchmark for Floating
Offshore Wind Turbine Tower Fatigue}

\author{%
  Jo\~ao Alves Ribeiro\thanks{Corresponding author. Also at: LAETA-INEGI,
    Faculty of Engineering, University of Porto, Portugal;\newline Center for
    Mechanical Technology and Automation, University of Aveiro, Portugal.} \\
  Department of Mechanical Engineering \\
  Massachusetts Institute of Technology \\
  Cambridge, MA 02139 USA \\
  \texttt{jpar@mit.edu} \\
  \And
  Bruno Alves Ribeiro\thanks{Also at: Faculty of Mechanical Engineering,
    Delft University of Technology, Netherlands.} \\
  School of Engineering \\
  Brown University \\
  Providence, RI 02912 USA \\
  \texttt{bruno\_ribeiro@brown.edu} \\
  \AND
  Francisco Pimenta \\
  CONSTRUCT, Faculty of Engineering \\
  University of Porto \\
  Porto, 4200-465 Portugal \\
  \texttt{fnpimenta@fe.up.pt} \\
  \And
  S\'ergio M.\,O.\ Tavares \\
  Center for Mechanical Technology and Automation \\
  University of Aveiro \\
  Aveiro, 3810-193 Portugal \\
  \texttt{smotavares@ua.pt} \\
  \AND
  Faez Ahmed \\
  Department of Mechanical Engineering \\
  Massachusetts Institute of Technology \\
  Cambridge, MA 02139 USA \\
  \texttt{faez@mit.edu} \\
}

\begin{document}

\maketitle

\begin{abstract}
Most of the world's offshore wind resource lies in waters too deep for fixed-bottom foundations, making floating offshore wind turbines (FOWTs) essential for deep-water deployment. As the industry scales toward $22$\,MW class designs, tower fatigue becomes increasingly critical because larger structures amplify the coupled aero-hydro-servo-elastic loads induced by continuous wind and wave excitation. Accurate fatigue-damage prediction is therefore central to certification, design optimization, and cost reduction. Yet the field lacks a shared surrogate benchmark: studies report different simulations, splits, and metrics, making methods difficult to compare. We present \textbf{FLOATBench}, a public tabular benchmark with $582{,}120$ per-section fatigue-damage labels across three $22$\,MW FOWT tower geometries, derived from $19{,}404$ high-fidelity OpenFAST simulations across the three towers ($6{,}468$ per tower: $1{,}078$ aligned wind/wave operating points $\times$ six turbulence seeds), labeled at $30$ cross-sections per tower. FLOATBench includes a regime-aware alpha-shape partition of the joint wind/wave operating envelope, stratifying test points into in-train, interpolation, and extrapolation regimes. It is paired with a reproducible evaluation harness covering three protocol levels: random validation (E1), within-tower regime-aware evaluation (E2), and cross-tower transfer (E3). The regime-aware protocol reveals rank shifts between global and extrapolation performance that random-split leaderboards cannot detect. To the authors' knowledge, FLOATBench is the first FOWT fatigue benchmark for tabular surrogate modeling, and offers an evaluation protocol that generalizes to engineering surrogates defined over physical operating envelopes. Dataset and code available at: \url{https://github.com/Joao97ribeiro/FLOATBench}.
\end{abstract}


\section{Introduction}
\label{sec:intro}

Floating offshore wind is a near-term frontier for deep-water
renewable expansion: most of the world's high-quality offshore wind
resource lies in waters beyond $\sim$50\,m depth, accessible only
through floating designs~\citep{ribeiro2025owt_review}. As turbines
scale toward 22\,MW the tower becomes a design bottleneck, with
fatigue coupling turbulence-driven aerodynamic loads on the rotor,
irregular hydrodynamic loads on the floating platform, nonlinear
mooring restoring forces, and gravitational loads from the heavy
rotor-nacelle assembly atop the tower, accumulated across billions
of stress cycles over a 20--25-year service life. High-fidelity aero-hydro-servo-elastic
simulators such as OpenFAST~\citep{openfast} resolve these dynamics,
but a single 10-minute condition costs roughly one CPU-hour, and a
single design iteration crosses thousands of operating conditions
per tower. Surrogate models are now standard practice in
scientific machine learning~\citep{drivAerNetPP,carbench,pdebench}
for exactly this kind of bottleneck, but the floating offshore wind
turbine (FOWT) fatigue community has no shared benchmark on which
to compare them.

Existing wind-fatigue datasets and works cover a single tower
geometry, sample sparse or site-specific wind/wave operating points,
provide labels for at most two tower cross-sections, and evaluate
surrogates under random train/test splits. As a consequence, every
surrogate study uses its own simulation set, its own split, and its
own metrics, and the field cannot adjudicate between competing
methods. The gap is more than a missing dataset: a benchmark for
engineering surrogates must test reliability in extrapolation
beyond the sampled envelope, not just interpolation inside it.

For FOWT fatigue, the deployment-relevant coordinates are physical and
interpretable (mean wind speed, wind-speed standard deviation as a
turbulence-intensity proxy, significant wave height $H_s$, peak
wave period $T_p$), so the operating envelope admits a clean
geometric stratification of test points into In-train,
Interpolation, and Extrapolation. The random partition default, common in scientific-ML
benchmarks, fails here: it leaves the extrapolation regimes
empty, so the leaderboard cannot distinguish a surrogate that
interpolates well from one that fails in extrapolation.

\begin{figure}[!b]
  \centering
    \includegraphics[width=\linewidth]{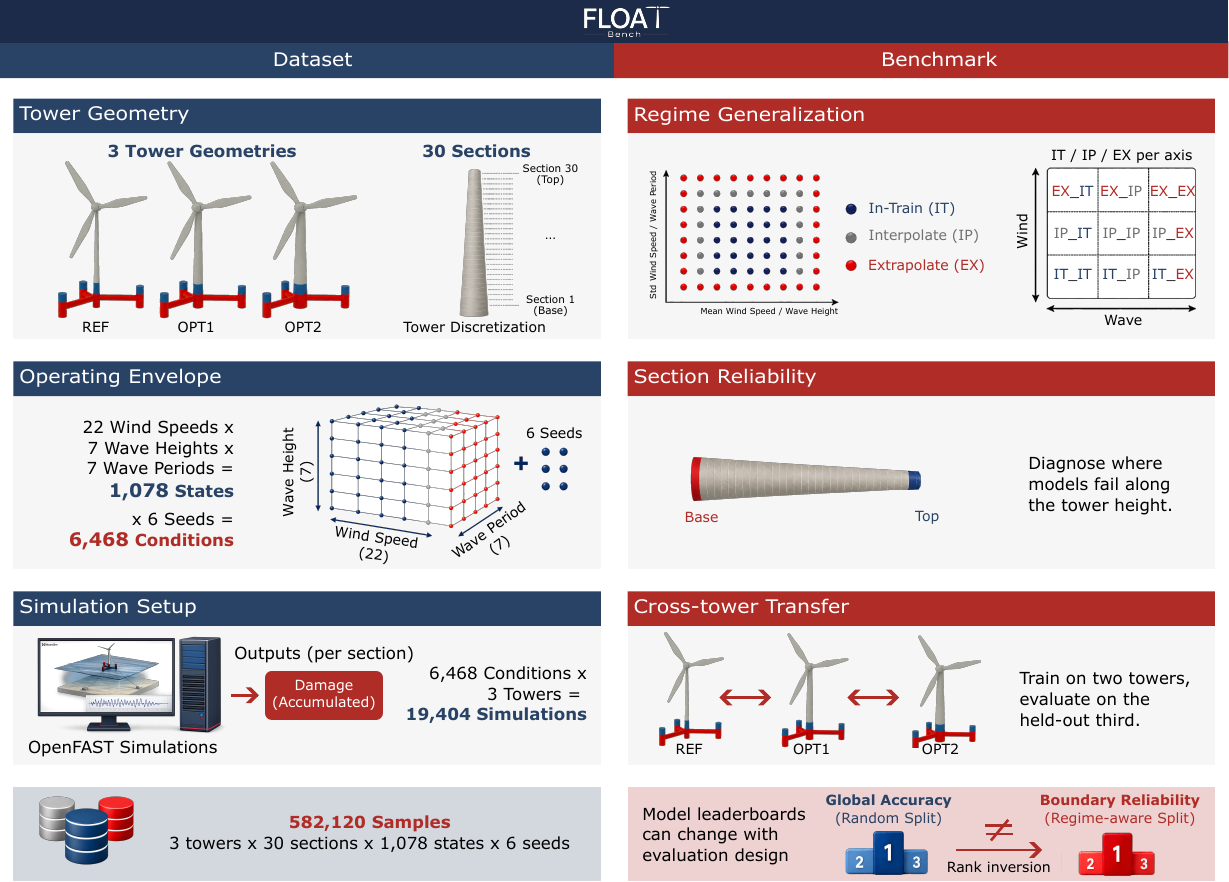}
  \caption{\textbf{FLOATBench:} a dataset and benchmark for
  22\,MW FOWT tower fatigue. Dataset: three 22-MW floating tower
    geometries, $1{,}078$ aligned wind and wave states, $6$ seeds,
    and $30$ sections yield damage labels. Benchmark: evaluation by regime, section
    reliability, and cross-tower transfer separate global accuracy
    from boundary reliability.}
  \label{fig:hero_main}
\end{figure}

\textbf{FLOATBench} closes this gap with a public, multi-geometry,
per-section fatigue benchmark paired with an evaluation
protocol matched to the operating envelope
(\autoref{fig:hero_main}). Our contributions are:
\textbf{(i) Dataset:} $582{,}120$ per-section fatigue labels from
$19{,}404$ OpenFAST simulations ($6{,}468$ per tower) across three
22\,MW FOWT towers, $1{,}078$ aligned wind/wave states, six
turbulence seeds, and 30 tower sections.
\textbf{(ii) Protocol:} a regime-aware alpha-shape partition of
the operating envelope covering all nine wind/wave regime cells,
with per-section reliability and a reproducible evaluation
harness.
\textbf{(iii) Benchmark:} a controlled comparison of tabular
surrogates that exposes rank shifts between global and
subgroup-specific performance (regimes, sections, cross-tower
transfer) invisible to random splits.


\section{Related work}
\label{sec:related_work}

\paragraph{Benchmarks for scientific ML in engineering.}
Recent scientific-ML benchmarks span PDE and multi-physics
surrogates (PDEBench, PINNacle, The Well,
LIPS)~\citep{pdebench,pinnacle,thewell,lips}, aerodynamics
(AirfRANS, DrivAerNet++,
CarBench)~\citep{airfrans,drivAerNetPP,carbench}, geophysics and
climate (OpenFWI, WeatherBench~2)~\citep{openfwi,weatherbench2},
catalyst chemistry (OC20)~\citep{oc20}, and tabular materials
(MatBench)~\citep{matbench}. These benchmarks have established the
practice of releasing simulator-surrogate datasets at scale with
reproducible splits and metrics. None, however, covers wind-turbine
fatigue, and none stratifies surrogate performance by coverage of
the physical operating envelope where the surrogate will be
deployed.

\paragraph{Surrogate evaluation by operating-envelope coverage.}
Engineering surrogates are deployed over physical operating
envelopes (wind/wave states, flight conditions, soil parameters),
and the practical question is whether predictions stay reliable as
inputs drift away from the training cloud. Alpha
shapes~\citep{edelsbrunner1983shape} estimate this cloud's boundary
from a finite sample, supporting a clean
In-train\,/\,Interpolation\,/\,Extrapolation partition over the
operating dimensions. Adopting this partition at the benchmark
level shifts the leaderboard from average accuracy to boundary
reliability, where deployment risk concentrates.

\paragraph{Datasets and surrogates for wind/offshore fatigue.}
Wind-turbine fatigue datasets are the closest domain-specific
resources, but public benchmark-ready releases remain limited
(\autoref{tab:dataset_comparison}): every prior release ships a
single geometry, $\le 2$ tower sections, and a random or
site-specific evaluation. The closest public analogue is Papi and
Bianchini's release of $\sim$447\,k hindcast-driven damage
equivalent load (DEL) records for a single NREL 5\,MW +
OC4-DeepCwind semisubmersible configuration~\citep{papi2024oc4};
no public release exists at the 20+\,MW scale where current
floating designs are converging. On the methods side, prior
surrogate studies for wind/offshore fatigue span Kriging, Gaussian
processes, polynomial chaos, neural networks, mixture-density
networks, and copula-based
models~\citep{dimitrov2018kriging,slot2020bayesian,singh2024fcnn,singh2024mdn,liu2024fowt,singh2025fowt,hu2023multisection,muller2018rbf,wei2024vinecopula,zhao2024fowt,schmidt2025kriging,avendano2021gp},
but each study uses its own simulation set, split, metrics, and
reporting protocol, so the relative merits of these families
remain unaudited and the field cannot adjudicate between them.
The missing artifact is a shared benchmark that decouples the
modeling question from the simulation question.

\begin{table}[!ht]
  \caption{Wind-turbine fatigue datasets: scope, envelope
  sampling, label coverage, and accessibility.}
  \label{tab:dataset_comparison}
  \centering
  \scriptsize
  \setlength{\tabcolsep}{3pt}
  \begin{tabular}{l c l c r c r c c}
    \toprule
    Reference & Power [MW] & Substructure & \#geom & \#cond (input space) & \#sect & \#labels & Regime & Public \\
    \midrule
    \multicolumn{9}{l}{\emph{Onshore (aero only)}} \\
    \citet{dimitrov2018kriging}                & 10           & Onshore                            & 1          & 10\,k (9-D)            & 2           & 160\,k    & \xmark & \xmark   \\
    \citet{slot2020bayesian}                   & 5            & Onshore                            & 1          & 0.6\,k (5-D)           & 1           & 0.6\,k    & \xmark & \xmark   \\
    IWT 7.5\,MW~\citep{iwt75mw}                & 7.5          & Onshore                            & 1          & 15\,k (4-D)            & 1           & 47\,k     & \xmark & \cmark   \\
    \citet{singh2024fcnn}                      & 5            & Onshore                            & 1          & 33\,k (3-D)            & 2           & 65\,k     & \xmark & \cmark   \\
    \midrule
    \multicolumn{9}{l}{\emph{Bottom-fixed offshore (aero + hydro on a static substructure)}} \\
    \citet{mueller2018database}                & 5            & Tripod                             & 1          & 0.7\,k (5-D)           & 2           & 9\,k      & \xmark & \xmark   \\
    \citet{singh2024mdn}                       & 10           & Onshore + monopile                  & 2          & 9\,k (5-D)             & 2           & 107\,k    & \xmark & \cmark   \\
    \midrule
    \multicolumn{9}{l}{\emph{Floating offshore (aero + hydro + mooring restoring and rigid-body floater motions)}} \\
    FLOATECH D2.3~\citep{lifes50_d23}          & 10/15        & Semisub.\ + Spar + Hexa             & 3          & 0.5\,k (5-D)           & 1           & 1.5\,k    & \xmark & \cmark   \\
    \citet{liu2024fowt}                        & 5            & Semisub.\ (OC4 DeepCw.)            & 1          & 32 (3-D)                     & 1           & 32              & \xmark & \xmark   \\
    \citet{papi2024oc4}                        & 5            & Semisub.\ (OC4 DeepCw.)            & 1          & 447\,k (hindcast)            & 2           & 894\,k    & \xmark & \cmark   \\
    \citet{singh2025fowt}                      & 6            & Spar (Hywind Scotland)             & 1          & 9\,k (7-D)             & 2           & 22\,k     & \xmark & \xmark   \\
    \addlinespace[1pt]
    \textbf{FLOATBench (ours)}                 & \textbf{22}  & \textbf{Semisub.\ (IEA-22)}        & \textbf{3} & \textbf{1{,}078 (3-D)}       & \textbf{30} & \textbf{582\,k} & \cmark & \cmark   \\
    \bottomrule
    \addlinespace[2pt]
    \multicolumn{9}{p{0.97\linewidth}}{\scriptsize\emph{\#geom}: distinct turbine/substructure geometries; \emph{\#cond}: distinct operating conditions; \emph{\#sect}: tower cross-sections with fatigue labels; \emph{\#labels}: per-section tower fatigue labels.}
  \end{tabular}
\end{table}

FLOATBench provides, to our knowledge, the first public
scientific-ML benchmark at this scale: three 22\,MW FOWT
geometries (vs.\ a single 5--15\,MW geometry in every prior
release), 30 tower sections (vs.\ 1--2), $582{,}120$ per-section
fatigue labels, and a regime-aware evaluation protocol derived
from physical-envelope coverage rather than a random split.


\section{Dataset}
\label{sec:dataset}
\label{sec:openfast_compute}
\begin{wrapfigure}{r}{0.30\linewidth}
  \vspace{-3.5em}
  \centering
  \includegraphics[width=\linewidth]{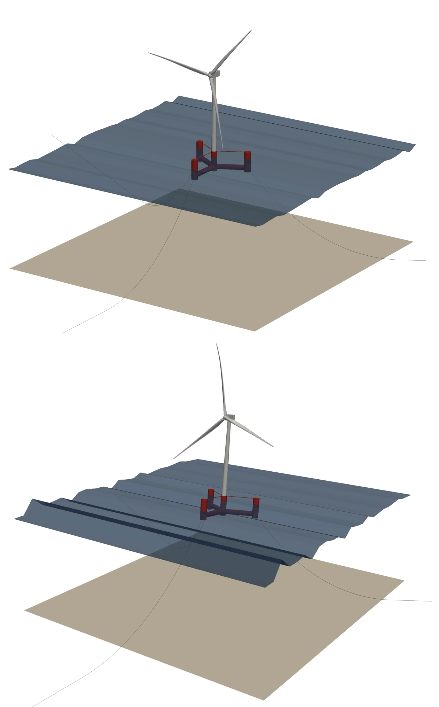}
  \caption{Example OpenFAST simulations at the operating envelope
  extremes: near cut-in wind ($V\!\approx\!4.5$\,m/s, top) and
  near cut-out wind ($V\!\approx\!24.5$\,m/s, bottom).}
  \label{fig:operating_envelope}
  \vspace{-3.5em}
\end{wrapfigure}

\paragraph{Operating envelope.}
Each tower is simulated across a $22\times7\times7$ grid of
environmental conditions (\autoref{fig:operating_envelope}): 22
mean hub-height wind speeds at the midpoints of 1\,m/s bins from
cut-in ($3$\,m/s) to cut-out ($25$\,m/s), and, for each wind speed,
seven $H_s$ levels with seven $T_p$ levels per $H_s$, sampled from
the joint wind--wave operating distribution defined in our prior
work~\citep{ribeiro2026float}, yielding 49 sea states per wind
speed and $1{,}078$ unique (wind, wave) operating points per tower.
Wind and waves are aligned.

\begin{figure}[!ht]
  \centering
  \includegraphics[width=\linewidth]{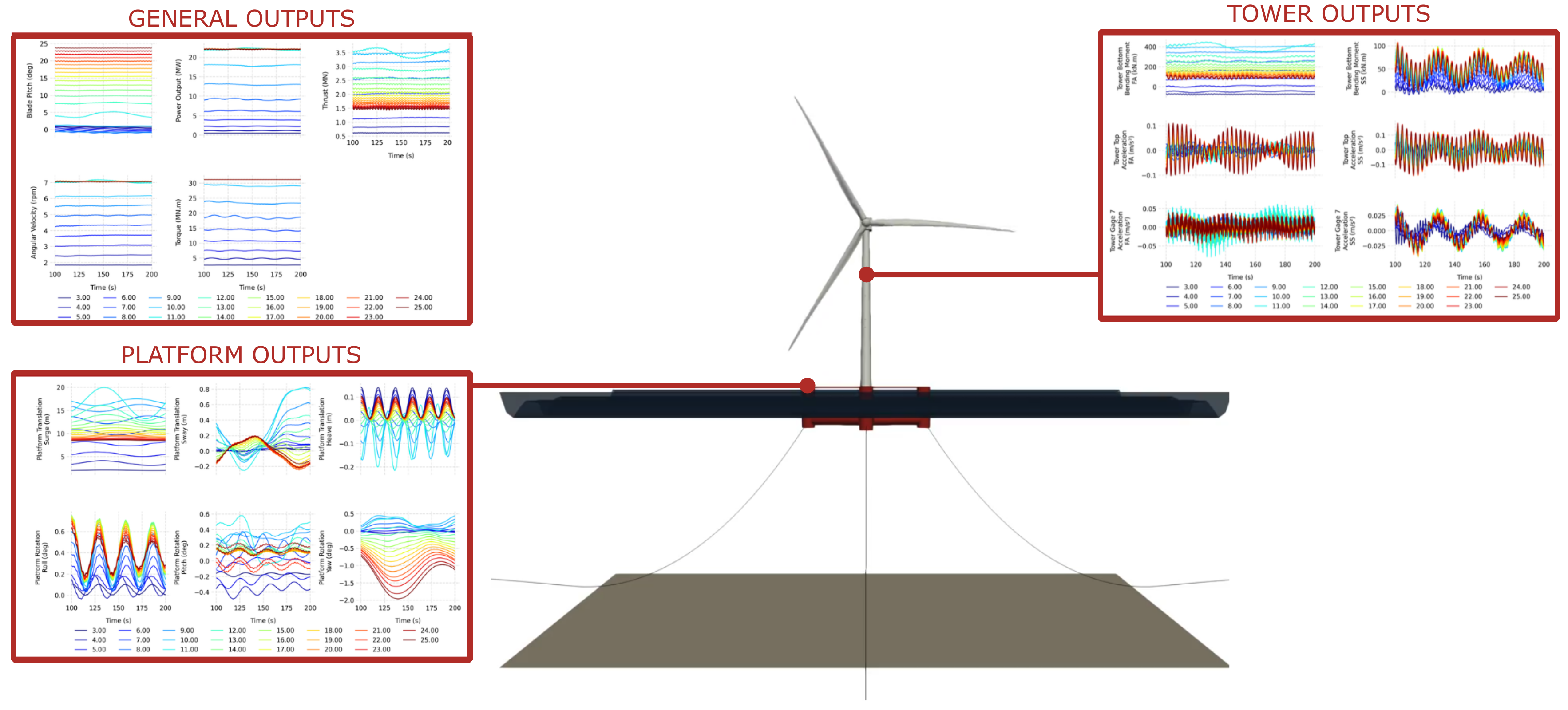}
  \caption{\textbf{FLOATBench OpenFAST outputs and damage
  pipeline.} Each of our OpenFAST simulations yields 88
  time-series at 10\,Hz around the IEA-22-280-RWT on a
  three-column semi-submersible. \emph{Pipeline:} our OpenFAST
  simulations $\to$ 88 time-series $\to$ bending moments $\to$
  stress $\to$ rainflow $\to$ S-N + Miner $\to$ damage labels.}
  \label{fig:simulation_outputs}
\end{figure}

\paragraph{Simulation setup.}
\label{sec:simulation_setup}
All simulations follow IEC~61400-3-2~\citep{iec61400_3_2} design
load case (DLC) 1.2 (normal power production), run with
OpenFAST~\citep{openfast} across the operating envelope. Each
operating point is simulated with 6 independent turbulence (wind)
seeds, giving $6{,}468$ ten-minute simulations per tower
($1{,}078\times6$); runs last $1{,}000$\,s with the final 600\,s
post-processed. Each run records $88$ time-series at 10\,Hz
organized into three groups
(\autoref{fig:simulation_outputs}): general turbine outputs
(e.g., rotor speed, blade pitch, generator power, rotor thrust), tower outputs (e.g., section bending
moments, accelerations, and deflections at multiple heights), and platform outputs (e.g., 6-DOF translation and rotation motions,
mooring fairlead and anchor tensions, wave elevation), totaling
$\approx\!190$\,GB
across the $19{,}404$ runs for the 3 towers. Setup details in our
previous work~\citep{ribeiro2026float}.

\paragraph{Compute.}
Simulations were executed on a commercial cloud HPC platform on
2-vCPU virtual-machine instances at $\approx\!30$\,min wall-clock
per run ($\approx\!1$ CPU-hour). Each tower required
$6{,}468$ simulations ($\approx\!6{,}500$ CPU-hours,
$\approx\!0.74$ core-years), and the full 3-tower release
consumed $\approx\!19{,}400$ CPU-hours in total
($\approx\!2.2$ core-years).

\paragraph{Damage outputs.}
\label{sec:damage_labelling}
Fatigue is quantified by the cumulative damage $D$. For each
tower, each
operating condition, and each of the 30 tower sections, OpenFAST
produces a bending moment time-series
over the post-processed window, which is converted to stress
and processed through rainflow
counting~\citep{matsuishi1968fatigue,downing1982rainflow} and
Miner summation against an S--N curve~\citep{dnv_rp_c203},
yielding the cumulative damage $D$, the released label.
Full pipeline details in our previous
work~\citep{ribeiro2026float}; pipeline overview in
\autoref{fig:simulation_outputs}. The
load-equivalent transform $\mathrm{DEL} \propto D^{1/m}$,
$m = 3$, computed from $D$, is also used as a regression target.

\begin{figure}[!ht]
  \centering
  \begin{subfigure}[b]{0.32\linewidth}
    \centering
    \includegraphics[width=\linewidth]{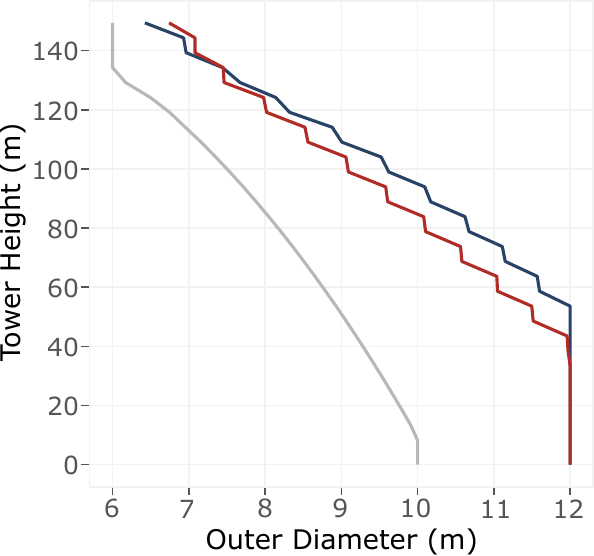}
  \end{subfigure}\hfill
  \begin{subfigure}[b]{0.32\linewidth}
    \centering
    \includegraphics[width=\linewidth]{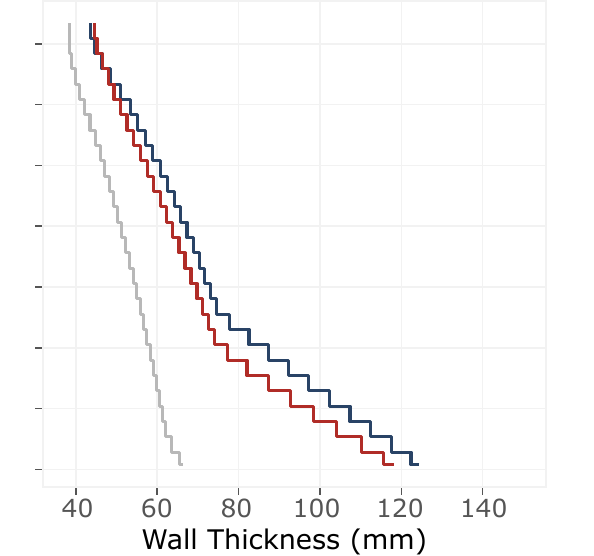}
  \end{subfigure}\hfill
  \begin{subfigure}[b]{0.32\linewidth}
    \centering
    \includegraphics[width=\linewidth]{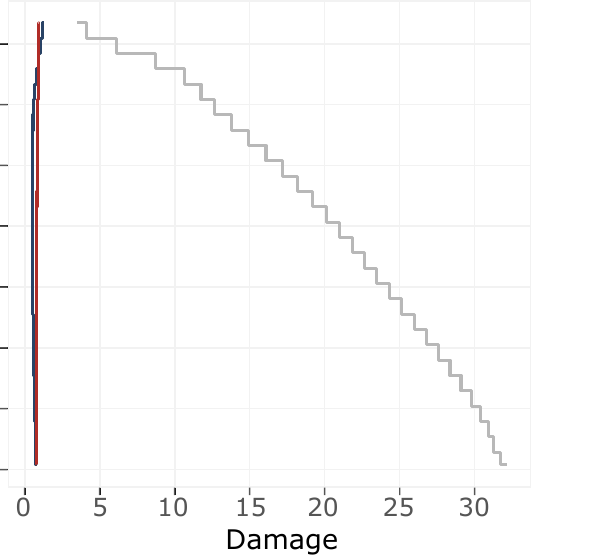}
  \end{subfigure}\\[2pt]
  \includegraphics[width=\linewidth]{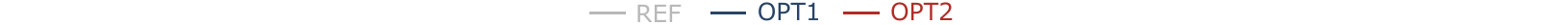}\\[4pt]
  \begin{subfigure}[b]{0.32\linewidth}
    \centering
    \caption{Outer diameter profile}
    \label{fig:tower_diameter}
  \end{subfigure}\hfill
  \begin{subfigure}[b]{0.32\linewidth}
    \centering
    \caption{Wall thickness profile}
    \label{fig:tower_thickness}
  \end{subfigure}\hfill
  \begin{subfigure}[b]{0.32\linewidth}
    \centering
    \caption{Damage profile}
    \label{fig:tower_damage}
  \end{subfigure}
  \caption{Geometry and FLOATBench lifetime damage profiles along
  the tower height for the three released towers \textsc{ref},
  \textsc{opt1}, and \textsc{opt2}.}
  \label{fig:tower_geom_damage}
\end{figure}

\paragraph{Tower geometries.}
Three towers are released, all on the same three-column
semi-submersible platform, controller, and mooring:
\textsc{ref}, the IEA-22-280-RWT~\citep{iea22mw} baseline tower
(designed for fixed-bottom conditions without explicit fatigue
constraints, $D\!\approx\!32$ at the tower base, $\approx\!9$ months under Miner's rule on a
$25$-yr horizon);
\textsc{opt1}, an intermediate fatigue-aware re-design from our
FLOAT method~\citep{ribeiro2026float} ($D\!\approx\!1.0$);
and \textsc{opt2}, a final iterate ($D\!\approx\!0.9$) targeting
$D\!\le\!0.90$.
The redesign sequence \textsc{ref}\,$\to$\,\textsc{opt1}\,$\to$\,\textsc{opt2}
varies the outer diameter and wall thickness profiles at
constant tower height ($148.385$\,m)
(\autoref{fig:tower_geom_damage}). All three towers are
discretized into $30$ sections with damage labels at midpoints.
Geometric profiles, damage profiles, and natural frequencies are
in \appref{app:tower_geom_damage}.

\paragraph{Dataset schema and access.}
\label{sec:dataset_schema}
Each tower contributes $194{,}040$ rows ($6{,}468 \times 30$
sections), totaling $582{,}120$ rows across the three towers; we
release three CSVs per tower: the training set, the test set with
regime labels, and a raw table holding all rows without split or
regime labels (the canonical source for reproducing the partition).
Each row carries identifiers (simulation, section, and grid
coordinates on the $22\times7\times7$ wind/wave envelope plus the
turbulence-seed index), environmental features (nominal, mean, and
std hub-height wind speed, $H_s$, $T_p$), tower section geometry
(height, radius, thickness), per-row regime labels on test rows
(wind and wave), the Miner-summed damage label, and a per-condition
lifetime weighting factor for aggregating damage over the 25-year
service life; full schema in \appref{app:schema_ref}.


\paragraph{Hosting and licensing.}
The dataset is hosted on Hugging Face under CC-BY-4.0; the
release includes a Croissant~1.1 manifest and a Datasheet for
Datasets~\citep{gebru2021datasheets}. The
benchmark code is released under MIT license. See
Appendices~\ref{app:reproducibility}--\ref{app:license} for details.



\section{Benchmark protocol}
\label{sec:benchmark_protocol}

\subsection{Regime-aware partition}
\label{sec:regime_split}

\paragraph{Train/test split.}
From the $22\times7\times7$ grid, test holds the extreme and
middle indices on both the wind and wave axes, leaving 18 of 22
wind speeds and a $4\times4$ inner wave patch
on the train side (all 6 seeds per condition kept in the
same fold to prevent train-test leakage; full index list in
\appref{app:train_indices}). This yields 51{,}840 train /
142{,}200 test rows per tower ($\approx 27/73$ train/test split). Unlike a
row-matched i.i.d.\ 80/20 baseline, the partition induces a
structured covariate shift along both wind and wave axes; the
test set is larger than the training set and lies outside its
support, keeping every extrapolation region densely sampled.

\begin{figure}[!ht]
  \centering
  \begin{subfigure}[b]{0.48\linewidth}
    \centering
    \includegraphics[width=\linewidth]{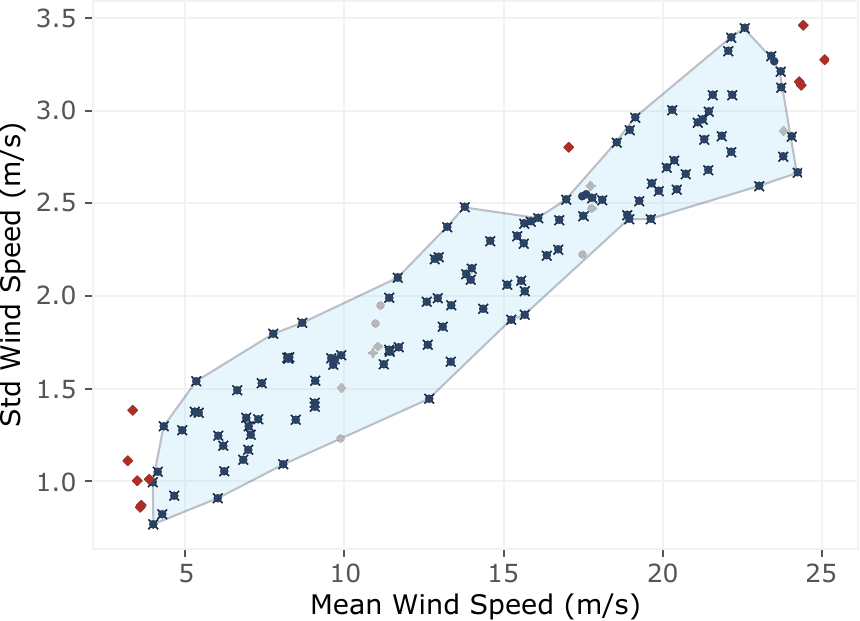}
  \end{subfigure}\hfill
  \begin{subfigure}[b]{0.48\linewidth}
    \centering
    \includegraphics[width=\linewidth]{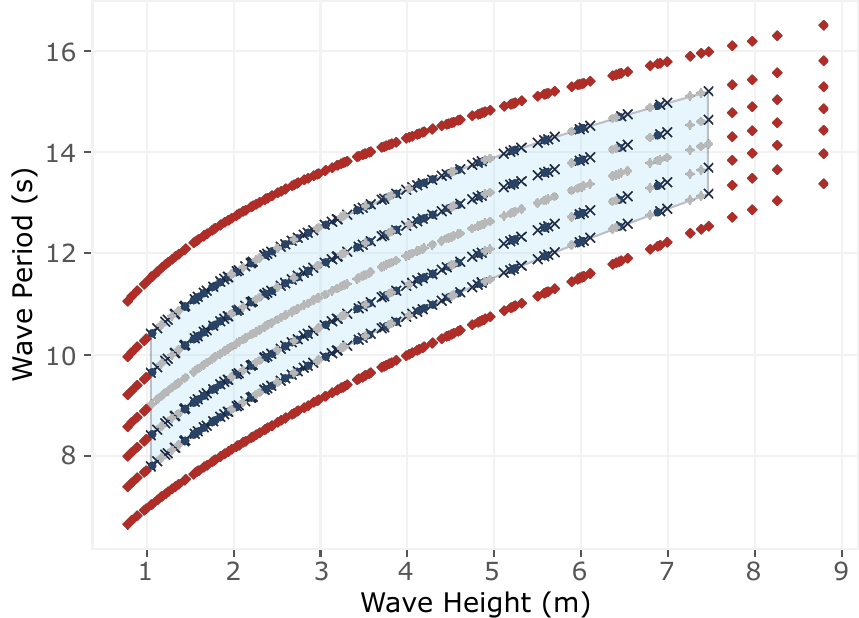}
  \end{subfigure}\\[2pt]
  \includegraphics[width=\linewidth]{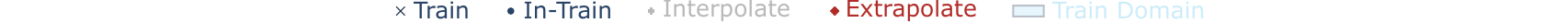}\\[2pt]
  \begin{subfigure}[b]{0.48\linewidth}
    \centering
    \caption{Wind subspace}
    \label{fig:regime_split_wind}
  \end{subfigure}\hfill
  \begin{subfigure}[b]{0.48\linewidth}
    \centering
    \caption{Wave subspace}
    \label{fig:regime_split_wave}
  \end{subfigure}
  \caption{FLOATBench regime partition; the training domain
  (alpha-shape hull) is shaded.}
  \label{fig:regime_split}
\end{figure}

\paragraph{Regime labeling.}
Each test point is labeled In-train, Interpolation, or
Extrapolation (\textsc{IT}, \textsc{IP}, \textsc{EX}) independently
in the wind $(\mathrm{mean}\,V, \mathrm{std}\,V)$ and wave $(H_s,
T_p)$ subspaces (\autoref{fig:regime_split}):
\begin{enumerate}
  \item \textbf{Train spacing (train$\to$train).} For each
  training point, the nearest-neighbor distance to the rest of
  the training set is computed in standardized feature units;
  the mean over all training points sets the spacing scale.
  \item \textbf{Distance-based grouping (test$\to$train).} For
  each test point, the nearest-neighbor distance to the training
  set is computed in standardized feature units and normalized by
  the spacing scale; points with normalized distance $\le 0.5$
  are labeled \textsc{IT}, the rest provisionally \textsc{IP}.
  \item \textbf{Boundary-based override.} A concave hull
  ($\alpha$-shape, $\alpha=0.1$) bounds the training support
  in the wind and wave subspaces; test points outside the hull
  \emph{and} farther than a small numerical tolerance from its
  boundary are re-labeled \textsc{EX}. Points just outside the
  hull but within the tolerance stay as \textsc{IP}, so the
  tolerance acts as a slack band against boundary-proximate
  misclassification.
\end{enumerate}
Pseudocode and method details are in \appref{app:alg_regime_label}. Crossing the per-axis labels
(each \textsc{IT}\,/\,\textsc{IP}\,/\,\textsc{EX}) gives nine joint
regimes, with \textsc{EX\_EX} (extrapolation on both axes) the
worst case. Random validation samples test points from the training
envelope and leaves every extrapolation regime empty by
construction; the regime-aware partition populates all nine
(random vs regime-aware composition in
\appref{app:partition_regime_comparison}).

\subsection{Evaluation levels}
\label{sec:eval_levels}

\paragraph{E1: baseline (random split).}
A random partition matched in size to E2
($\approx\!27/73$ train/test split), run independently on each tower
(\textsc{ref}, \textsc{opt1}, \textsc{opt2}). The same surrogate
set and training-set size are used in both E1 and E2, so any
performance difference between them reflects the partitioning
strategy alone.

\paragraph{E2: within-tower (by regime and section).}
Models are evaluated under the regime-aware partition, run
independently on each tower (\textsc{ref}, \textsc{opt1},
\textsc{opt2}); results are reported per regime (with
\textsc{EX\_EX} as worst-case extrapolation) and per section (Section~1
base vs Section~30 top).

\paragraph{E3: cross-tower transfer.}
A leave-one-tower-out scheme trains on two towers and evaluates on
the held-out third
(\textsc{ref}+\textsc{opt1}$\to$\textsc{opt2},
\textsc{ref}+\textsc{opt2}$\to$\textsc{opt1},
\textsc{opt1}+\textsc{opt2}$\to$\textsc{ref}), measuring whether
the regime hierarchy survives a change of geometry while holding
the operating envelope fixed.

\subsection{Tabular surrogates}
\label{sec:baselines}

We evaluate the same surrogate models across all three levels
(E1, E2, E3): XGBoost~\citep{chen2016xgboost},
LightGBM~\citep{ke2017lightgbm},
CatBoost~\citep{prokhorenkova2018catboost},
NeuralNetFastAI~\citep{howard2020fastai},
NeuralNetTorch~\citep{erickson2020autogluon},
ExtraTrees~\citep{geurts2006extratrees},
RandomForest~\citep{breiman2001randomforest}, and
TabM~\citep{gorishniy2024tabm}. To cover a broad set of surrogates at each (tower-configuration,
experiment level) combination, all baselines are trained through
AutoGluon-Tabular
1.5.0~\citep{erickson2020autogluon}, which provides pre-tuned
hyperparameter portfolios and automatic stacking. We use the
\texttt{best} (\texttt{hyperparameters=zeroshot}) and
\texttt{extreme} (\texttt{hyperparameters=zeroshot\_2025\_tabfm})
presets, taking as inputs four environmental conditions (mean
wind speed, std wind speed, $H_s$, $T_p$) and three
section-geometry descriptors (section height, section radius,
section thickness; full schema in \appref{app:schema_ref}), and the cube-root damage transform
$D^{1/3}$ (i.e.\ the DEL-format target) as the regression target. Each (tower-configuration, experiment level, preset) combination
runs under a 4-hour wall-clock budget, yielding
82--96 models per tower for E1/E2 and 58--63 per fold for E3
(E3 trains on two towers instead of one, so each model takes
longer to fit and fewer portfolio variants finish within the same
budget).
Extending FLOATBench to in-context tabular foundation models
is left to future work; more on baselines and training in
\appref{app:baseline_models} and \appref{app:training_config}.

\subsection{Metrics and statistical significance}
\label{sec:metrics}

We assess surrogate predictions of damage (in DEL format) using two error
metrics: \textbf{Rel L\textsuperscript{2}}, the relative
$L^2$-norm error, captures global prediction quality on a
dimensionless scale and is used as the primary metric for ranking
models; \textbf{MRE} (point-wise mean relative error) captures the
average per-point degradation and is used as the per-regime
metric. Lower is better
in both.
Auxiliary error metrics (MSE, MAE, RMSE, R\textsuperscript{2},
MaxErr) and compute metrics (inference latency, throughput,
training time) are also reported in the leaderboard tables. Every
leaderboard number is reported as $\bar\theta_{\mathrm{boot}} \pm
\sigma_{\mathrm{boot}}$ (bootstrap mean and standard deviation, $B=2{,}000$
resamples); the corresponding $95\%$ percentile-bootstrap intervals
are computed by the released harness and emitted into the leaderboard
CSVs (a top-3 selection, Rel L\textsuperscript{2}, is reproduced in
\appref{app:top3_ci_rell2}).
Formulas, rationale, and the bootstrap algorithm are in
\appref{app:metrics_definitions} and
\appref{app:bootstrap_ci_method}.


\section{Results}
\label{sec:results}

We organize the FLOATBench evaluation along the three protocol levels declared in \autoref{sec:eval_levels}: the random baseline (E1, \autoref{sec:f1}), the within-tower regime-aware partition (E2, \autoref{sec:f2}), and cross-tower transfer (E3, \autoref{sec:f4}). Full top-10 leaderboards for E1, E2, and E3 are in \appref{app:leaderboards}, reported with 95\,\% percentile-bootstrap confidence intervals.


\subsection{Random split (E1): everything looks easy, the wrong models win}
\label{sec:f1}

Under E1's random split, all top-10 surrogates (\appref{app:top10_e1}) appear strong: roughly $2$--$4\%$ Rel L\textsuperscript{2} DEL across towers, with bootstrap CIs overlapping between adjacent ranks. But the test set contains no hard regime: random validation leaves every extrapolation regime empty (see \appref{app:partition_regime_comparison} for regime composition), so the regimes that matter are invisible by construction. The leaderboard rewards fit inside the training domain, and the rank-1 surrogate under E1 is \emph{not} the rank-1 once boundary regimes appear.

\noindent\emph{A damage benchmark needs a regime-aware split, not a random one, to choose the right model.}

\subsection{Within-tower regime-aware (E2): different regimes pick different winners}
\label{sec:f2}

\begin{figure}[!htbp]
  \centering
  \includegraphics[width=0.65\linewidth]{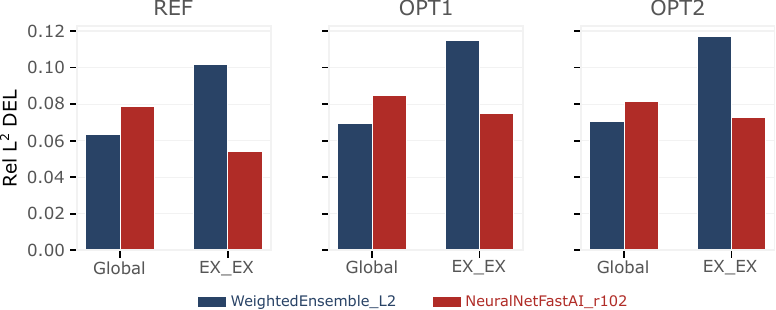}
  \caption{\textbf{Within-tower cross-over (E2):} on each tower, the Global winner \texttt{WeightedEnsemble\_L2} (blue) drops at the wind-and-wave extrapolation regime \textsc{EX\_EX}, where \texttt{NeuralNetFastAI\_r102\_BAG\_L1} (red) wins despite being ranked low globally.}
  \label{fig:crossover_hero}
\end{figure}

\paragraph{The Global winner loses at the wind-and-wave extrapolation regime.}
On every tower, the worst-case regime is \textsc{EX\_EX} (extrapolation on both wind and wave), which dominates the error (see \appref{app:ex_ex_dominance} showing \textsc{EX\_EX} dominance over the other regimes). Ranking on all test points (Global) versus on \textsc{EX\_EX} points only gives different winners on every tower: the Global rank-1 \texttt{WeightedEnsemble\_L2} drops to \textsc{EX\_EX} ranks 23\,/\,11\,/\,11 (\textsc{ref}\,/\,\textsc{opt1}\,/\,\textsc{opt2}), while the \textsc{EX\_EX} rank-1 \texttt{NeuralNetFastAI\_r102\_BAG\_L1} sits at Global ranks 79\,/\,73\,/\,69 (\autoref{fig:crossover_hero}).

\paragraph{NeuralNet variants are strongest at extrapolation; TabM weakest.}
The \textsc{EX\_EX} top-10 (\appref{app:f1_exex_top10}) is dominated by \texttt{NeuralNet} BAG\_L1 variants (FastAI and Torch backends) across all three towers. The per-model scatter of MRE DEL on Global vs.\ \textsc{EX\_EX} (\appref{app:scatter_global_exex}) confirms the family pattern: \texttt{NeuralNet} variants stay closest to the diagonal (similar errors on Global and \textsc{EX\_EX}), while \texttt{TabM} falls farthest below (heaviest degradation at \textsc{EX\_EX}).

\paragraph{Wind drives the extrapolation; wave axis stays flat.}
\label{sec:f3}
\label{sec:f3_sections}
Aggregating errors by model family (see \appref{app:bar_family_regime} for family-aggregated MRE DEL by wind and wave regime), the wind axis inflates by $6$--$20\times$ from In-train to Extrapolation, while the wave axis stays flat ($1.1$--$1.4\times$). \texttt{NeuralNet} is the best family on wind Extrapolation and \texttt{TabM} the worst.

\paragraph{Sections pick different winners than Global.}
Ranking on each section's test points alone gives different winners than the Global ranking. On \textsc{ref}, no model wins both sections: \texttt{NeuralNetFastAI\_r11\_BAG\_L2} wins Section~1 (Global rank 5), \texttt{RandomForestMSE\_BAG\_L1} wins Section~30 (Global rank 60). On \textsc{opt1}/\textsc{opt2}, \texttt{WeightedEnsemble\_L2} (the Global rank-1) keeps rank-1 at the base (Section~1) but loses Section~30 to \texttt{NeuralNetTorch\_r22\_BAG\_L1} (Global ranks 71/44). The base carries more error than the top on every tower: $+32\%$ on \textsc{ref} and $\approx\!+45\%$ on \textsc{opt1}/\textsc{opt2}. The gap is larger on the re-designed towers, reflecting the high-stress regime at the tower base. Full per-section top-10 in \appref{app:top10_section1}.

\noindent\emph{For damage prediction, the Global rank-1 does not remain best across regimes and sections: it loses at \textsc{EX\_EX} and at the tower top, while remaining strongest at the base for \textsc{opt1}/\textsc{opt2}.}

\subsection{Cross-tower transfer (E3): the rank-1 fails again, on geometry and on sections}
\label{sec:f4}

\begin{figure}[!htbp]
  \centering
  \includegraphics[width=0.5\linewidth]{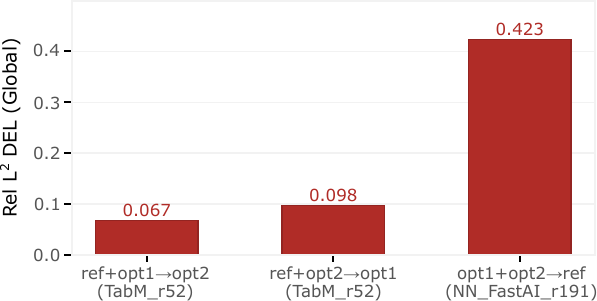}
  \caption{\textbf{Cross-tower transfer (E3):} rank-1 Rel L\textsuperscript{2} DEL per fold (model name in parentheses). Reaches $0.067$/$0.098$ when training includes \textsc{ref}, and $0.423$ when training without \textsc{ref}.}
  \label{fig:cross_tower_bars}
\end{figure}

\paragraph{Transfer collapses on geometries far from training.}
E3 trains on two towers and evaluates on the held-out third (E3 top-10 per fold in \appref{app:cross_tower_pool}). The cross-tower error depends sharply on which towers are in training (\autoref{fig:cross_tower_bars}): training on a set that includes \textsc{ref} generalizes to the re-designed geometries with rank-1 Rel L\textsuperscript{2} DEL of $0.067$/$0.098$ (\textsc{ref+opt1}\,$\to$\,\textsc{opt2} and \textsc{ref+opt2}\,$\to$\,\textsc{opt1}), while training without \textsc{ref} (\textsc{opt1+opt2}\,$\to$\,\textsc{ref}) reaches $0.423$, a $4$--$6\times$ increase. The fold that holds \textsc{ref} out under-predicts, consistent with \textsc{ref}'s wider damage profile and most-distinct geometry (\autoref{fig:tower_geom_damage}); see \appref{app:scatter_e3} for the cross-tower asymmetry in the per-fold predicted-vs-true scatter.

\paragraph{Sections pick different winners than Global, across folds.}
On every fold, no model wins all three rankings: the Section~1 (base) winner is a \texttt{NeuralNet} variant on all three folds, while \texttt{TabM\_r52\_BAG\_L1} wins Section~30 (top) on all three. The Global winner is \texttt{TabM\_r52\_BAG\_L1} on the two folds whose training set includes \textsc{ref}, and \texttt{NeuralNetFastAI\_r191\_BAG\_L1} on \textsc{opt1+opt2}\,$\to$\,\textsc{ref}. Full per-section top-10 in \appref{app:cross_tower_sections}.

\noindent\emph{For damage prediction across towers, the Global rank-1 again does not remain best across geometry and sections.}


\section{Limitations and future work}
\label{sec:limitations}

\paragraph{Single design family.}
The release covers the IEA-22 FLOAT baseline and two fatigue-aware re-designs derived from it. The cross-tower setup is well-suited to surrogate evaluation within a single design family (the realistic setting for fatigue-aware re-design iteration), but it does not test cross-design transfer to NREL\,5\,MW / DTU\,10\,MW / IEA-15; extending the benchmark to broader designs is deferred to follow-up work.

\paragraph{Tabular foundation models.}
The current pool uses AutoGluon's \texttt{best} and \texttt{extreme} presets; future work would extend it with tabular foundation models (TabPFN-v2, TabICL, Mitra), whose in-context learning offers a different inductive bias for generalising to unseen operating conditions with limited samples.

\paragraph{Time-series benchmark.}
The full $\approx\!190$\,GB OpenFAST time-series record (88 outputs at 10\,Hz across the $19{,}404$ runs) was generated to compute the tabular labels released here but is not released in v1; lifting FLOATBench to time-series targets (tower bending moments used to compute the stress and damage) is future work, alongside extending the benchmark beyond the tower to cover other FOWT components (platforms and blades).

\paragraph{High-fidelity 3D extensions.}
A planned next step is to release the structural and aerodynamic meshes for the three towers, then re-run selected operating conditions with high-fidelity FEA and CFD. This would enable neural surrogate benchmarks on high-fidelity simulations for flow fields and 3D stress, which can in turn be used to compute damage, building on prior work that applies deep learning to structural stress prediction and ML-assisted mechanical design~\citep{ribeiro2023simustruct,ribeiro2021stressstrain,mlp_ribeiro,designs8020029}. It would also provide aligned multi-fidelity data spanning meshes and time series, supporting multi-fidelity surrogates that combine the OpenFAST tabular benchmark with sparse high-fidelity FEA/CFD simulations.


\section{Conclusion}
\label{sec:conclusion}

\textbf{FLOATBench} addresses the lack of a shared FOWT fatigue benchmark with $582{,}120$ per-section damage labels across three $22$\,MW floating-tower geometries, a regime-aware alpha-shape partition exposing the IT/IP/EX regime structure of the joint wind/wave envelope, and three protocol levels (random, within-tower regime-aware, cross-tower transfer). Across up to $96$ tabular surrogates per tower (E1/E2) and up to $63$ per fold (E3), totalling $735$ trained surrogates over the three protocols, the regime-aware protocol reveals rank shifts between global and extrapolation performance (the global rank-1 surrogate is not the rank-1 at the worst-case wind-and-wave extrapolation cell), which random-split leaderboards systematically miss, and a related rank inversion appears under cross-tower transfer. Beyond the empirical finding, the release of the dataset, evaluation harness, and trained surrogates establishes common ground for adjudicating competing tabular surrogates on this domain. Looking forward, the time-series record underlying the released labels and the planned multi-fidelity FEA/CFD extensions position FLOATBench as a foundation for surrogate research beyond the tabular task. To the authors' knowledge, FLOATBench is the first FOWT fatigue benchmark for surrogate development, and its regime-aware protocol generalizes to any tabular surrogate over a low-dimensional physical operating envelope.

\clearpage
\begin{ack}
\textbf{João Alves Ribeiro} acknowledges funding from the Luso-American Development Foundation (FLAD) and the doctoral grant SFRH/BD/151362/2021 (DOI: \href{https://doi.org/10.54499/SFRH/BD/151364/2021}{10.54499/SFRH/BD/151364/2021} (accessed on November 18, 2024)), financed by the Portuguese Foundation for Science and Technology (FCT), Ministério da Ciência, Tecnologia e Ensino Superior (MCTES), Portugal, with funds from the State Budget (OE), European Social Fund (ESF), and PorNorte under the MIT Portugal Program, and by the Alliance for the Energy Transition (56) co-financed by the Recovery and Resilience Plan (PRR) through the European Union.

\textbf{Bruno Alves Ribeiro} acknowledges financial support from FCT through the doctoral grant 2021/08659/BD.

\textbf{Francisco Pimenta} acknowledges the financial support for project 2022.08120.PTDC, M4WIND (DOI: \href{https://doi.org/10.54499/2022.08120.PTDC}{10.54499/2022.08120.PTDC} (accessed on November 18, 2024)), funded by national funds through FCT/MCTES (PIDDAC), and for UID/ECI/04708/2020-CONSTRUCT-Instituto de I\&D em Estruturas e Construções, also funded by national funds through FCT/MCTES (PIDDAC).
\end{ack}

\appendix


\clearpage
\section*{Table of Contents for Appendices}

\begingroup
\renewcommand{\arraystretch}{1.15}
\begin{tabular}{@{}rl@{}}
\textbf{A} & \textbf{Reproducibility statement} \dotfill \pageref{app:reproducibility} \\
\textbf{B} & \textbf{URL and links} \dotfill \pageref{app:urls} \\
\textbf{C} & \textbf{Author statement and data license confirmation} \dotfill \pageref{app:license} \\
\textbf{D} & \textbf{Broader impact} \dotfill \pageref{app:broader_impact} \\
\textbf{E} & \textbf{Dataset details} \dotfill \pageref{app:dataset_details} \\
           & \quad E.1 Tower geometry, damage, and natural frequencies (\autoref{fig:tower_damage_compare}, \autoref{tab:geometric_comparison}, \autoref{tab:damage_comparison}, \autoref{tab:nat_freqs}) \\
           & \quad E.2 Schema reference (\autoref{tab:schema}) \\
\textbf{F} & \textbf{Benchmark details} \dotfill \pageref{app:benchmark} \\
           & \quad F.1 Train/test selection (\autoref{tab:train_test_indices}) \\
           & \quad F.2 Regime-labeling algorithm and distance distributions (\autoref{alg:regime_label}, \autoref{fig:train_spacing}, \autoref{tab:train_spacing_summary}) \\
           & \quad F.3 Random vs regime-aware split: regime composition (\autoref{tab:partition_regime_comparison}, \autoref{fig:app_random_subspaces}) \\
           & \quad F.4 Baseline models (\autoref{tab:model_pool}) \\
           & \quad F.5 Training configuration \\
           & \quad F.6 Metric definitions \\
           & \quad F.7 Bootstrap confidence intervals (\autoref{alg:bootstrap_ci}) \\
\textbf{G} & \textbf{Leaderboards (E1, E2 Global, E3 cross-tower)} \dotfill \pageref{app:leaderboards} \\
           & \quad G.1 Top-10 under random (E1), per tower (\autoref{tab:app_top10_e1}) \\
           & \quad G.2 Top-10 under regime-aware (E2), per tower (\autoref{tab:app_top10_e2}) \\
           & \quad G.3 Top-10 under cross-tower (E3), per fold (\autoref{tab:app_top10_e3}) \\
           & \quad G.4 Bootstrap CIs for top-3 surrogates (Rel L\textsuperscript{2}, E2 and E3) (\autoref{tab:app_top3_ci_rell2}) \\
\textbf{H} & \textbf{E2: how the best surrogate varies by regime, family, and section} \dotfill \pageref{app:e2_evidence} \\
           & \quad H.1 Wind and wave Extrapolation dominates over other regimes (\autoref{fig:app_heatmap_3towers}) \\
           & \quad H.2 Top-10 surrogates on the \textsc{EX\_EX} regime under E2, per tower (\autoref{tab:app_top10_exex}) \\
           & \quad H.3 \texttt{NeuralNet} family stays closest to the Global-vs-\textsc{EX\_EX} diagonal (\autoref{fig:app_scatter_global_exex_3towers}) \\
           & \quad H.4 Wind Extrapolation dominates over wave at the family level (\autoref{fig:app_bar_family_regime_3towers}) \\
           & \quad H.5 Top-10 surrogates on Section~1 (base) and Section~30 (top) under E2, per tower (\autoref{tab:app_top10_section1}, \autoref{tab:app_top10_section30}) \\
\textbf{I} & \textbf{E3: how cross-tower transfer behaves across folds and sections} \dotfill \pageref{app:cross_tower_appendix} \\
           & \quad I.1 Cross-tower transfer is asymmetric across folds (\autoref{fig:app_scatter_e3}) \\
           & \quad I.2 Top-10 surrogates on Section~1 (base) and Section~30 (top) under E3, per fold (\autoref{tab:app_e3_section1_top10}, \autoref{tab:app_e3_section30_top10}) 
         \\
\end{tabular}
\endgroup

\clearpage


\clearpage
\section{Reproducibility statement}
\label{app:reproducibility}

We provide a GitHub repository to facilitate the reproduction of all
experiments in the main paper, along with links to download the
dataset. The repository includes code to replicate the benchmark
experiments and code to generate the figures presented in the paper.
The full dataset is publicly available at the links provided in
\autoref{app:urls}. The training and experiments were performed
on a machine equipped with an Intel Core i9-14900K processor
(24 cores, 32 threads), 128\,GB RAM, and an NVIDIA GeForce RTX 4090
GPU (24\,GB VRAM).


\section{URL and links}
\label{app:urls}

\paragraph{FLOATBench dataset.}
The FLOATBench dataset is hosted on Hugging Face under CC-BY-4.0 at
\url{https://huggingface.co/datasets/DeCoDELab/FLOATBench}.

\paragraph{Tower geometry artifacts.}
The baseline IEA-22-280-RWT reference turbine is available at
\url{https://github.com/IEAWindSystems/IEA-22-280-RWT}.
The FLOAT-derived re-designs \textsc{opt1} and \textsc{opt2} are
released at
\url{https://github.com/Joao97ribeiro/FLOAT-22-280-RWT-Semi}.

\paragraph{Croissant metadata.}
A Croissant 1.1 metadata record documenting the dataset is
available at
\url{https://huggingface.co/api/datasets/DeCoDELab/FLOATBench/croissant}.

\paragraph{Benchmark code.}
The source code for model training, evaluation, and figure generation
is available at
\url{https://github.com/Joao97ribeiro/FLOATBench}.


\section{Author statement and data license confirmation}
\label{app:license}

We, the authors, hereby declare that we bear full responsibility for
any violations of rights or other issues arising from the use of the
data. We confirm that the FLOATBench dataset is released under the
Creative Commons Attribution 4.0 International license (CC-BY-4.0),
which allows others to share and adapt the work for any purpose,
including commercial use, provided that appropriate credit is given.
The benchmark code is released under the MIT license.

\paragraph{Third-party assets and their licenses.} The existing
assets used in this work are released under permissive open-source
licenses, all compatible with the redistribution terms of FLOATBench:
IEA-22-280-RWT~\citep{iea22mw} (Apache-2.0); OpenFAST~\citep{openfast}
(Apache-2.0); AutoGluon-Tabular 1.5.0 (Apache-2.0); XGBoost
(Apache-2.0); LightGBM (MIT); CatBoost (Apache-2.0); scikit-learn
(BSD-3-Clause), which provides the ExtraTrees and RandomForest
baselines; the FastAI- and PyTorch-based neural-network learners
included in AutoGluon's portfolio (Apache-2.0); and TabM (MIT).
All licenses have been reviewed and respected.


\clearpage
\section{Broader impact}
\label{app:broader_impact}

Tower fatigue is the binding design constraint for the 22\,MW
floating turbines that will harvest the world's deep-water wind
resource, and surrogate models are the only way to make the design
loop tractable. FLOATBench is, to the authors' knowledge, the first principled,
multi-geometry, per-section fatigue benchmark for this problem,
and its impact spans several research communities:

\begin{enumerate}
\item \textbf{Advancing floating wind design and accelerating design
cycles:} Enabling detailed studies in tower fatigue to foster
innovation in larger-class FOWT design, assisting structural
engineers in creating more accurate surrogate models, and supporting
the deployment of next-generation floating wind farms.

\item \textbf{Accelerating fatigue assessment:} FLOATBench provides
damage labels computed from high-fidelity OpenFAST simulations,
supporting the training of surrogate models that reduce the
computational cost and time required for fatigue analysis and enable
faster iteration over the design space.

\item \textbf{Machine learning integration:} Offering a rich resource
for training and testing tabular surrogates on a physically structured
regression task with realistic boundary regimes.

\item \textbf{Benchmarking and validation:} Serving as a benchmark
dataset with a regime-aware evaluation protocol, aiding the
validation of boundary reliability for surrogate models and exposing
global-versus-boundary failures invisible under random splits; this
matters operationally because real-world wind and wave conditions
routinely fall at the boundary of any sampled training envelope,
exactly where these failures become consequential.

\item \textbf{Environmental impact:} Contributing to the development
of more cost-effective floating wind farms, supporting the deployment
of deep-water renewable energy and reductions in levelized cost of
energy.

\item \textbf{Safety and certification support:} Accelerating early
design exploration with fast surrogate predictions while
certification-grade DLC analysis remains required for final
verification. The released labels are single-slope S--N values under
DLC~1.2 only and are not a substitute for certification-grade
analysis.

\item \textbf{Cross-disciplinary insights:} Offering insights at the
intersection of structural fatigue, machine learning, and floating
wind design, encouraging cross-disciplinary research and
collaboration.

\item \textbf{Foundation for follow-up ML research:} The
$\sim$190\,GB OpenFAST time-series record (88 outputs at 10\,Hz
across the $19{,}404$ runs) used to compute the released tabular
labels, together with the planned multi-fidelity FEA/CFD extensions
(\autoref{sec:limitations}), positions FLOATBench as a foundation
for follow-up ML research on time-series surrogates, neural
operators, and multi-fidelity learning beyond the tabular task
released here.
\end{enumerate}

Beyond floating wind, FLOATBench provides a template for any
engineering surrogate task defined over a physical operating envelope:
a regime-aware partition that exposes failures random splits cannot
see, per-section resolution that connects model accuracy to
component-level reliability, and multi-geometry evaluation that tests
transfer across design families. We expect this template to inform
surrogate development in adjacent engineering domains where realistic
deployment shifts make boundary reliability the operationally
binding metric.

\clearpage
\section{Dataset details}
\label{app:dataset_details}

\subsection{Tower geometry, damage, and natural frequencies}
\label{app:tower_geom_damage}

\paragraph{Tower geometry.}
\autoref{tab:geometric_comparison} reports the diameter and wall
thickness at the bottom and top of each released tower. Both
optimized towers thicken toward the base and taper toward
the top; \textsc{opt2} distributes material slightly more evenly
than \textsc{opt1} (thinner base, thicker top). Full geometry
artifacts (CAD profiles and OpenFAST input decks) for
\textsc{opt1}/\textsc{opt2} are released at
\url{https://github.com/Joao97ribeiro/FLOAT-22-280-RWT-Semi};
upstream \textsc{ref} artifacts at
\url{https://github.com/IEAWindSystems/IEA-22-280-RWT}.

\begin{table}[!ht]
\centering
\scriptsize
\caption{FLOATBench tower diameter and wall thickness at base
and top for the three released towers; percentage change
relative to \textsc{ref} in parentheses.}
\label{tab:geometric_comparison}
\begin{tabular}{lccc}
\toprule
Parameter              & \textsc{ref} & \textsc{opt1}            & \textsc{opt2}            \\
\midrule
Diameter, bottom (m)   & $10.000$     & $12.000$ ($+20.0\,\%$)   & $12.000$ ($+20.0\,\%$)   \\
Diameter, top (m)      &  $6.000$     &  $6.424$ ($+7.1\,\%$)    &  $6.741$ ($+12.4\,\%$)   \\
Thickness, bottom (mm) &  $66$        &  $124$ ($+87.6\,\%$)     &  $118$ ($+78.2\,\%$)     \\
Thickness, top (mm)    &  $38$        &  $43$ ($+13.1\,\%$)      &  $45$ ($+15.8\,\%$)      \\
\midrule
Total mass (t)         &  $1{,}574$   &  $2{,}899$ ($+84.2\,\%$) &  $2{,}656$ ($+68.8\,\%$) \\
\bottomrule
\end{tabular}
\end{table}

\begin{figure}[!ht]
  \centering
  \begin{subfigure}[b]{0.48\linewidth}
    \centering
    \includegraphics[width=\linewidth]{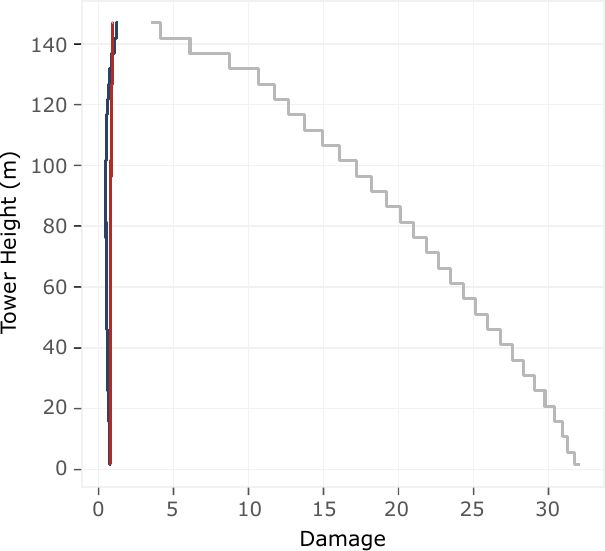}
  \end{subfigure}\hfill
  \begin{subfigure}[b]{0.48\linewidth}
    \centering
    \includegraphics[width=\linewidth]{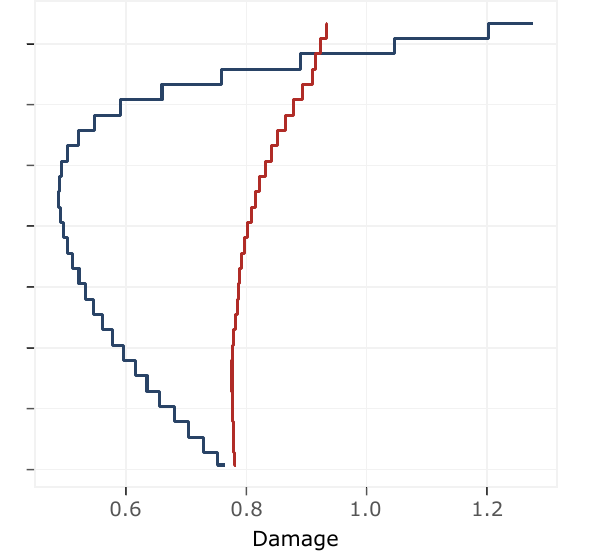}
  \end{subfigure}\\[2pt]
  \includegraphics[width=\linewidth]{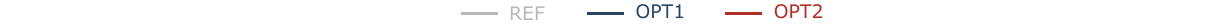}\\[2pt]
  \begin{subfigure}[b]{0.48\linewidth}
    \centering
    \caption{All three towers}
    \label{fig:tower_damage_compare_all}
  \end{subfigure}\hfill
  \begin{subfigure}[b]{0.48\linewidth}
    \centering
    \caption{Zoom on the two re-designs}
    \label{fig:tower_damage_compare_zoom}
  \end{subfigure}
  \caption{FLOATBench lifetime weighted section damage along the
  tower.}
  \label{fig:tower_damage_compare}
\end{figure}

\paragraph{Damage profile.}
\autoref{fig:tower_damage_compare} and \autoref{tab:damage_comparison} report the per-section
lifetime weighted damage profile and the boundary values.
\textsc{ref} varies by an order of magnitude along the height
($\approx\!4$ to $\approx\!32$); \textsc{opt1} varies modestly ($\approx\!0.5$--$1.25$); \textsc{opt2} is essentially constant at
$\approx\!0.9$. This progression reflects the
FLOAT~\citep{ribeiro2026float} re-design objective of reducing
and stabilising fatigue damage along the tower below
$D\le 0.90$.

\begin{table}[!ht]
\centering
\scriptsize
\caption{FLOATBench lifetime weighted section damage at base
and top; percentage change relative to \textsc{ref} in
parentheses.}
\label{tab:damage_comparison}
\begin{tabular}{lccc}
\toprule
              & \textsc{ref} & \textsc{opt1}            & \textsc{opt2}            \\
\midrule
Damage, bottom & $32.128$    & $0.764$ ($-97.6\,\%$)   & $0.781$ ($-97.6\,\%$)   \\
Damage, top    & $3.471$     & $1.277$ ($-63.2\,\%$)   & $0.932$ ($-73.1\,\%$)   \\
\bottomrule
\end{tabular}
\end{table}

\paragraph{Natural frequencies.}
\autoref{tab:nat_freqs} lists the first fore-aft (FA1) natural
frequency per tower. \textsc{ref} FA1 lies close to the 3P
excitation frequency ($\approx\!0.35$\,Hz), indicating a potential
resonance risk under specific operating conditions; the
FLOAT~\citep{ribeiro2026float} re-design shifts FA1 above this
frequency, reducing from \textsc{opt1} to \textsc{opt2}.

\begin{table}[!ht]
  \caption{FLOATBench first fore-aft natural frequency per
  tower (Hz); percentage change relative to \textsc{ref} in
  parentheses.}
  \label{tab:nat_freqs}
  \centering
  \small
  \begin{tabular}{lccc}
    \toprule
              & \textsc{ref} & \textsc{opt1}            & \textsc{opt2}            \\
    \midrule
    FA1 (Hz)  & $0.336$      & $0.573$ ($+70.5\,\%$)    & $0.537$ ($+59.8\,\%$)    \\
    \bottomrule
  \end{tabular}
\end{table}

\subsection{Schema reference}
\label{app:schema_ref}
\autoref{tab:schema} documents the schema of the released CSVs.
For each tower (\textsc{ref}, \textsc{opt1}, \textsc{opt2}) we
release three files: \texttt{train\_damage.csv} (51,840 rows, 18
columns), \texttt{test\_damage.csv} (142,200 rows, 18 columns), and
\texttt{data.csv} (194,040 rows, 16 columns: the raw rows without
split or regime labels, kept as the canonical source for
reproducing the partition). All three files carry three
grid-coordinate columns (\texttt{wind\_speed\_id},
\texttt{wave\_hs\_id}, \texttt{wave\_tp\_id}) that index each row's
position on the $22\times7\times7$ envelope; the deterministic
train/test split is fully recovered by filtering on these IDs
(\autoref{app:train_indices}). The regime-label columns
(\texttt{wind\_group}, \texttt{wave\_group}) appear only in
\texttt{train\_damage.csv} and \texttt{test\_damage.csv} and are
populated by the regime-aware splitter (\autoref{sec:regime_split}):
on train rows they are trivially \texttt{In-train}, and on test
rows they hold the actual \textsc{IT}/\textsc{IP}/\textsc{EX}
regime; the joint label \texttt{wind\_wave\_group} can be derived
at load time as \texttt{wind\_group + "\_" + wave\_group}.

\begin{table}[!ht]
  \caption{FLOATBench released CSV schema; one folder is released
  per tower (\textsc{ref}, \textsc{opt1}, \textsc{opt2}), so no
  tower identifier column is needed. Columns are grouped by
  category. Grid-coordinate IDs identify each row's position on
  the $22\times7\times7$ envelope and appear in all three files.
  Regime-label columns are \texttt{In-train} on train rows and
  carry the \textsc{IT}/\textsc{IP}/\textsc{EX} regime on test rows;
  they appear only in
  \texttt{train\_damage.csv}/\texttt{test\_damage.csv}.}
  \label{tab:schema}
  \centering
  \small
  \renewcommand{\arraystretch}{1.15}
  \begin{tabular}{@{}l l p{0.62\linewidth}@{}}
    \toprule
    Column & Type & Description \\
    \midrule
    \multicolumn{3}{@{}l}{\emph{Identifiers}} \\
    \quad\texttt{sim\_id}              & int   & Unique simulation identifier (ties together the 30 sections of one run) \\
    \quad\texttt{section\_id}          & int   & Tower section index $\in\{1,\dots,30\}$, 1 (base) to 30 (top) \\
    \quad\texttt{wind\_speed\_id}      & int   & Grid index $\in\{1,\dots,22\}$, ordered by \texttt{wind\_speed} ascending \\
    \quad\texttt{wave\_hs\_id}         & int   & Grid index $\in\{1,\dots,7\}$ within each \texttt{wind\_speed} \\
    \quad\texttt{wave\_tp\_id}         & int   & Grid index $\in\{1,\dots,7\}$ within each (\texttt{wind\_speed}, \texttt{wave\_hs}) \\
    \quad\texttt{wind\_seed\_id}       & int   & Turbulence seed index $\in\{1,\dots,6\}$ \\
    \cmidrule(l){1-3}
    \multicolumn{3}{@{}l}{\emph{Environmental features}} \\
    \quad\texttt{wind\_speed}          & float & Nominal hub-height wind speed (m/s) \\
    \quad\texttt{mean\_wind\_speed}    & float & Realised 10-min mean hub-height wind speed (m/s) \\
    \quad\texttt{std\_wind\_speed}     & float & Realised 10-min std of hub-height wind speed (m/s) \\
    \quad\texttt{wave\_hs}             & float & Significant wave height (m) \\
    \quad\texttt{wave\_tp}             & float & Wave peak period (s) \\
    \cmidrule(l){1-3}
    \multicolumn{3}{@{}l}{\emph{Tower section geometry}} \\
    \quad\texttt{section\_height\_m}   & float & Tower section midpoint height along tower axis (m) \\
    \quad\texttt{section\_radius\_m}   & float & Tower section outer radius (m) \\
    \quad\texttt{section\_thickness\_m}& float & Tower section wall thickness (m) \\
    \cmidrule(l){1-3}
    \multicolumn{3}{@{}l}{\emph{Regime labels} (only in \texttt{train\_damage.csv}/\texttt{test\_damage.csv})} \\
    \quad\texttt{wind\_group}          & str   & \texttt{In-train} on train rows; \textsc{IT}/\textsc{IP}/\textsc{EX} on test rows (wind axis) \\
    \quad\texttt{wave\_group}          & str   & \texttt{In-train} on train rows; \textsc{IT}/\textsc{IP}/\textsc{EX} on test rows (wave axis) \\
    \cmidrule(l){1-3}
    \multicolumn{3}{@{}l}{\emph{Damage targets}} \\
    \quad\texttt{damage}               & float & Miner-summed fatigue damage at the section (dimensionless) \\
    \quad\texttt{damage\_weight}       & float & Probability of occurrence over the 25-year service life; lifetime damage is $\sum_i \texttt{damage}_i \cdot \texttt{damage\_weight}_i$ over all conditions \\
    \bottomrule
  \end{tabular}
\end{table}

\clearpage
\section{Benchmark details}
\label{app:benchmark}

\subsection{Train/test selection (regime-label reference)}
\label{app:train_indices}
The train/test partition is defined by a deterministic preset
on the $22\times7\times7$ envelope;
\autoref{tab:train_test_indices} lists the train and test IDs
per variable.

\begin{table}[!ht]
\centering
\scriptsize
\caption{FLOATBench train/test partition: train and test IDs
per variable.}
\label{tab:train_test_indices}
\begin{tabular}{lllcc}
\toprule
Variable & Train IDs & Test IDs & Train count & Test count \\
\midrule
Wind speed, $V$ (22 levels)            & $\{2$--$7,\, 9$--$14,\, 16$--$21\}$ & $\{1, 8, 15, 22\}$ & 18 & 4 \\
Wave height, $H_s$ (7 levels)           & $\{2, 3, 5, 6\}$                       & $\{1, 4, 7\}$      & 4 & 3 \\
Wave peak period, $T_p$ (7 levels)      & $\{2, 3, 5, 6\}$                       & $\{1, 4, 7\}$      & 4 & 3 \\
\bottomrule
\end{tabular}
\end{table}

\subsection{Regime-labeling algorithm and distance distributions}
\label{app:alg_regime_label}

\paragraph{Algorithm.}
\autoref{alg:regime_label} formalizes the three-stage regime
labeling described in \autoref{sec:regime_split}, applied
independently on the wind $(\mathrm{mean}\,V, \mathrm{std}\,V)$ and
wave $(H_s, T_p)$ subspaces.

\begin{algorithm}[!ht]
\caption{\textsc{RegimeLabel}: distance + alpha-shape regime
labeling.}
\label{alg:regime_label}
\small
\begin{algorithmic}[1]
\Require Training points $X_\mathrm{tr}$, test points
$X_\mathrm{te}$ (per subspace), threshold $\tau=0.5$, hull
parameter $\alpha=0.1$
\Ensure Labels
$L\in\{\textsc{IT},\textsc{IP},\textsc{EX}\}^{|X_\mathrm{te}|}$
(In-train, Interpolation, Extrapolation)
\State \textbf{Stage 1 (train$\to$train spacing)}
\For{each $x\in X_\mathrm{tr}^\mathrm{unique}$}
  \State $d_x \gets \min_{y\in X_\mathrm{tr},\, y\neq x}\|x-y\|$ \Comment{standardized units}
\EndFor
\State $s \gets \mathrm{mean}(d_x)$ \Comment{spacing scale; see \autoref{fig:train_spacing}, \autoref{tab:train_spacing_summary}}
\State \textbf{Stage 2 (test$\to$train, distance-based grouping)}
\For{each $x\in X_\mathrm{te}$}
  \State $\hat d_x \gets \min_{y\in X_\mathrm{tr}}\|x-y\|/s$
  \If{$\hat d_x \le \tau$}
    \State $L_x \gets \textsc{IT}$
  \Else
    \State $L_x \gets \textsc{IP}$
  \EndIf
\EndFor
\State \textbf{Stage 3 (boundary-based override)}
\State $H \gets \mathrm{alpha\text{-}shape}(X_\mathrm{tr};\,\alpha)$
\State $\varepsilon \gets $ small numerical tolerance \Comment{slack against boundary-proximate misclassification}
\For{each $x\in X_\mathrm{te}$}
  \If{$x\notin H$ \textbf{and} $\mathrm{dist}(x,\partial H)\ge \varepsilon$}
    \State $L_x \gets \textsc{EX}$
  \EndIf
\EndFor
\State \Return $L$
\end{algorithmic}
\end{algorithm}

\begin{figure}[!ht]
  \centering
  \begin{subfigure}[b]{0.48\linewidth}
    \centering
    \includegraphics[width=\linewidth]{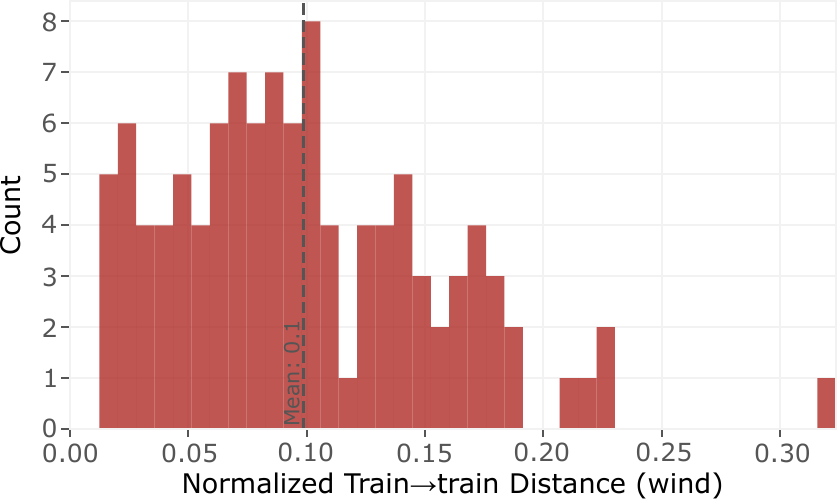}
    \caption{Wind subspace}
    \label{fig:train_spacing_wind}
  \end{subfigure}\hfill
  \begin{subfigure}[b]{0.48\linewidth}
    \centering
    \includegraphics[width=\linewidth]{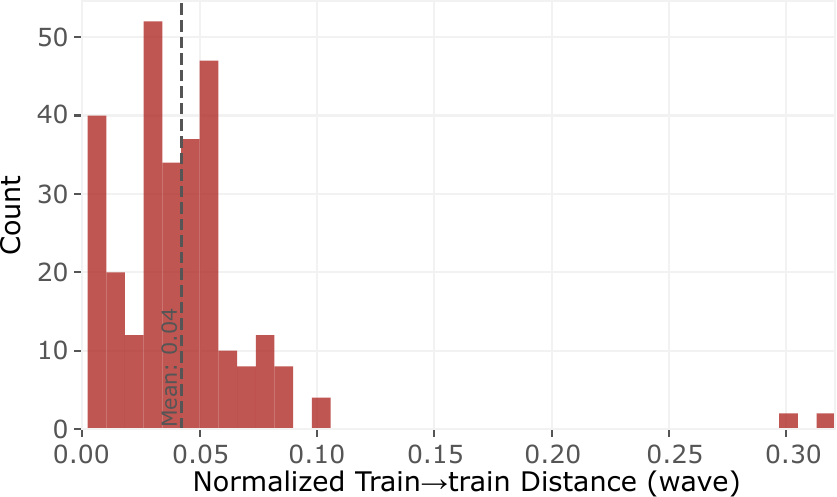}
    \caption{Wave subspace}
    \label{fig:train_spacing_wave}
  \end{subfigure}
  \caption{Train-train nearest-neighbor spacing on the standardized
  wind and wave subspaces. The dashed line marks the mean (spacing
  scale).}
  \label{fig:train_spacing}
\end{figure}

\paragraph{Distance histograms and threshold derivation.}
\label{app:distance_distrib}
\autoref{fig:train_spacing} and \autoref{tab:train_spacing_summary}
summarize the train-spacing statistics; $\tau=0.5$ places the
IT/IP boundary at half the mean train-train spacing per subspace.

\begin{table}[!ht]
\centering
\scriptsize
\caption{Train-spacing summary on the standardized wind and wave
subspaces. The \emph{mean} column is the spacing scale used to
normalize test-to-train distances.}
\label{tab:train_spacing_summary}
\begin{tabular}{lcccccc}
\toprule
subspace & $n$ unique train pts & mean & median & p25 & p90 \\
\midrule
wind & 108 & 0.099 & 0.089 & 0.056 & 0.171 \\
wave & 288 & 0.042 & 0.036 & 0.026 & 0.073 \\
\bottomrule
\end{tabular}
\end{table}

\newpage
\subsection{Random vs regime-aware split: regime composition}
\label{app:partition_regime_comparison}

\begin{figure}[!htbp]
  \centering
  \begin{subfigure}[b]{0.48\linewidth}
    \centering
    \includegraphics[width=\linewidth]{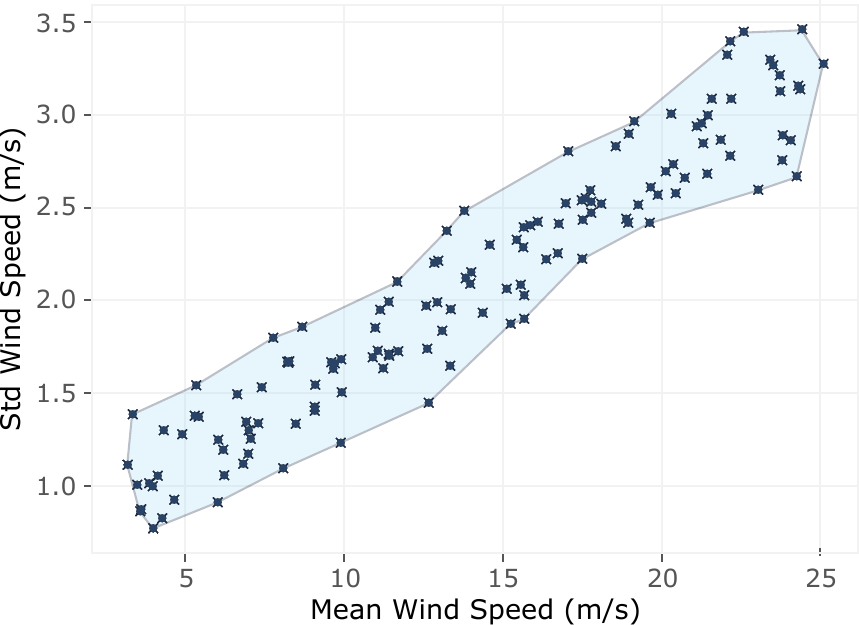}
  \end{subfigure}\hfill
  \begin{subfigure}[b]{0.48\linewidth}
    \centering
    \includegraphics[width=\linewidth]{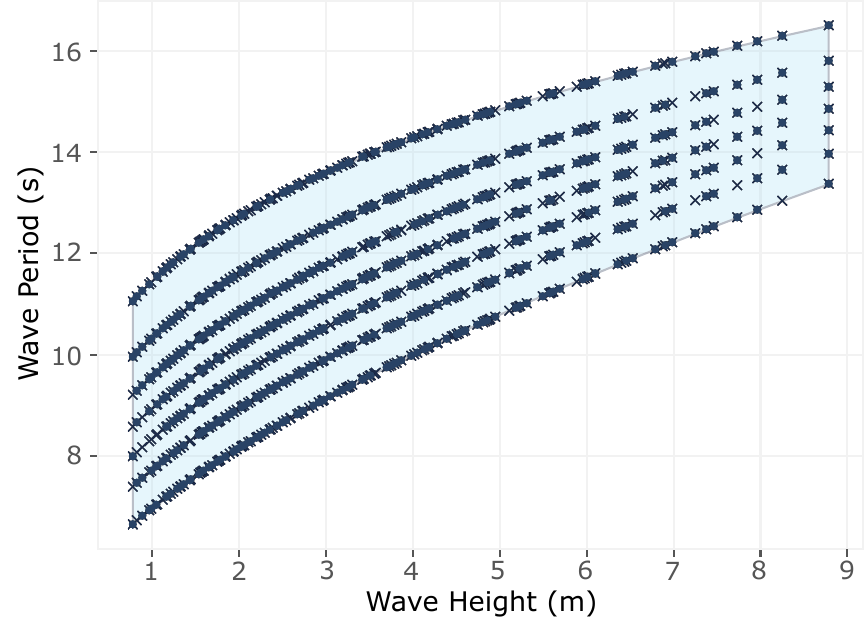}
  \end{subfigure}\\[2pt]
  \includegraphics[width=\linewidth]{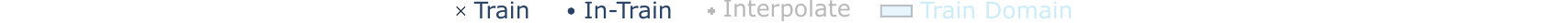}\\[2pt]
  \begin{subfigure}[b]{0.48\linewidth}
    \centering
    \caption{Wind subspace}
    \label{fig:app_random_subspaces_wind}
  \end{subfigure}\hfill
  \begin{subfigure}[b]{0.48\linewidth}
    \centering
    \caption{Wave subspace}
    \label{fig:app_random_subspaces_wave}
  \end{subfigure}
  \caption{Random validation regime partition; the alpha-shape hull is shaded. All test points fall in \textsc{IT}/\textsc{IP} regimes, with no \textsc{EX} samples.}
  \label{fig:app_random_subspaces}
\end{figure}

\autoref{tab:partition_regime_comparison} contrasts the regime composition of the test set under the random and regime-aware partitions; both use $142{,}200$ test rows per tower. Under the random partition (\autoref{fig:app_random_subspaces}), training samples cover the full envelope, so the alpha-shape hull covers the entire envelope: no test point lies outside the hull, every \textsc{EX} regime is empty, and the test set collapses to \textsc{IT\_IT} ($86\%$) and \textsc{IT\_IP} ($14\%$). Under the regime-aware partition, training is restricted to the inner envelope.

\begin{table}[!htbp]
  \caption{Test-set composition by regime, under random vs regime-aware partitions.}
  \label{tab:partition_regime_comparison}
  \centering
  \scriptsize
  \begin{tabular}{l r r r r}
    \toprule
    & \multicolumn{2}{c}{Random} & \multicolumn{2}{c}{Regime-aware} \\
    \cmidrule(lr){2-3}\cmidrule(lr){4-5}
    Regime cell (wind\_wave) & rows & \% & rows & \% \\
    \midrule
    IT\_IT & $122{,}760$ & $86.33$ & $22{,}560$ & $15.86$ \\
    IT\_IP & $19{,}440$  & $13.67$ & $37{,}350$ & $26.27$ \\
    IT\_EX & $0$  & $0.00$  & $51{,}420$ & $36.16$ \\
    IP\_IT & $0$  & $0.00$  & $4{,}920$  & $3.46$  \\
    IP\_IP & $0$  & $0.00$  & $5{,}280$  & $3.71$  \\
    IP\_EX & $0$  & $0.00$  & $4{,}500$  & $3.16$  \\
    EX\_IT & $0$  & $0.00$  & $2{,}760$  & $1.94$  \\
    EX\_IP & $0$  & $0.00$  & $5{,}790$  & $4.07$  \\
    \textbf{EX\_EX} & \textbf{0}  & $\mathbf{0.00}$ & $\mathbf{7{,}620}$ & $\mathbf{5.36}$ \\
    \midrule
    Total & $142{,}200$ & $100$ & $142{,}200$ & $100$ \\
    \bottomrule
  \end{tabular}
\end{table}

\vspace*{1em}
\subsection{Baseline models}
\label{app:baseline_models}

We describe the eight surrogate families evaluated. Per-cell
counts of configurations actually trained are in
\autoref{tab:model_pool}.

\paragraph{XGBoost~\citep{chen2016xgboost}.}
Gradient-boosted decision trees with histogram splits and
second-order gradient information; portfolio configurations
vary tree depth, learning rate, and regularisation.

\paragraph{LightGBM~\citep{ke2017lightgbm}.}
Histogram-based gradient boosting with leaf-wise tree growth;
portfolio configurations vary leaves, learning rate, and
feature/row sub-sampling.

\paragraph{CatBoost~\citep{prokhorenkova2018catboost}.}
Gradient boosting with oblivious (symmetric) trees and ordered
boosting to mitigate prediction shift; portfolio configurations
vary depth, learning rate, and regularisation.

\paragraph{NeuralNetFastAI~\citep{howard2020fastai}.}
Tabular MLP from the fastai library with 1-cycle learning-rate
schedule and embedded categorical features; portfolio
configurations vary layer widths, dropout, learning rate, and
weight decay.

\paragraph{NeuralNetTorch~\citep{erickson2020autogluon}.}
PyTorch tabular MLP with configurable depth, width, dropout,
and categorical embeddings, trained with Adam and early
stopping.

\paragraph{TabM~\citep{gorishniy2024tabm}.}
Parameter-efficient tabular model that ensembles multiple
prediction heads on top of a shared MLP backbone; available
under AutoGluon's \texttt{extreme} preset.

\paragraph{ExtraTrees~\citep{geurts2006extratrees}.}
Bagged decision trees with fully randomised feature splits;
single default configuration only.

\paragraph{RandomForest~\citep{breiman2001randomforest}.}
Bagged decision trees with bootstrap row sampling and random
feature subsets per split; single default configuration only.

\paragraph{Stacked ensembles via AutoGluon.}
\label{app:autogluon_presets}
The eight families above are trained as independent base
learners through AutoGluon
Tabular~\citep{erickson2020autogluon}, which constructs stacked
ensembles via greedy
\texttt{EnsembleSelection}~\citep{caruana2004ensembleselection}.
AutoGluon stacks models in three layers: base learners (L1),
level-2 ensembles (the global weighted ensemble WE\_L2 plus
per-family BAG\_L2 stackers), and an optional level-3 weighted
ensemble (WE\_L3).
\begin{itemize}
  \setlength\itemsep{2pt}
  \item \textbf{L1} (base learners): each family fits one or
  more configurations under 8-fold bagging, named
  \texttt{<Family>\_BAG\_L1} (e.g.,
  \texttt{CatBoost\_BAG\_L1},
  \texttt{NeuralNetFastAI\_r102\_BAG\_L1}); held-out fold
  predictions are stored.
  \item \textbf{WE\_L2} (greedy weighted ensemble of L1): sparse
  non-negative weights over L1 held-out predictions; the
  canonical ensemble baseline.
  \item \textbf{WE\_L3} (weighted ensemble on top of L2 + L1):
  a further weighted ensemble over L2 stacker outputs together
  with L1 predictions.
  \item \textbf{BAG\_L2} stackers: alternative L2 layer where
  each base family is refit on L1 held-out predictions plus
  original features, named \texttt{<Family>\_BAG\_L2} (e.g.,
  \texttt{NeuralNetFastAI\_BAG\_L2}); distinct from the
  WE\_L\{2,3\} weighted ensembles.
\end{itemize}

\paragraph{Foundation model exclusion.}
\label{app:model_pool}
AutoGluon's \texttt{extreme} preset nominally includes
TabPFN-v2~\citep{hollmann2024tabpfn},
TabICL~\citep{qu2025tabicl}, and
Mitra~\citep{mehta2025mitra}, but AutoGluon caps the input row
count for these models at $10{,}000$; FLOATBench's
$\approx\!41{,}500$-row training fold ($51{,}840$ rows minus the
$20\%$ internal validation hold-out, \autoref{app:training_config})
is $\approx\!4\times$ the cap, and the models abort before
training.

\begin{table}[!ht]
  \caption{Trained models per (experiment, tower or fold,
  preset), by family.}
  \label{tab:model_pool}
  \centering
  \scriptsize
  \begin{tabular}{lllrrrrrrrrrr}
    \toprule
    Exp.\ & Split & Preset & XGB & LGBM & CB & NN-F & NN-T & TabM & ET & RF & WE & Total \\
    \midrule
    \multirow{9}{*}{E1}
        & \multirow{3}{*}{\textsc{ref}}       & \texttt{best}    &  8 & 15 & 14 & 13 & 12 & 0 & 6 & 4 & 2 &  74 \\
        &                                     & \texttt{extreme} &  2 &  3 &  5 &  0 &  0 & 6 & 0 & 0 & 1 &  17 \\
        &                                     & \textbf{Total}   & \textbf{10} & \textbf{18} & \textbf{19} & \textbf{13} & \textbf{12} & \textbf{6} & \textbf{6} & \textbf{4} & \textbf{3} & \textbf{91} \\
    \cmidrule(lr){2-13}
        & \multirow{3}{*}{\textsc{opt1}}      & \texttt{best}    &  8 & 17 & 15 & 14 & 12 & 0 & 6 & 4 & 2 &  78 \\
        &                                     & \texttt{extreme} &  2 &  3 &  5 &  0 &  0 & 6 & 0 & 0 & 1 &  17 \\
        &                                     & \textbf{Total}   & \textbf{10} & \textbf{20} & \textbf{20} & \textbf{14} & \textbf{12} & \textbf{6} & \textbf{6} & \textbf{4} & \textbf{3} & \textbf{95} \\
    \cmidrule(lr){2-13}
        & \multirow{3}{*}{\textsc{opt2}}      & \texttt{best}    &  7 & 15 & 12 & 10 & 10 & 0 & 5 & 4 & 2 &  65 \\
        &                                     & \texttt{extreme} &  2 &  3 &  5 &  0 &  0 & 6 & 0 & 0 & 1 &  17 \\
        &                                     & \textbf{Total}   &  \textbf{9} & \textbf{18} & \textbf{17} & \textbf{10} & \textbf{10} & \textbf{6} & \textbf{5} & \textbf{4} & \textbf{3} & \textbf{82} \\
    \midrule
    \multirow{9}{*}{E2}
        & \multirow{3}{*}{\textsc{ref}}       & \texttt{best}    &  8 & 16 & 15 & 15 & 12 & 0 & 6 & 4 & 2 &  78 \\
        &                                     & \texttt{extreme} &  2 &  3 &  5 &  0 &  0 & 6 & 0 & 0 & 1 &  17 \\
        &                                     & \textbf{Total}   & \textbf{10} & \textbf{19} & \textbf{20} & \textbf{15} & \textbf{12} & \textbf{6} & \textbf{6} & \textbf{4} & \textbf{3} & \textbf{95} \\
    \cmidrule(lr){2-13}
        & \multirow{3}{*}{\textsc{opt1}}      & \texttt{best}    &  8 & 17 & 15 & 15 & 12 & 0 & 6 & 4 & 2 &  79 \\
        &                                     & \texttt{extreme} &  2 &  3 &  5 &  0 &  0 & 6 & 0 & 0 & 1 &  17 \\
        &                                     & \textbf{Total}   & \textbf{10} & \textbf{20} & \textbf{20} & \textbf{15} & \textbf{12} & \textbf{6} & \textbf{6} & \textbf{4} & \textbf{3} & \textbf{96} \\
    \cmidrule(lr){2-13}
        & \multirow{3}{*}{\textsc{opt2}}      & \texttt{best}    &  8 & 16 & 15 & 15 & 12 & 0 & 6 & 4 & 2 &  78 \\
        &                                     & \texttt{extreme} &  2 &  3 &  5 &  0 &  0 & 6 & 0 & 0 & 1 &  17 \\
        &                                     & \textbf{Total}   & \textbf{10} & \textbf{19} & \textbf{20} & \textbf{15} & \textbf{12} & \textbf{6} & \textbf{6} & \textbf{4} & \textbf{3} & \textbf{95} \\
    \midrule
    \multirow{9}{*}{E3}
        & \multirow{3}{*}{\textsc{ref+opt1}}  & \texttt{best}    &  5 & 12 & 10 &  7 &  7 & 0 & 4 & 3 & 2 &  50 \\
        &                                     & \texttt{extreme} &  1 &  2 &  4 &  0 &  0 & 5 & 0 & 0 & 1 &  13 \\
        &                                     & \textbf{Total}   &  \textbf{6} & \textbf{14} & \textbf{14} &  \textbf{7} &  \textbf{7} & \textbf{5} & \textbf{4} & \textbf{3} & \textbf{3} & \textbf{63} \\
    \cmidrule(lr){2-13}
        & \multirow{3}{*}{\textsc{ref+opt2}}  & \texttt{best}    &  4 & 11 & 10 &  7 &  6 & 0 & 4 & 3 & 2 &  47 \\
        &                                     & \texttt{extreme} &  1 &  2 &  4 &  0 &  0 & 5 & 0 & 0 & 1 &  13 \\
        &                                     & \textbf{Total}   &  \textbf{5} & \textbf{13} & \textbf{14} &  \textbf{7} &  \textbf{6} & \textbf{5} & \textbf{4} & \textbf{3} & \textbf{3} & \textbf{60} \\
    \cmidrule(lr){2-13}
        & \multirow{3}{*}{\textsc{opt1+opt2}} & \texttt{best}    &  4 & 11 &  9 &  6 &  6 & 0 & 4 & 3 & 2 &  45 \\
        &                                     & \texttt{extreme} &  1 &  2 &  4 &  0 &  0 & 5 & 0 & 0 & 1 &  13 \\
        &                                     & \textbf{Total}   &  \textbf{5} & \textbf{13} & \textbf{13} &  \textbf{6} &  \textbf{6} & \textbf{5} & \textbf{4} & \textbf{3} & \textbf{3} & \textbf{58} \\
    \bottomrule
    \addlinespace[2pt]
    \multicolumn{13}{p{0.97\linewidth}}{\scriptsize\emph{Columns:} XGB = XGBoost; LGBM = LightGBM (incl.\ LightGBMXT, LightGBMLarge); CB = CatBoost; NN-F = NeuralNetFastAI; NN-T = NeuralNetTorch; ET = ExtraTrees (incl.\ ExtraTreesMSE); RF = RandomForest (incl.\ RandomForestMSE); WE = WeightedEnsemble.}
  \end{tabular}
\end{table}

\newpage
\subsection{Training configuration}
\label{app:training_config}

\paragraph{Validation split.}
When no explicit validation CSV is provided, the trainer carves
a $20\%$ holdout from the training partition, grouped by
simulation: every row of a given simulation lands entirely in
either the training or validation fold. This avoids the leakage
that would otherwise arise because the 30 per-section rows of
one simulation share the same wind/wave conditions and are
strongly correlated.

\paragraph{Compute.}
AutoGluon runs were trained on the local machine described in
\autoref{app:reproducibility}. Each (experiment, tower or fold,
preset) cell received 24 vCPUs and 1 GPU (12 vCPUs and 1 shared GPU per bagging fold, 8 folds) under a 4-hour wall-clock budget. The 18 cells (E1, E2, E3 across 3 towers or 3 cross-tower folds, each under \texttt{best} and \texttt{extreme}) total at
most $72$ hours of wall-clock time.

\clearpage
\subsection{Metric definitions}
\label{app:metrics_definitions}

To rigorously evaluate surrogate predictions $\hat y_i$ against
ground-truth DEL values $y_i$ ($N$ test points), we employ a set
of well-established metrics that quantify accuracy, error, and
generalization. These metrics are defined as follows.

\paragraph{Mean Absolute Error (MAE).}
The average magnitude of absolute errors between predictions and
ground truth,
\begin{equation*}
\mathrm{MAE} = \frac{1}{N}\sum_{i=1}^{N} |\hat y_i - y_i|.
\end{equation*}
This metric provides an intuitive measure of the average error
magnitude, making it easy to interpret.

\paragraph{Mean Squared Error (MSE).}
The average squared difference between predictions and ground
truth,
\begin{equation*}
\mathrm{MSE} = \frac{1}{N}\sum_{i=1}^{N} (\hat y_i - y_i)^2.
\end{equation*}
Penalizes larger deviations more severely, emphasizing outliers
and high-error regimes. Common as a regression training loss.

\paragraph{Root Mean Square Error (RMSE).}
The square root of MSE,
\begin{equation*}
\mathrm{RMSE} = \sqrt{\frac{1}{N}\sum_{i=1}^{N} (\hat y_i - y_i)^2}.
\end{equation*}
Penalizes large errors more heavily than MAE due to the squaring
operation, making it sensitive to outliers.

\paragraph{Coefficient of Determination (R\textsuperscript{2}).}
The proportion of variance in the ground truth explained by the
model,
\begin{equation*}
R^{2} = 1 - \frac{\sum_{i=1}^{N} (y_i - \hat y_i)^{2}}{\sum_{i=1}^{N} (y_i - \bar y)^{2}},
\end{equation*}
where $\bar y$ is the mean of $y_i$. A value close to $1$
indicates that the model explains most of the variance; values
close to $0$ (or negative) suggest poor predictive performance.

\paragraph{Relative L\textsuperscript{2} Error.}
The Euclidean-norm error normalized by the ground-truth norm,
\begin{equation*}
\mathrm{Rel}\,L^{2} = \frac{\|\hat y - y\|_{2}}{\|y\|_{2}}.
\end{equation*}
Scale-invariant; particularly useful for comparing models across
regimes of different magnitudes. Used as the cross-model anchor
in the leaderboards.

\paragraph{Mean Relative Error (MRE).}
The point-wise mean of relative deviations,
\begin{equation*}
\mathrm{MRE} = \frac{1}{N}\sum_{i=1}^{N} \frac{|\hat y_i - y_i|}{|y_i|}.
\end{equation*}
Each test point contributes equally, preserving the regime
hierarchy under skewed magnitude distributions. Used as the
per-regime error metric.

\paragraph{Maximum Error (MaxErr).}
The largest absolute error across the test set,
\begin{equation*}
\mathrm{MaxErr} = \max_{i} |\hat y_i - y_i|.
\end{equation*}
Captures worst-case performance, critical in fatigue applications
where large prediction errors can have safety consequences.

These metrics collectively capture absolute and relative errors,
variance explained, and worst-case behavior, providing a
comprehensive evaluation framework. Lower is better for every
metric except R\textsuperscript{2}.

\subsection{Bootstrap confidence intervals}
\label{app:bootstrap_ci_method}

To quantify the statistical uncertainty of the leaderboard
metrics, we use percentile bootstrap resampling~\citep{efron1979bootstrap}.
This nonparametric approach makes no distributional assumptions
and is robust to the heavy-tailed damage residuals.

\paragraph{Bootstrap methodology.}
Given $N$ test pairs $\{(\hat y_i, y_i)\}_{i=1}^{N}$, we generate
$B$ bootstrap replicates by sampling $N$ indices with replacement;
for each replicate $b$ we compute the metric
$\theta^{(b)} = \mathcal{M}\!\left(\{(\hat y_i, y_i) : i \in I^{(b)}\}\right)$. The bootstrap estimates of
the mean and standard deviation are
\begin{equation*}
\bar\theta_{\mathrm{boot}}
= \frac{1}{B}\sum_{b=1}^{B}\theta^{(b)},
\qquad
\sigma_{\mathrm{boot}}
= \sqrt{\frac{1}{B-1}\sum_{b=1}^{B}\left(\theta^{(b)} - \bar\theta_{\mathrm{boot}}\right)^{2}},
\end{equation*}
and the percentile-method confidence interval is
$\mathrm{CI}_{1-\alpha} = [\theta^{*}_{\alpha/2}, \theta^{*}_{1-\alpha/2}]$,
where $\theta^{*}_{p}$ is the $p$-th percentile of the bootstrap
distribution.

\paragraph{Implementation details.}
We draw $B = 2{,}000$ resamples at the $95\%$ confidence level
($\alpha = 0.05$) with RNG seed $s = 42$. The eight leaderboard
metrics (\autoref{app:metrics_definitions}) are computed in a
single bootstrap pass with shared residual, absolute-error and
norm intermediates (\autoref{alg:bootstrap_ci}).

\begin{algorithm}[!ht]
\caption{\textsc{BootstrapRegressionMetrics}: percentile
bootstrap for the leaderboard metrics.}
\label{alg:bootstrap_ci}
\small
\begin{algorithmic}[1]
\Require Predictions and ground truth $\{(\hat y_i, y_i)\}_{i=1}^{N}$,
metric set $\mathcal{M} = \{f_m\}_{m=1}^{M}$, resamples $B$
(default 2000), significance level $\alpha$ (default 0.05),
RNG seed $s$ (default 42)
\Ensure For each metric $m$: bootstrap mean $\bar\theta_m$,
standard deviation $\sigma_m$, percentile CI
$[\theta^{*}_{m,\alpha/2}, \theta^{*}_{m,1-\alpha/2}]$
\State Allocate $\Theta_m \in \mathbb{R}^{B}$ for every $m \in \mathcal{M}$ and seed RNG with $s$
\For{$b = 1,\dots,B$}
  \State Sample $N$ indices with replacement: $I^{(b)} \sim \mathrm{Uniform}\{1,\dots,N\}$
  \State $(\hat y^{(b)}, y^{(b)}) \gets (\hat y_{I^{(b)}},\, y_{I^{(b)}})$
  \For{$m = 1,\dots,M$}
    \State $\Theta_m[b] \gets f_m(\hat y^{(b)}, y^{(b)})$
    \Comment{shared intermediates ($\hat y - y$, $|\hat y - y|$, $\|y\|$) reused across $m$}
  \EndFor
\EndFor
\For{$m = 1,\dots,M$}
  \State $\bar\theta_m \gets \tfrac{1}{B}\sum_{b=1}^{B}\Theta_m[b]$
  \State $\sigma_m \gets \sqrt{\tfrac{1}{B-1}\sum_{b=1}^{B}(\Theta_m[b] - \bar\theta_m)^{2}}$ \Comment{$\mathrm{ddof}=1$}
  \State $\theta^{*}_{m,\alpha/2},\, \theta^{*}_{m,1-\alpha/2} \gets \mathrm{percentile}_{\mathrm{linear}}(\Theta_m;\, \alpha/2,\, 1-\alpha/2)$
\EndFor
\State \Return $\{(\bar\theta_m,\, \sigma_m,\, [\theta^{*}_{m,\alpha/2},\, \theta^{*}_{m,1-\alpha/2}])\}_{m=1}^{M}$
\end{algorithmic}
\end{algorithm}

\paragraph{Interpretation.}
The bootstrap CIs serve four roles:
(i) \emph{uncertainty quantification}: distinguishes genuine
model-quality gaps from finite-sample fluctuations;
(ii) \emph{distribution-free robustness}: avoids normality and
homoscedasticity assumptions, important for the heavy-tailed
fatigue damage distribution;
(iii) \emph{statistical comparability}: overlapping $95\%$ CIs
indicate that apparent gaps may not be statistically significant;
(iv) \emph{stability assessment}: narrow CIs indicate consistent
generalization, wider CIs flag sensitivity to test-set
composition. A worked example contrasting overlapping and
disjoint CIs is shown in \autoref{tab:bootstrap_example}.

\begin{table}[!b]
\centering
\small
\caption{Two hypothetical surrogate pairs illustrating overlapping
vs.\ disjoint $95\%$ CIs. Each row is one case; the two surrogates
are labelled $S_1$ and $S_2$.}
\label{tab:bootstrap_example}
\begin{tabular}{c c c c}
\toprule
Case & $S_1$: $R^2$, $95\%$ CI & $S_2$: $R^2$, $95\%$ CI & Verdict \\
\midrule
A & $0.950$, $[0.940, 0.960]$ & $0.945$, $[0.929, 0.961]$ & CIs overlap; not distinguishable \\
B & $0.950$, $[0.946, 0.954]$ & $0.930$, $[0.926, 0.934]$ & CIs disjoint; $S_1$ significantly better \\
\bottomrule
\end{tabular}
\end{table}

\clearpage
\section{Leaderboards}
\label{app:leaderboards}
\label{app:family_ranks}

This appendix collects the top-10 leaderboards used in the main text. All three tables share the same 13 columns (latency, throughput, training time, MSE, MAE, RMSE, MRE, R\textsuperscript{2}, Rel L\textsuperscript{2} DEL, MaxErr, plus 95\,\% percentile-bootstrap CIs) and are sorted by Rel L\textsuperscript{2} DEL.

\subsection{Top-10 surrogates under random (E1), per tower}
\label{app:top10_e1}

\autoref{tab:app_top10_e1} covers the E1 random partition, per tower.

\begin{sidewaystable}
  \caption{\textbf{Top-10 random (E1) per tower, sorted by Rel L\textsuperscript{2} DEL.} Error metrics on the DEL target; cells are $\bar\theta_{\mathrm{boot}} \pm \sigma_{\mathrm{boot}}$ (bootstrap mean $\pm$ bootstrap standard deviation, $B=2{,}000$ resamples). Full $95\%$ percentile-bootstrap intervals are computed by \autoref{alg:bootstrap_ci} and released alongside the leaderboard CSVs.}
  \label{tab:app_top10_e1}
  \centering
  \resizebox{\textheight}{!}{%
  \scriptsize
  \setlength{\tabcolsep}{2.5pt}
  \begin{tabular}{c l l r r r l l l l l l l}
    \toprule
    Rk & Model & Preset & Lat. & Thru & Train & MSE & MAE & RMSE & MRE\,(\%) & R\textsuperscript{2} & Rel L\textsuperscript{2} & MaxErr \\
    \midrule

    \multicolumn{13}{l}{\emph{\textsc{ref}}} \\
    1 & \texttt{WeightedEnsemble\_L2} & extreme & 0.052 & 19374 & 1237 & $(2.882\pm0.016)\!\cdot\!10^{-7}$ & $(3.9628\pm0.0096)\!\cdot\!10^{-4}$ & $(5.368\pm0.015)\!\cdot\!10^{-4}$ & $2.2910\pm0.0096$ & $0.998110\pm0.000011$ & $0.019100\pm0.000053$ & $0.004139\pm0.000053$ \\
    2 & \texttt{TabM\_r191\_BAG\_L1} & extreme & 0.039 & 25449 & 911 & $(3.147\pm0.017)\!\cdot\!10^{-7}$ & $(4.1844\pm0.0099)\!\cdot\!10^{-4}$ & $(5.610\pm0.015)\!\cdot\!10^{-4}$ & $2.892\pm0.017$ & $0.997936\pm0.000011$ & $0.019960\pm0.000054$ & $0.00472\pm0.00011$ \\
    3 & \texttt{CatBoost\_r177\_BAG\_L1} & best & 0.002 & 612323 & 201 & $(3.178\pm0.017)\!\cdot\!10^{-7}$ & $(4.171\pm0.010)\!\cdot\!10^{-4}$ & $(5.637\pm0.015)\!\cdot\!10^{-4}$ & $2.1368\pm0.0067$ & $0.997916\pm0.000012$ & $0.020058\pm0.000053$ & $0.003952\pm0.000040$ \\
    4 & \texttt{WeightedEnsemble\_L2} & best & 1.006 & 994 & 2546 & $(3.180\pm0.017)\!\cdot\!10^{-7}$ & $(4.158\pm0.010)\!\cdot\!10^{-4}$ & $(5.639\pm0.015)\!\cdot\!10^{-4}$ & $2.1043\pm0.0065$ & $0.997915\pm0.000012$ & $0.020063\pm0.000053$ & $0.004072\pm0.000049$ \\
    5 & \texttt{CatBoost\_r69\_BAG\_L1} & best & 0.002 & 496013 & 234 & $(3.180\pm0.017)\!\cdot\!10^{-7}$ & $(4.186\pm0.010)\!\cdot\!10^{-4}$ & $(5.639\pm0.015)\!\cdot\!10^{-4}$ & $2.1494\pm0.0068$ & $0.997915\pm0.000012$ & $0.020064\pm0.000052$ & $0.003958\pm0.000037$ \\
    6 & \texttt{WeightedEnsemble\_L3} & best & 1.001 & 999 & 1136 & $(3.191\pm0.017)\!\cdot\!10^{-7}$ & $(4.158\pm0.010)\!\cdot\!10^{-4}$ & $(5.649\pm0.015)\!\cdot\!10^{-4}$ & $2.0788\pm0.0063$ & $0.997908\pm0.000012$ & $0.020098\pm0.000053$ & $0.004068\pm0.000044$ \\
    7 & \texttt{CatBoost\_BAG\_L1} & best & 0.001 & 668353 & 197 & $(3.250\pm0.017)\!\cdot\!10^{-7}$ & $(4.251\pm0.010)\!\cdot\!10^{-4}$ & $(5.701\pm0.015)\!\cdot\!10^{-4}$ & $2.2015\pm0.0070$ & $0.997869\pm0.000012$ & $0.020283\pm0.000053$ & $0.003933\pm0.000012$ \\
    8 & \texttt{TabM\_BAG\_L1} & extreme & 0.010 & 97355 & 195 & $(3.266\pm0.019)\!\cdot\!10^{-7}$ & $(4.190\pm0.010)\!\cdot\!10^{-4}$ & $(5.715\pm0.017)\!\cdot\!10^{-4}$ & $2.2206\pm0.0075$ & $0.997858\pm0.000013$ & $0.020335\pm0.000058$ & $0.004474\pm0.000028$ \\
    9 & \texttt{CatBoost\_BAG\_L1} & extreme & 0.001 & 730642 & 183 & $(3.287\pm0.017)\!\cdot\!10^{-7}$ & $(4.285\pm0.010)\!\cdot\!10^{-4}$ & $(5.733\pm0.015)\!\cdot\!10^{-4}$ & $2.2221\pm0.0070$ & $0.997845\pm0.000012$ & $0.020398\pm0.000052$ & $0.003940\pm0.000034$ \\
    10 & \texttt{CatBoost\_r91\_BAG\_L1} & extreme & 0.002 & 490570 & 131 & $(3.289\pm0.018)\!\cdot\!10^{-7}$ & $(4.234\pm0.010)\!\cdot\!10^{-4}$ & $(5.735\pm0.016)\!\cdot\!10^{-4}$ & $2.1415\pm0.0067$ & $0.997843\pm0.000013$ & $0.020404\pm0.000054$ & $0.004057\pm0.000032$ \\
    \midrule
    \multicolumn{13}{l}{\emph{\textsc{opt1}}} \\
    1 & \texttt{WeightedEnsemble\_L2} & extreme & 0.041 & 24668 & 1234 & $(1.2542\pm0.0078)\!\cdot\!10^{-7}$ & $(2.4943\pm0.0067)\!\cdot\!10^{-4}$ & $(3.541\pm0.011)\!\cdot\!10^{-4}$ & $3.5025\pm0.0095$ & $0.989303\pm0.000080$ & $0.04065\pm0.00012$ & $0.002524\pm0.000037$ \\
    2 & \texttt{CatBoost\_BAG\_L1} & best & 0.002 & 443508 & 268 & $(1.2552\pm0.0079)\!\cdot\!10^{-7}$ & $(2.5228\pm0.0067)\!\cdot\!10^{-4}$ & $(3.543\pm0.011)\!\cdot\!10^{-4}$ & $3.5071\pm0.0092$ & $0.989294\pm0.000081$ & $0.04067\pm0.00013$ & $0.002579\pm0.000038$ \\
    3 & \texttt{CatBoost\_BAG\_L1} & extreme & 0.002 & 538195 & 249 & $(1.2555\pm0.0078)\!\cdot\!10^{-7}$ & $(2.5266\pm0.0067)\!\cdot\!10^{-4}$ & $(3.543\pm0.011)\!\cdot\!10^{-4}$ & $3.5177\pm0.0092$ & $0.989292\pm0.000081$ & $0.04068\pm0.00012$ & $0.002576\pm0.000038$ \\
    4 & \texttt{WeightedEnsemble\_L2} & best & 0.296 & 3374 & 843 & $(1.2608\pm0.0080)\!\cdot\!10^{-7}$ & $(2.5125\pm0.0067)\!\cdot\!10^{-4}$ & $(3.551\pm0.011)\!\cdot\!10^{-4}$ & $3.4575\pm0.0088$ & $0.989247\pm0.000082$ & $0.04076\pm0.00013$ & $0.002627\pm0.000033$ \\
    5 & \texttt{WeightedEnsemble\_L3} & best & 0.296 & 3374 & 843 & $(1.2608\pm0.0080)\!\cdot\!10^{-7}$ & $(2.5125\pm0.0067)\!\cdot\!10^{-4}$ & $(3.551\pm0.011)\!\cdot\!10^{-4}$ & $3.4575\pm0.0088$ & $0.989247\pm0.000082$ & $0.04076\pm0.00013$ & $0.002627\pm0.000033$ \\
    6 & \texttt{CatBoost\_r177\_BAG\_L1} & best & 0.002 & 424469 & 267 & $(1.2629\pm0.0080)\!\cdot\!10^{-7}$ & $(2.5136\pm0.0068)\!\cdot\!10^{-4}$ & $(3.554\pm0.011)\!\cdot\!10^{-4}$ & $3.4596\pm0.0089$ & $0.989229\pm0.000082$ & $0.04079\pm0.00013$ & $0.002618\pm0.000036$ \\
    7 & \texttt{CatBoost\_r137\_BAG\_L1} & best & 0.002 & 444628 & 215 & $(1.2810\pm0.0080)\!\cdot\!10^{-7}$ & $(2.5747\pm0.0067)\!\cdot\!10^{-4}$ & $(3.579\pm0.011)\!\cdot\!10^{-4}$ & $3.6193\pm0.0096$ & $0.989075\pm0.000082$ & $0.04109\pm0.00013$ & $0.002606\pm0.000037$ \\
    8 & \texttt{CatBoost\_r13\_BAG\_L1} & best & 0.002 & 447179 & 390 & $(1.3071\pm0.0076)\!\cdot\!10^{-7}$ & $(2.6290\pm0.0067)\!\cdot\!10^{-4}$ & $(3.615\pm0.011)\!\cdot\!10^{-4}$ & $3.7186\pm0.0097$ & $0.988852\pm0.000080$ & $0.04150\pm0.00012$ & $0.002517\pm0.000042$ \\
    9 & \texttt{CatBoost\_r51\_BAG\_L1} & extreme & 0.003 & 361347 & 129 & $(1.3269\pm0.0078)\!\cdot\!10^{-7}$ & $(2.6245\pm0.0068)\!\cdot\!10^{-4}$ & $(3.643\pm0.011)\!\cdot\!10^{-4}$ & $3.6281\pm0.0092$ & $0.988683\pm0.000082$ & $0.04182\pm0.00012$ & $0.002569\pm0.000034$ \\
    10 & \texttt{CatBoost\_r69\_BAG\_L1} & best & 0.001 & 1928116 & 66 & $(1.3523\pm0.0077)\!\cdot\!10^{-7}$ & $(2.6976\pm0.0067)\!\cdot\!10^{-4}$ & $(3.677\pm0.010)\!\cdot\!10^{-4}$ & $3.829\pm0.010$ & $0.988467\pm0.000081$ & $0.04221\pm0.00012$ & $0.002515\pm0.000039$ \\
    \midrule
    \multicolumn{13}{l}{\emph{\textsc{opt2}}} \\
    1 & \texttt{WeightedEnsemble\_L2} & extreme & 0.079 & 12612 & 1892 & $(1.4854\pm0.0095)\!\cdot\!10^{-7}$ & $(2.6936\pm0.0073)\!\cdot\!10^{-4}$ & $(3.854\pm0.012)\!\cdot\!10^{-4}$ & $3.514\pm0.010$ & $0.988611\pm0.000084$ & $0.04051\pm0.00013$ & $0.002900\pm0.000016$ \\
    2 & \texttt{CatBoost\_BAG\_L1} & best & 0.002 & 466466 & 281 & $(1.5114\pm0.0093)\!\cdot\!10^{-7}$ & $(2.7614\pm0.0072)\!\cdot\!10^{-4}$ & $(3.888\pm0.012)\!\cdot\!10^{-4}$ & $3.4771\pm0.0090$ & $0.988412\pm0.000083$ & $0.04086\pm0.00012$ & $0.002819\pm0.000016$ \\
    3 & \texttt{CatBoost\_BAG\_L1} & extreme & 0.003 & 304684 & 396 & $(1.5120\pm0.0093)\!\cdot\!10^{-7}$ & $(2.7603\pm0.0072)\!\cdot\!10^{-4}$ & $(3.888\pm0.012)\!\cdot\!10^{-4}$ & $3.4748\pm0.0090$ & $0.988407\pm0.000083$ & $0.04087\pm0.00012$ & $0.002838\pm0.000017$ \\
    4 & \texttt{CatBoost\_r177\_BAG\_L1} & best & 0.002 & 498429 & 251 & $(1.5150\pm0.0093)\!\cdot\!10^{-7}$ & $(2.7559\pm0.0072)\!\cdot\!10^{-4}$ & $(3.892\pm0.012)\!\cdot\!10^{-4}$ & $3.4596\pm0.0090$ & $0.988384\pm0.000083$ & $0.04091\pm0.00012$ & $0.002762\pm0.000017$ \\
    5 & \texttt{CatBoost\_r137\_BAG\_L1} & best & 0.002 & 517907 & 217 & $(1.5367\pm0.0093)\!\cdot\!10^{-7}$ & $(2.8171\pm0.0072)\!\cdot\!10^{-4}$ & $(3.920\pm0.012)\!\cdot\!10^{-4}$ & $3.5862\pm0.0094$ & $0.988218\pm0.000083$ & $0.04120\pm0.00012$ & $0.002864\pm0.000019$ \\
    6 & \texttt{WeightedEnsemble\_L3} & best & 0.009 & 113961 & 244 & $(1.5420\pm0.0095)\!\cdot\!10^{-7}$ & $(2.7934\pm0.0073)\!\cdot\!10^{-4}$ & $(3.927\pm0.012)\!\cdot\!10^{-4}$ & $3.5078\pm0.0090$ & $0.988177\pm0.000085$ & $0.04127\pm0.00012$ & $0.002839\pm0.000017$ \\
    7 & \texttt{WeightedEnsemble\_L2} & best & 0.009 & 113969 & 244 & $(1.5420\pm0.0095)\!\cdot\!10^{-7}$ & $(2.7934\pm0.0073)\!\cdot\!10^{-4}$ & $(3.927\pm0.012)\!\cdot\!10^{-4}$ & $3.5078\pm0.0090$ & $0.988177\pm0.000085$ & $0.04127\pm0.00012$ & $0.002839\pm0.000017$ \\
    8 & \texttt{CatBoost\_r13\_BAG\_L1} & best & 0.002 & 523928 & 404 & $(1.5543\pm0.0090)\!\cdot\!10^{-7}$ & $(2.8567\pm0.0071)\!\cdot\!10^{-4}$ & $(3.942\pm0.011)\!\cdot\!10^{-4}$ & $3.6681\pm0.0096$ & $0.988083\pm0.000082$ & $0.04144\pm0.00012$ & $0.002705\pm0.000011$ \\
    9 & \texttt{CatBoost\_r51\_BAG\_L1} & extreme & 0.003 & 342615 & 154 & $(1.5851\pm0.0092)\!\cdot\!10^{-7}$ & $(2.8685\pm0.0072)\!\cdot\!10^{-4}$ & $(3.981\pm0.012)\!\cdot\!10^{-4}$ & $3.6095\pm0.0091$ & $0.987847\pm0.000083$ & $0.04185\pm0.00012$ & $0.002750\pm0.000016$ \\
    10 & \texttt{CatBoost\_r50\_BAG\_L1} & best & 0.007 & 146355 & 27 & $(1.6129\pm0.0100)\!\cdot\!10^{-7}$ & $(2.8419\pm0.0075)\!\cdot\!10^{-4}$ & $(4.016\pm0.012)\!\cdot\!10^{-4}$ & $3.5390\pm0.0091$ & $0.987633\pm0.000089$ & $0.04221\pm0.00013$ & $0.002813\pm0.000016$ \\
    \bottomrule
  \end{tabular}%
  }
  \par\smallskip
  {\footnotesize ``Lat.'' = mean inference latency (ms); ``Thru'' = throughput (samples/s); ``Train'' = training time (s).}
\end{sidewaystable}

\subsection{Top-10 surrogates under regime-aware (E2), per tower}
\label{app:top10_e2_global}

\autoref{tab:app_top10_e2} covers the E2 regime-aware partition, per tower.

\begin{sidewaystable}
  \caption{\textbf{Top-10 regime-aware (E2) per tower, sorted by Rel L\textsuperscript{2} DEL.} Error metrics on the DEL target; cells are $\bar\theta_{\mathrm{boot}} \pm \sigma_{\mathrm{boot}}$ (bootstrap mean $\pm$ bootstrap standard deviation, $B=2{,}000$ resamples). Full $95\%$ percentile-bootstrap intervals are computed by \autoref{alg:bootstrap_ci} and released alongside the leaderboard CSVs; top-3 Rel L\textsuperscript{2} bounds are reproduced in \autoref{tab:app_top3_ci_rell2}.}
  \label{tab:app_top10_e2}
  \centering
  \resizebox{\textheight}{!}{%
  \scriptsize
  \setlength{\tabcolsep}{2.5pt}
  \begin{tabular}{c l l r r r l l l l l l l}
    \toprule
    Rk & Model & Preset & Lat. & Thru & Train & MSE & MAE & RMSE & MRE\,(\%) & R\textsuperscript{2} & Rel L\textsuperscript{2} & MaxErr \\
    \midrule

    \multicolumn{13}{l}{\emph{\textsc{ref}}} \\
    1 & \texttt{WeightedEnsemble\_L2} & best & 0.112 & 8943 & 1368 & $(3.217\pm0.029)\!\cdot\!10^{-6}$ & $0.0010049\pm0.0000040$ & $0.0017935\pm0.0000080$ & $7.335\pm0.050$ & $0.97961\pm0.00019$ & $0.06344\pm0.00028$ & $0.0120379\pm0.0000075$ \\
    2 & \texttt{NeuralNetTorch\_r22\_BAG\_L1} & best & 0.004 & 224561 & 849 & $(3.228\pm0.028)\!\cdot\!10^{-6}$ & $0.0010003\pm0.0000040$ & $0.0017966\pm0.0000078$ & $6.789\pm0.042$ & $0.97953\pm0.00019$ & $0.06355\pm0.00027$ & $0.011621\pm0.000024$ \\
    3 & \texttt{LightGBMLarge\_BAG\_L1} & best & 0.399 & 2507 & 830 & $(3.233\pm0.028)\!\cdot\!10^{-6}$ & $0.0010307\pm0.0000039$ & $0.0017979\pm0.0000079$ & $7.133\pm0.049$ & $0.97951\pm0.00019$ & $0.06359\pm0.00028$ & $0.0122058\pm0.0000051$ \\
    4 & \texttt{NeuralNetFastAI\_r145\_BAG\_L2} & best & 3.012 & 332 & 9724 & $(3.245\pm0.028)\!\cdot\!10^{-6}$ & $0.0010269\pm0.0000039$ & $0.0018014\pm0.0000077$ & $7.127\pm0.049$ & $0.97943\pm0.00018$ & $0.06372\pm0.00027$ & $0.0121078\pm0.0000087$ \\
    5 & \texttt{NeuralNetFastAI\_r11\_BAG\_L2} & best & 3.016 & 332 & 9702 & $(3.256\pm0.028)\!\cdot\!10^{-6}$ & $0.0010286\pm0.0000039$ & $0.0018044\pm0.0000078$ & $7.297\pm0.050$ & $0.97936\pm0.00019$ & $0.06382\pm0.00028$ & $0.012219\pm0.000010$ \\
    6 & \texttt{NeuralNetFastAI\_r191\_BAG\_L2} & best & 2.987 & 335 & 9649 & $(3.259\pm0.028)\!\cdot\!10^{-6}$ & $0.0010249\pm0.0000039$ & $0.0018053\pm0.0000079$ & $7.082\pm0.048$ & $0.97934\pm0.00019$ & $0.06386\pm0.00028$ & $0.0122622\pm0.0000093$ \\
    7 & \texttt{LightGBM\_r130\_BAG\_L2} & best & 3.187 & 314 & 9786 & $(3.274\pm0.029)\!\cdot\!10^{-6}$ & $0.0010241\pm0.0000040$ & $0.0018095\pm0.0000079$ & $7.069\pm0.048$ & $0.97924\pm0.00019$ & $0.06401\pm0.00028$ & $0.01258\pm0.00013$ \\
    8 & \texttt{XGBoost\_r194\_BAG\_L2} & best & 2.950 & 339 & 9541 & $(3.281\pm0.029)\!\cdot\!10^{-6}$ & $0.0010248\pm0.0000040$ & $0.0018115\pm0.0000079$ & $7.084\pm0.048$ & $0.97919\pm0.00019$ & $0.06407\pm0.00028$ & $0.012592\pm0.000071$ \\
    9 & \texttt{NeuralNetFastAI\_BAG\_L2} & best & 2.980 & 336 & 9630 & $(3.282\pm0.028)\!\cdot\!10^{-6}$ & $0.0010383\pm0.0000039$ & $0.0018117\pm0.0000078$ & $7.368\pm0.051$ & $0.97919\pm0.00019$ & $0.06408\pm0.00028$ & $0.0123405\pm0.0000085$ \\
    10 & \texttt{XGBoost\_BAG\_L2} & best & 2.957 & 338 & 9541 & $(3.283\pm0.029)\!\cdot\!10^{-6}$ & $0.0010248\pm0.0000040$ & $0.0018120\pm0.0000079$ & $7.076\pm0.048$ & $0.97918\pm0.00019$ & $0.06409\pm0.00028$ & $0.012648\pm0.000087$ \\
    \midrule
    \multicolumn{13}{l}{\emph{\textsc{opt1}}} \\
    1 & \texttt{WeightedEnsemble\_L2} & best & 0.006 & 177121 & 1941 & $(3.667\pm0.020)\!\cdot\!10^{-7}$ & $(4.096\pm0.012)\!\cdot\!10^{-4}$ & $(6.056\pm0.017)\!\cdot\!10^{-4}$ & $9.333\pm0.057$ & $0.97078\pm0.00019$ & $0.06916\pm0.00020$ & $0.003335\pm0.000033$ \\
    2 & \texttt{CatBoost\_r13\_BAG\_L1} & best & 0.002 & 461304 & 427 & $(3.883\pm0.022)\!\cdot\!10^{-7}$ & $(4.176\pm0.012)\!\cdot\!10^{-4}$ & $(6.231\pm0.018)\!\cdot\!10^{-4}$ & $9.449\pm0.057$ & $0.96906\pm0.00020$ & $0.07116\pm0.00021$ & $0.003705\pm0.000032$ \\
    3 & \texttt{WeightedEnsemble\_L3} & best & 2.879 & 347 & 9553 & $(3.889\pm0.022)\!\cdot\!10^{-7}$ & $(4.161\pm0.012)\!\cdot\!10^{-4}$ & $(6.236\pm0.018)\!\cdot\!10^{-4}$ & $9.374\pm0.057$ & $0.96901\pm0.00020$ & $0.07122\pm0.00021$ & $0.003597\pm0.000041$ \\
    4 & \texttt{CatBoost\_r69\_BAG\_L1} & best & 0.001 & 876283 & 125 & $(3.926\pm0.022)\!\cdot\!10^{-7}$ & $(4.220\pm0.012)\!\cdot\!10^{-4}$ & $(6.266\pm0.018)\!\cdot\!10^{-4}$ & $9.370\pm0.055$ & $0.96871\pm0.00021$ & $0.07156\pm0.00021$ & $0.003922\pm0.000083$ \\
    5 & \texttt{CatBoost\_BAG\_L1} & extreme & 0.002 & 560342 & 221 & $(3.931\pm0.023)\!\cdot\!10^{-7}$ & $(4.185\pm0.012)\!\cdot\!10^{-4}$ & $(6.270\pm0.018)\!\cdot\!10^{-4}$ & $9.329\pm0.056$ & $0.96868\pm0.00021$ & $0.07160\pm0.00022$ & $0.004083\pm0.000048$ \\
    6 & \texttt{CatBoost\_BAG\_L1} & best & 0.002 & 482266 & 235 & $(3.936\pm0.023)\!\cdot\!10^{-7}$ & $(4.190\pm0.012)\!\cdot\!10^{-4}$ & $(6.274\pm0.018)\!\cdot\!10^{-4}$ & $9.348\pm0.056$ & $0.96863\pm0.00021$ & $0.07165\pm0.00022$ & $0.004065\pm0.000045$ \\
    7 & \texttt{LightGBM\_r196\_BAG\_L2} & best & 2.904 & 344 & 9609 & $(3.968\pm0.023)\!\cdot\!10^{-7}$ & $(4.261\pm0.012)\!\cdot\!10^{-4}$ & $(6.299\pm0.018)\!\cdot\!10^{-4}$ & $9.227\pm0.052$ & $0.96838\pm0.00019$ & $0.07194\pm0.00021$ & $0.004502\pm0.000075$ \\
    8 & \texttt{CatBoost\_r137\_BAG\_L1} & best & 0.002 & 531274 & 205 & $(3.987\pm0.023)\!\cdot\!10^{-7}$ & $(4.267\pm0.012)\!\cdot\!10^{-4}$ & $(6.314\pm0.018)\!\cdot\!10^{-4}$ & $9.445\pm0.055$ & $0.96823\pm0.00021$ & $0.07211\pm0.00021$ & $0.003831\pm0.000054$ \\
    9 & \texttt{NeuralNetFastAI\_r11\_BAG\_L2} & best & 2.939 & 340 & 9684 & $(3.995\pm0.022)\!\cdot\!10^{-7}$ & $(4.223\pm0.012)\!\cdot\!10^{-4}$ & $(6.320\pm0.018)\!\cdot\!10^{-4}$ & $9.070\pm0.052$ & $0.96816\pm0.00020$ & $0.07218\pm0.00021$ & $0.003388\pm0.000026$ \\
    10 & \texttt{CatBoost\_r177\_BAG\_L1} & best & 0.002 & 411336 & 215 & $(4.021\pm0.023)\!\cdot\!10^{-7}$ & $(4.224\pm0.012)\!\cdot\!10^{-4}$ & $(6.341\pm0.019)\!\cdot\!10^{-4}$ & $9.496\pm0.058$ & $0.96796\pm0.00022$ & $0.07242\pm0.00022$ & $0.004087\pm0.000046$ \\
    \midrule
    \multicolumn{13}{l}{\emph{\textsc{opt2}}} \\
    1 & \texttt{WeightedEnsemble\_L2} & best & 0.010 & 100699 & 1854 & $(4.529\pm0.024)\!\cdot\!10^{-7}$ & $(4.552\pm0.013)\!\cdot\!10^{-4}$ & $(6.730\pm0.018)\!\cdot\!10^{-4}$ & $9.344\pm0.056$ & $0.96769\pm0.00019$ & $0.07036\pm0.00020$ & $0.003209\pm0.000017$ \\
    2 & \texttt{WeightedEnsemble\_L3} & best & 0.007 & 139137 & 449 & $(4.726\pm0.026)\!\cdot\!10^{-7}$ & $(4.612\pm0.013)\!\cdot\!10^{-4}$ & $(6.874\pm0.019)\!\cdot\!10^{-4}$ & $9.435\pm0.057$ & $0.96629\pm0.00020$ & $0.07187\pm0.00020$ & $0.003371\pm0.000035$ \\
    3 & \texttt{CatBoost\_r13\_BAG\_L1} & best & 0.002 & 450942 & 448 & $(4.727\pm0.026)\!\cdot\!10^{-7}$ & $(4.621\pm0.013)\!\cdot\!10^{-4}$ & $(6.876\pm0.019)\!\cdot\!10^{-4}$ & $9.516\pm0.058$ & $0.96627\pm0.00020$ & $0.07188\pm0.00021$ & $0.003511\pm0.000025$ \\
    4 & \texttt{CatBoost\_r137\_BAG\_L1} & best & 0.002 & 491779 & 209 & $(4.834\pm0.026)\!\cdot\!10^{-7}$ & $(4.723\pm0.013)\!\cdot\!10^{-4}$ & $(6.953\pm0.019)\!\cdot\!10^{-4}$ & $9.464\pm0.055$ & $0.96551\pm0.00021$ & $0.07269\pm0.00021$ & $0.003685\pm0.000091$ \\
    5 & \texttt{CatBoost\_r51\_BAG\_L1} & extreme & 0.003 & 367571 & 137 & $(4.841\pm0.027)\!\cdot\!10^{-7}$ & $(4.670\pm0.013)\!\cdot\!10^{-4}$ & $(6.958\pm0.019)\!\cdot\!10^{-4}$ & $9.501\pm0.057$ & $0.96546\pm0.00021$ & $0.07274\pm0.00021$ & $0.003780\pm0.000061$ \\
    6 & \texttt{CatBoost\_r69\_BAG\_L1} & best & 0.001 & 863375 & 137 & $(4.878\pm0.027)\!\cdot\!10^{-7}$ & $(4.705\pm0.013)\!\cdot\!10^{-4}$ & $(6.984\pm0.019)\!\cdot\!10^{-4}$ & $9.526\pm0.057$ & $0.96520\pm0.00021$ & $0.07302\pm0.00021$ & $0.003655\pm0.000089$ \\
    7 & \texttt{CatBoost\_r177\_BAG\_L1} & best & 0.002 & 586907 & 206 & $(4.941\pm0.027)\!\cdot\!10^{-7}$ & $(4.707\pm0.013)\!\cdot\!10^{-4}$ & $(7.029\pm0.019)\!\cdot\!10^{-4}$ & $9.645\pm0.059$ & $0.96475\pm0.00021$ & $0.07349\pm0.00021$ & $0.003667\pm0.000076$ \\
    8 & \texttt{NeuralNetTorch\_BAG\_L1} & best & 0.003 & 365100 & 1405 & $(4.966\pm0.025)\!\cdot\!10^{-7}$ & $(5.037\pm0.013)\!\cdot\!10^{-4}$ & $(7.047\pm0.017)\!\cdot\!10^{-4}$ & $9.681\pm0.054$ & $0.96457\pm0.00020$ & $0.07367\pm0.00019$ & $0.0032082\pm0.0000013$ \\
    9 & \texttt{CatBoost\_BAG\_L1} & best & 0.002 & 539980 & 229 & $(4.985\pm0.027)\!\cdot\!10^{-7}$ & $(4.724\pm0.014)\!\cdot\!10^{-4}$ & $(7.060\pm0.019)\!\cdot\!10^{-4}$ & $9.671\pm0.059$ & $0.96444\pm0.00022$ & $0.07382\pm0.00021$ & $0.003656\pm0.000073$ \\
    10 & \texttt{CatBoost\_BAG\_L1} & extreme & 0.003 & 384745 & 259 & $(4.989\pm0.027)\!\cdot\!10^{-7}$ & $(4.723\pm0.014)\!\cdot\!10^{-4}$ & $(7.063\pm0.019)\!\cdot\!10^{-4}$ & $9.655\pm0.059$ & $0.96441\pm0.00022$ & $0.07385\pm0.00021$ & $0.003656\pm0.000067$ \\
    \bottomrule
  \end{tabular}%
  }
  \par\smallskip
  {\footnotesize ``Lat.'' = mean inference latency (ms); ``Thru'' = throughput (samples/s); ``Train'' = training time (s).}
\end{sidewaystable}

\subsection{Top-10 surrogates under cross-tower (E3), per fold}
\label{app:cross_tower_pool}

\autoref{tab:app_top10_e3} covers the E3 cross-tower transfer protocol, per leave-one-tower-out fold.

\begin{sidewaystable}
  \caption{\textbf{Top-10 cross-tower (E3) per fold, sorted by Rel L\textsuperscript{2} DEL.} Error metrics on the DEL target; cells are $\bar\theta_{\mathrm{boot}} \pm \sigma_{\mathrm{boot}}$ (bootstrap mean $\pm$ bootstrap standard deviation, $B=2{,}000$ resamples). Full $95\%$ percentile-bootstrap intervals are computed by \autoref{alg:bootstrap_ci} and released alongside the leaderboard CSVs; top-3 Rel L\textsuperscript{2} bounds are reproduced in \autoref{tab:app_top3_ci_rell2}.}
  \label{tab:app_top10_e3}
  \centering
  \resizebox{\textheight}{!}{%
  \scriptsize
  \setlength{\tabcolsep}{2.5pt}
  \begin{tabular}{c l l r r r l l l l l l l}
    \toprule
    Rk & Model & Preset & Lat. & Thru & Train & MSE & MAE & RMSE & MRE\,(\%) & R\textsuperscript{2} & Rel L\textsuperscript{2} & MaxErr \\
    \midrule

    \multicolumn{13}{l}{\emph{\textsc{ref}\,+\,\textsc{opt1}\,$\to$\,\textsc{opt2}}} \\
    1 & \texttt{TabM\_r52\_BAG\_L1} & extreme & 0.050 & 19841 & 2847 & $(4.095\pm0.018)\!\cdot\!10^{-7}$ & $(4.9001\pm0.0090)\!\cdot\!10^{-4}$ & $(6.399\pm0.014)\!\cdot\!10^{-4}$ & $6.075\pm0.011$ & $0.96880\pm0.00016$ & $0.06701\pm0.00014$ & $0.00561\pm0.00032$ \\
    2 & \texttt{TabM\_BAG\_L1} & extreme & 0.010 & 101296 & 554 & $(5.953\pm0.033)\!\cdot\!10^{-7}$ & $(5.946\pm0.011)\!\cdot\!10^{-4}$ & $(7.715\pm0.021)\!\cdot\!10^{-4}$ & $8.057\pm0.019$ & $0.95465\pm0.00028$ & $0.08080\pm0.00022$ & $0.006991\pm0.000080$ \\
    3 & \texttt{NeuralNetFastAI\_r102\_BAG\_L1} & best & 0.007 & 144090 & 106 & $(6.894\pm0.023)\!\cdot\!10^{-7}$ & $(6.588\pm0.012)\!\cdot\!10^{-4}$ & $(8.303\pm0.014)\!\cdot\!10^{-4}$ & $9.221\pm0.022$ & $0.94747\pm0.00024$ & $0.08695\pm0.00014$ & $0.003351\pm0.000015$ \\
    4 & \texttt{NeuralNetFastAI\_r191\_BAG\_L1} & best & 0.036 & 27606 & 854 & $(7.115\pm0.024)\!\cdot\!10^{-7}$ & $(6.677\pm0.012)\!\cdot\!10^{-4}$ & $(8.435\pm0.014)\!\cdot\!10^{-4}$ & $9.439\pm0.024$ & $0.94579\pm0.00023$ & $0.08833\pm0.00014$ & $0.003732\pm0.000028$ \\
    5 & \texttt{TabM\_r184\_BAG\_L1} & extreme & 0.028 & 35584 & 5384 & $(7.421\pm0.020)\!\cdot\!10^{-7}$ & $(7.155\pm0.011)\!\cdot\!10^{-4}$ & $(8.614\pm0.012)\!\cdot\!10^{-4}$ & $8.494\pm0.011$ & $0.94346\pm0.00023$ & $0.09021\pm0.00011$ & $0.003495\pm0.000050$ \\
    6 & \texttt{NeuralNetFastAI\_r145\_BAG\_L1} & best & 0.060 & 16550 & 477 & $(7.704\pm0.026)\!\cdot\!10^{-7}$ & $(6.888\pm0.012)\!\cdot\!10^{-4}$ & $(8.777\pm0.015)\!\cdot\!10^{-4}$ & $9.420\pm0.021$ & $0.94130\pm0.00025$ & $0.09192\pm0.00015$ & $0.0037972\pm0.0000060$ \\
    7 & \texttt{NeuralNetTorch\_r22\_BAG\_L1} & best & 0.003 & 334705 & 1178 & $(8.918\pm0.031)\!\cdot\!10^{-7}$ & $(7.386\pm0.014)\!\cdot\!10^{-4}$ & $(9.443\pm0.016)\!\cdot\!10^{-4}$ & $8.626\pm0.014$ & $0.93206\pm0.00030$ & $0.09889\pm0.00015$ & $0.003836\pm0.000023$ \\
    8 & \texttt{ExtraTreesMSE\_BAG\_L1} & best & 0.003 & 370004 & 2 & $(8.989\pm0.036)\!\cdot\!10^{-7}$ & $(6.940\pm0.015)\!\cdot\!10^{-4}$ & $(9.481\pm0.019)\!\cdot\!10^{-4}$ & $8.139\pm0.016$ & $0.93151\pm0.00036$ & $0.09929\pm0.00020$ & $0.004230\pm0.000072$ \\
    9 & \texttt{NeuralNetFastAI\_BAG\_L1} & best & 0.017 & 59050 & 325 & $(9.137\pm0.035)\!\cdot\!10^{-7}$ & $(7.368\pm0.013)\!\cdot\!10^{-4}$ & $(9.559\pm0.018)\!\cdot\!10^{-4}$ & $10.708\pm0.026$ & $0.93038\pm0.00036$ & $0.10010\pm0.00020$ & $0.004778\pm0.000046$ \\
    10 & \texttt{NeuralNetTorch\_BAG\_L1} & best & 0.003 & 336238 & 1068 & $(1.0142\pm0.0035)\!\cdot\!10^{-6}$ & $(7.982\pm0.014)\!\cdot\!10^{-4}$ & $0.0010071\pm0.0000017$ & $11.517\pm0.030$ & $0.92272\pm0.00032$ & $0.10546\pm0.00016$ & $0.004177\pm0.000017$ \\
    \midrule
    \multicolumn{13}{l}{\emph{\textsc{ref}\,+\,\textsc{opt2}\,$\to$\,\textsc{opt1}}} \\
    1 & \texttt{TabM\_r52\_BAG\_L1} & extreme & 0.050 & 19848 & 3777 & $(7.271\pm0.018)\!\cdot\!10^{-7}$ & $(7.182\pm0.010)\!\cdot\!10^{-4}$ & $(8.527\pm0.011)\!\cdot\!10^{-4}$ & $10.253\pm0.016$ & $0.93836\pm0.00028$ & $0.09753\pm0.00013$ & $0.002829\pm0.000013$ \\
    2 & \texttt{NeuralNetFastAI\_r191\_BAG\_L1} & best & 0.036 & 27461 & 856 & $(7.412\pm0.024)\!\cdot\!10^{-7}$ & $(6.839\pm0.012)\!\cdot\!10^{-4}$ & $(8.609\pm0.014)\!\cdot\!10^{-4}$ & $10.344\pm0.022$ & $0.93716\pm0.00028$ & $0.09848\pm0.00016$ & $0.004430\pm0.000039$ \\
    3 & \texttt{TabM\_BAG\_L1} & extreme & 0.010 & 100599 & 673 & $(9.150\pm0.024)\!\cdot\!10^{-7}$ & $(7.948\pm0.012)\!\cdot\!10^{-4}$ & $(9.565\pm0.013)\!\cdot\!10^{-4}$ & $11.486\pm0.019$ & $0.92243\pm0.00037$ & $0.10941\pm0.00016$ & $0.00352\pm0.00012$ \\
    4 & \texttt{NeuralNetFastAI\_r102\_BAG\_L1} & best & 0.007 & 143286 & 106 & $(9.215\pm0.029)\!\cdot\!10^{-7}$ & $(7.587\pm0.013)\!\cdot\!10^{-4}$ & $(9.599\pm0.015)\!\cdot\!10^{-4}$ & $11.339\pm0.023$ & $0.92188\pm0.00037$ & $0.10980\pm0.00019$ & $0.004687\pm0.000050$ \\
    5 & \texttt{NeuralNetFastAI\_BAG\_L1} & best & 0.017 & 59323 & 320 & $(9.771\pm0.028)\!\cdot\!10^{-7}$ & $(7.936\pm0.013)\!\cdot\!10^{-4}$ & $(9.885\pm0.014)\!\cdot\!10^{-4}$ & $11.820\pm0.022$ & $0.91716\pm0.00039$ & $0.11307\pm0.00018$ & $0.00404\pm0.00011$ \\
    6 & \texttt{TabM\_r69\_BAG\_L1} & extreme & 0.028 & 35763 & 4769 & $(1.1261\pm0.0029)\!\cdot\!10^{-6}$ & $(8.853\pm0.013)\!\cdot\!10^{-4}$ & $0.0010612\pm0.0000014$ & $11.980\pm0.016$ & $0.90453\pm0.00043$ & $0.12138\pm0.00016$ & $0.0033189\pm0.0000081$ \\
    7 & \texttt{RandomForestMSE\_BAG\_L1} & best & 0.002 & 402738 & 6 & $(1.1544\pm0.0029)\!\cdot\!10^{-6}$ & $(9.071\pm0.013)\!\cdot\!10^{-4}$ & $0.0010744\pm0.0000014$ & $12.491\pm0.017$ & $0.90213\pm0.00043$ & $0.12290\pm0.00016$ & $0.003683\pm0.000034$ \\
    8 & \texttt{LightGBMXT\_BAG\_L1} & best & 0.276 & 3624 & 299 & $(1.1983\pm0.0054)\!\cdot\!10^{-6}$ & $(7.450\pm0.019)\!\cdot\!10^{-4}$ & $0.0010947\pm0.0000025$ & $9.472\pm0.021$ & $0.89841\pm0.00053$ & $0.12521\pm0.00026$ & $0.005260\pm0.000054$ \\
    9 & \texttt{XGBoost\_BAG\_L1} & best & 0.003 & 344109 & 10 & $(1.2114\pm0.0036)\!\cdot\!10^{-6}$ & $(8.955\pm0.015)\!\cdot\!10^{-4}$ & $0.0011006\pm0.0000017$ & $12.447\pm0.020$ & $0.89730\pm0.00050$ & $0.12589\pm0.00019$ & $0.00418\pm0.00014$ \\
    10 & \texttt{NeuralNetFastAI\_r145\_BAG\_L1} & best & 0.058 & 17177 & 60 & $(1.2215\pm0.0037)\!\cdot\!10^{-6}$ & $(8.933\pm0.015)\!\cdot\!10^{-4}$ & $0.0011052\pm0.0000017$ & $14.085\pm0.030$ & $0.89644\pm0.00047$ & $0.12642\pm0.00020$ & $0.005567\pm0.000044$ \\
    \midrule
    \multicolumn{13}{l}{\emph{\textsc{opt1}\,+\,\textsc{opt2}\,$\to$\,\textsc{ref}}} \\
    1 & \texttt{NeuralNetFastAI\_r191\_BAG\_L1} & best & 0.037 & 27248 & 853 & $(1.4226\pm0.0039)\!\cdot\!10^{-4}$ & $0.009574\pm0.000017$ & $0.011927\pm0.000016$ & $33.140\pm0.034$ & $0.0710\pm0.0026$ & $0.42280\pm0.00023$ & $0.03337\pm0.00025$ \\
    2 & \texttt{NeuralNetFastAI\_BAG\_L1} & best & 0.017 & 59640 & 322 & $(1.4934\pm0.0039)\!\cdot\!10^{-4}$ & $0.010019\pm0.000016$ & $0.012220\pm0.000016$ & $35.230\pm0.032$ & $0.0248\pm0.0027$ & $0.43319\pm0.00022$ & $0.03337\pm0.00025$ \\
    3 & \texttt{NeuralNetFastAI\_r102\_BAG\_L1} & best & 0.007 & 145622 & 106 & $(1.9132\pm0.0045)\!\cdot\!10^{-4}$ & $0.011539\pm0.000018$ & $0.013832\pm0.000016$ & $41.036\pm0.031$ & $-0.2493\pm0.0034$ & $0.49031\pm0.00019$ & $0.03353\pm0.00021$ \\
    4 & \texttt{NeuralNetTorch\_BAG\_L1} & best & 0.003 & 360937 & 1232 & $(2.3072\pm0.0053)\!\cdot\!10^{-4}$ & $0.012926\pm0.000018$ & $0.015189\pm0.000017$ & $47.642\pm0.027$ & $-0.5066\pm0.0040$ & $0.53843\pm0.00018$ & $0.03553\pm0.00018$ \\
    5 & \texttt{NeuralNetTorch\_r79\_BAG\_L1} & best & 0.004 & 264425 & 1130 & $(2.4820\pm0.0056)\!\cdot\!10^{-4}$ & $0.013215\pm0.000020$ & $0.015754\pm0.000018$ & $47.483\pm0.033$ & $-0.6208\pm0.0044$ & $0.55846\pm0.00021$ & $0.03555\pm0.00021$ \\
    6 & \texttt{NeuralNetFastAI\_r145\_BAG\_L1} & best & 0.058 & 17180 & 60 & $(2.4988\pm0.0057)\!\cdot\!10^{-4}$ & $0.013381\pm0.000019$ & $0.015808\pm0.000018$ & $48.142\pm0.029$ & $-0.6317\pm0.0044$ & $0.56034\pm0.00019$ & $0.035892\pm0.000098$ \\
    7 & \texttt{NeuralNetTorch\_r22\_BAG\_L1} & best & 0.003 & 361820 & 2805 & $(2.5743\pm0.0056)\!\cdot\!10^{-4}$ & $0.013816\pm0.000019$ & $0.016044\pm0.000017$ & $51.624\pm0.027$ & $-0.6810\pm0.0045$ & $0.56874\pm0.00018$ & $0.03531\pm0.00018$ \\
    8 & \texttt{LightGBMXT\_BAG\_L1} & best & 0.289 & 3460 & 302 & $(2.7952\pm0.0061)\!\cdot\!10^{-4}$ & $0.014407\pm0.000019$ & $0.016719\pm0.000018$ & $54.261\pm0.023$ & $-0.8253\pm0.0049$ & $0.59265\pm0.00020$ & $0.03658\pm0.00011$ \\
    9 & \texttt{LightGBM\_r188\_BAG\_L1} & best & 0.284 & 3527 & 431 & $(2.8330\pm0.0062)\!\cdot\!10^{-4}$ & $0.014515\pm0.000019$ & $0.016831\pm0.000018$ & $54.725\pm0.023$ & $-0.8500\pm0.0050$ & $0.59664\pm0.00019$ & $0.03703\pm0.00016$ \\
    10 & \texttt{TabM\_r52\_BAG\_L1} & extreme & 0.050 & 19846 & 5428 & $(2.8577\pm0.0070)\!\cdot\!10^{-4}$ & $0.013861\pm0.000022$ & $0.016905\pm0.000021$ & $49.893\pm0.042$ & $-0.8661\pm0.0052$ & $0.59924\pm0.00030$ & $0.03784\pm0.00019$ \\
    \bottomrule
  \end{tabular}%
  }
  \par\smallskip
  {\footnotesize ``Lat.'' = mean inference latency (ms); ``Thru'' = throughput (samples/s); ``Train'' = training time (s).}
\end{sidewaystable}

\clearpage
\subsection{Bootstrap CIs for top-3 surrogates (Rel L\texorpdfstring{\textsuperscript{2}}{2}, E2 and E3)}
\label{app:top3_ci_rell2}

The Rel L\textsuperscript{2} DEL leaderboards in
\autoref{tab:app_top10_e2} and \autoref{tab:app_top10_e3} report
$\bar\theta_{\mathrm{boot}}\pm\sigma_{\mathrm{boot}}$ for column-budget
reasons; \autoref{tab:app_top3_ci_rell2} below adds the explicit
$95\%$ percentile-bootstrap interval
$[\theta^{*}_{0.025}, \theta^{*}_{0.975}]$ for the top-3 surrogates of
each E2 tower and each E3 fold, to make CI overlap directly
inspectable. Full bounds for every metric and every rank are released
alongside the leaderboard CSVs.

\begin{table}[!htbp]
  \centering
  \caption{\textbf{Rel L\textsuperscript{2} DEL $95\%$ percentile-bootstrap CIs for the top-3 surrogates}, per E2 tower and per E3 fold. Cells are $\bar\theta_{\mathrm{boot}} \pm \sigma_{\mathrm{boot}}\,[\theta^{*}_{0.025}, \theta^{*}_{0.975}]$, with $B=2{,}000$ resamples (\autoref{alg:bootstrap_ci}).}
  \label{tab:app_top3_ci_rell2}
  \small
  \setlength{\tabcolsep}{4pt}
  \begin{tabular}{c l l}
    \toprule
    Rank & Model & Rel L\textsuperscript{2} DEL $\bar\theta \pm \sigma\,[\theta^{*}_{0.025}, \theta^{*}_{0.975}]$ \\
    \midrule
    \multicolumn{3}{l}{\emph{E2 (regime-aware), per tower}} \\
    \cmidrule(l){1-3}
    \multicolumn{3}{l}{\textsc{ref}} \\
    1 & \texttt{WeightedEnsemble\_L2}             & $0.06344\pm0.00028\,[0.0629, 0.0640]$ \\
    2 & \texttt{NeuralNetTorch\_r22\_BAG\_L1}     & $0.06355\pm0.00027\,[0.0630, 0.0641]$ \\
    3 & \texttt{LightGBMLarge\_BAG\_L1}           & $0.06359\pm0.00028\,[0.0631, 0.0641]$ \\
    \cmidrule(l){1-3}
    \multicolumn{3}{l}{\textsc{opt1}} \\
    1 & \texttt{WeightedEnsemble\_L2}             & $0.06916\pm0.00020\,[0.0688, 0.0696]$ \\
    2 & \texttt{CatBoost\_r13\_BAG\_L1}           & $0.07116\pm0.00021\,[0.0707, 0.0716]$ \\
    3 & \texttt{WeightedEnsemble\_L3}             & $0.07122\pm0.00021\,[0.0708, 0.0716]$ \\
    \cmidrule(l){1-3}
    \multicolumn{3}{l}{\textsc{opt2}} \\
    1 & \texttt{WeightedEnsemble\_L2}             & $0.07036\pm0.00020\,[0.0700, 0.0707]$ \\
    2 & \texttt{WeightedEnsemble\_L3}             & $0.07187\pm0.00020\,[0.0715, 0.0723]$ \\
    3 & \texttt{CatBoost\_r13\_BAG\_L1}           & $0.07188\pm0.00021\,[0.0715, 0.0723]$ \\
    \midrule
    \multicolumn{3}{l}{\emph{E3 (cross-tower), per fold}} \\
    \cmidrule(l){1-3}
    \multicolumn{3}{l}{\textsc{ref+opt1}\,$\to$\,\textsc{opt2}} \\
    1 & \texttt{TabM\_r52\_BAG\_L1}               & $0.06701\pm0.00014\,[0.0667, 0.0673]$ \\
    2 & \texttt{TabM\_BAG\_L1}                    & $0.08080\pm0.00022\,[0.0804, 0.0812]$ \\
    3 & \texttt{NeuralNetFastAI\_r102\_BAG\_L1}   & $0.08695\pm0.00014\,[0.0867, 0.0872]$ \\
    \cmidrule(l){1-3}
    \multicolumn{3}{l}{\textsc{ref+opt2}\,$\to$\,\textsc{opt1}} \\
    1 & \texttt{TabM\_r52\_BAG\_L1}               & $0.09753\pm0.00013\,[0.0973, 0.0978]$ \\
    2 & \texttt{NeuralNetFastAI\_r191\_BAG\_L1}   & $0.09848\pm0.00016\,[0.0982, 0.0988]$ \\
    3 & \texttt{TabM\_BAG\_L1}                    & $0.10941\pm0.00016\,[0.1090, 0.1100]$ \\
    \cmidrule(l){1-3}
    \multicolumn{3}{l}{\textsc{opt1+opt2}\,$\to$\,\textsc{ref}} \\
    1 & \texttt{NeuralNetFastAI\_r191\_BAG\_L1}   & $0.42280\pm0.00023\,[0.4220, 0.4230]$ \\
    2 & \texttt{NeuralNetFastAI\_BAG\_L1}         & $0.43319\pm0.00022\,[0.4330, 0.4340]$ \\
    3 & \texttt{NeuralNetFastAI\_r102\_BAG\_L1}   & $0.49031\pm0.00019\,[0.4900, 0.4910]$ \\
    \bottomrule
  \end{tabular}
\end{table}

\clearpage
\section{E2: how the best surrogate varies by regime, family, and section}
\label{app:e2_evidence}

\subsection{Wind and wave Extrapolation dominates over other regimes}
\label{app:ex_ex_dominance}

\begin{figure}[!htbp]
  \centering
  \begin{subfigure}{\linewidth}
    \centering
    \includegraphics[width=0.78\linewidth]{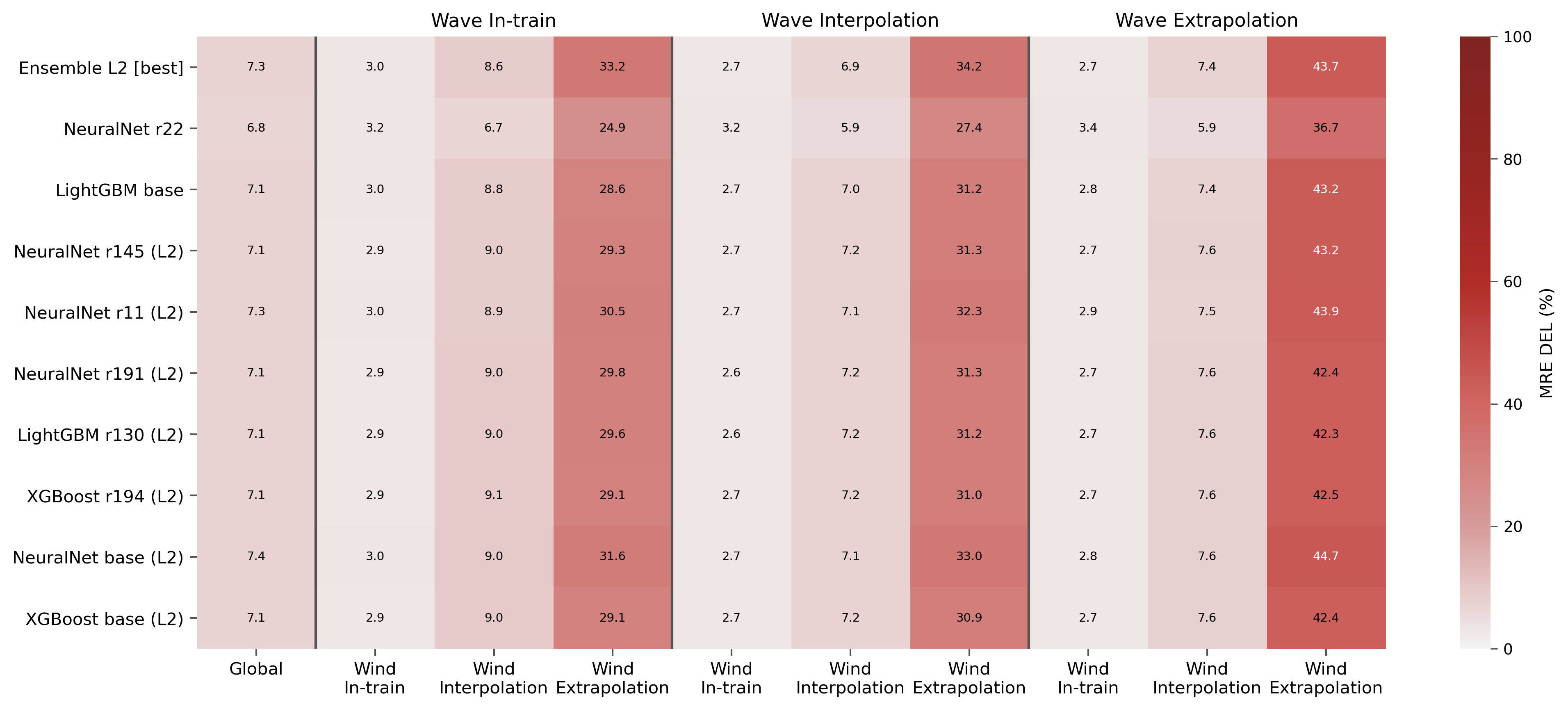}
    \caption{\textsc{ref}.}
    \label{fig:app_heatmap_ref}
  \end{subfigure}\\[6pt]
  \begin{subfigure}{\linewidth}
    \centering
    \includegraphics[width=0.78\linewidth]{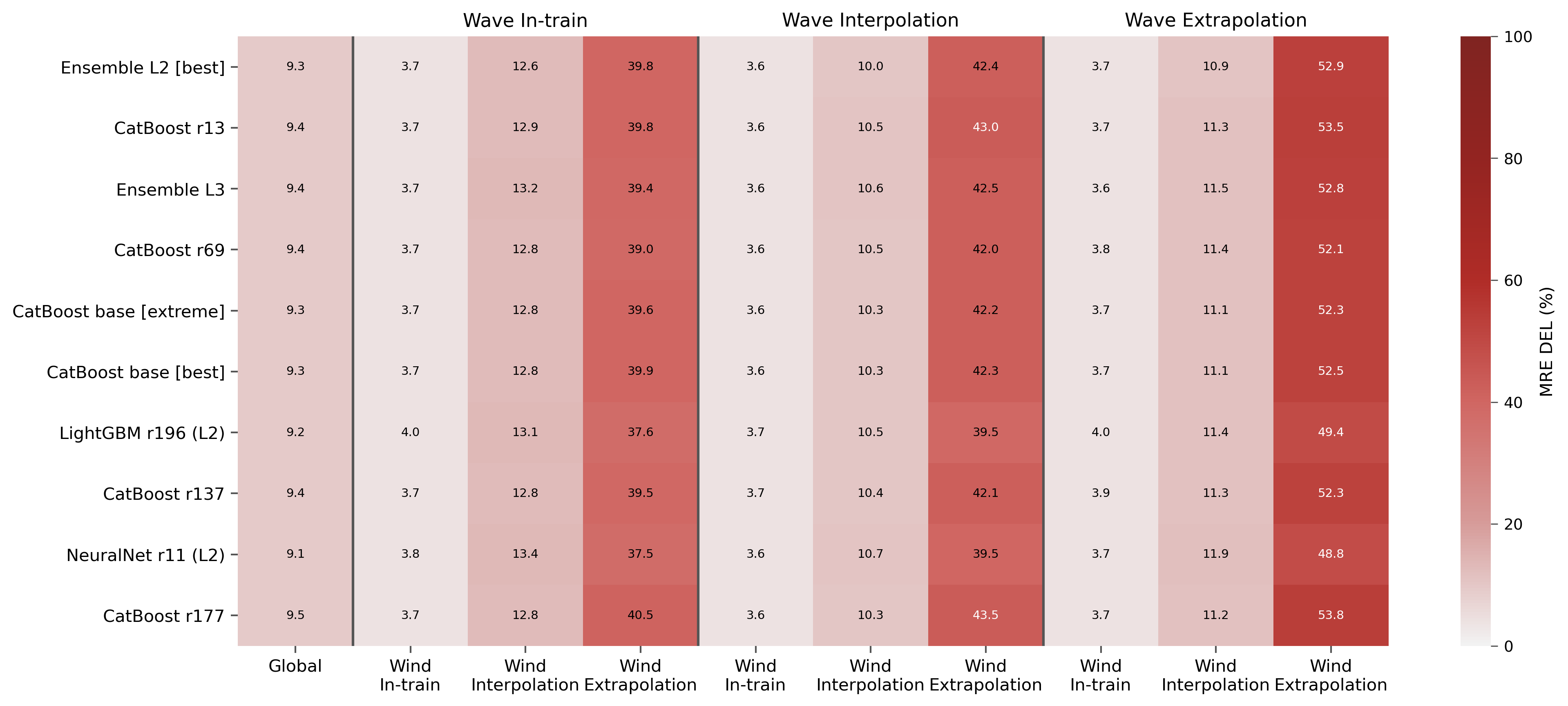}
    \caption{\textsc{opt1}.}
    \label{fig:app_heatmap_opt1}
  \end{subfigure}\\[6pt]
  \begin{subfigure}{\linewidth}
    \centering
    \includegraphics[width=0.78\linewidth]{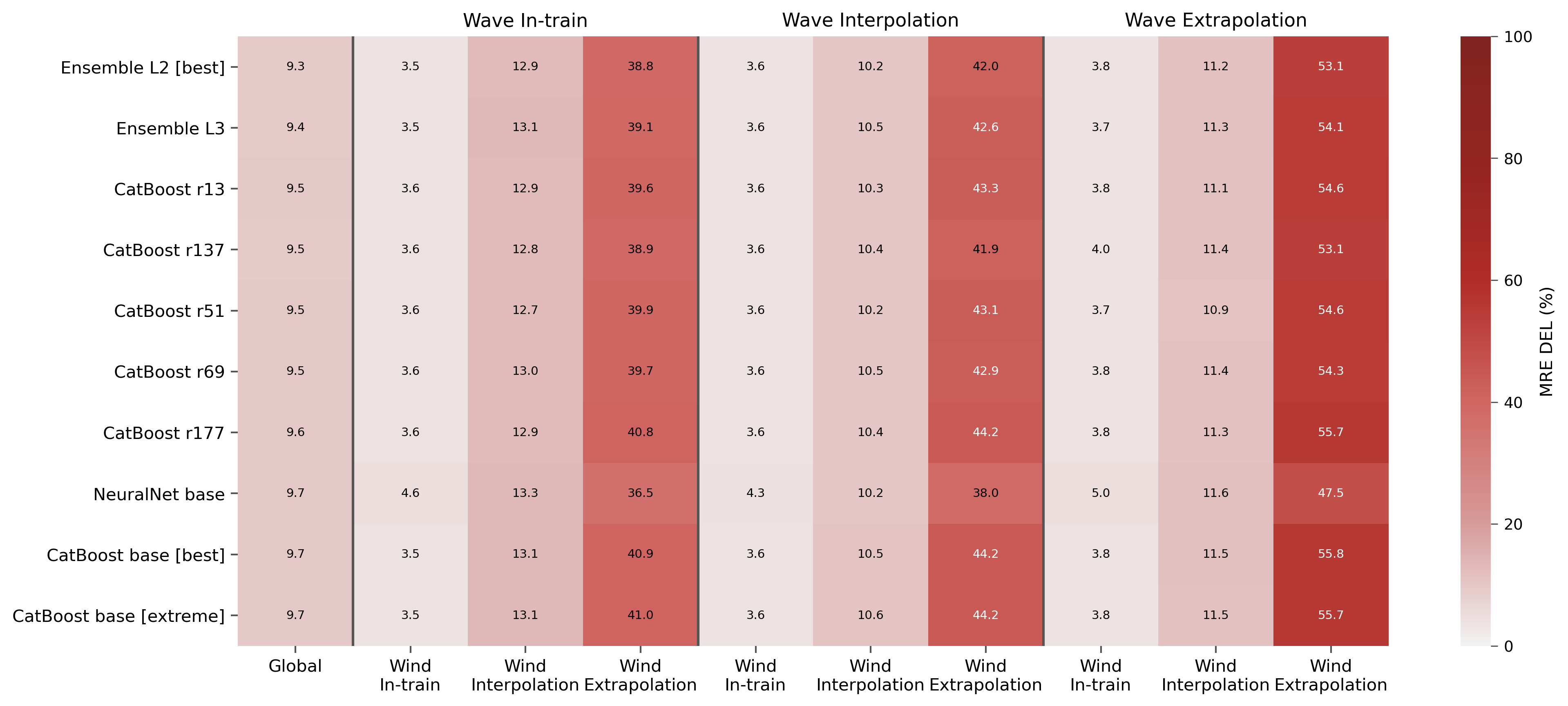}
    \caption{\textsc{opt2}.}
    \label{fig:app_heatmap_opt2}
  \end{subfigure}
  \caption{\textbf{MRE DEL across the nine regimes, three towers (E2).} Top-10 global surrogates per tower (sorted by Rel L\textsuperscript{2} DEL, \appref{app:top10_e2_global}). Columns group wave regimes; within each group, the three sub-columns are wind regimes. The rightmost \textsc{EX\_EX} column is the formal worst-case extrapolation regime by construction.}
  \label{fig:app_heatmap_3towers}
\end{figure}

The extreme wind and wave regime (\textsc{EX\_EX}) is the formal worst-case by construction. \autoref{fig:app_heatmap_3towers} shows the top-10 global surrogates (sorted by Rel L\textsuperscript{2} DEL, \appref{app:top10_e2_global}) and their MRE DEL across the nine regimes per tower; the rightmost \textsc{EX\_EX} column is the highest-error regime on all three towers.

\subsection{Top-10 surrogates on the \textsc{EX\_EX} regime under E2, per tower}
\label{app:f1_exex_top10}

\autoref{tab:app_top10_exex} lists the top-10 surrogates ranked by Rel L\textsuperscript{2} DEL on the \textsc{EX\_EX} regime per tower. Every top-10 \textsc{EX\_EX} model is a \texttt{NeuralNetFastAI} or \texttt{NeuralNetTorch} BAG\_L1 variant; most have low global ranks (bottom of the global leaderboard), while \textsc{opt2} includes two higher-global-rank \texttt{NeuralNetTorch} variants. The family-level signature is examined in \appref{app:scatter_global_exex}.

\begin{table}[!htbp]
  \caption{\textbf{Top-10 \textsc{EX\_EX} (E2) per tower, sorted by Rel L\textsuperscript{2} DEL on \textsc{EX\_EX}.} Error metric on the DEL target.}
  \label{tab:app_top10_exex}
  \label{tab:f1_exex_top5}
  \centering
  \scriptsize
  \setlength{\tabcolsep}{4pt}
  \begin{tabular}{c c l l r r}
    \toprule
    \makecell{Rank\\\textsc{EX\_EX}} & \makecell{Rank\\Global} & Model & Preset & Rel L\textsuperscript{2} Global & Rel L\textsuperscript{2} \textsc{EX\_EX} \\
    \midrule

    \multicolumn{6}{l}{\emph{\textsc{ref}}} \\
    1 & 79 & \texttt{NeuralNetFastAI\_r102\_BAG\_L1} & best & $0.079$ & $\mathbf{0.054}$ \\
    2 & 84 & \texttt{NeuralNetFastAI\_r156\_BAG\_L1} & best & $0.082$ & $\mathbf{0.062}$ \\
    3 & 80 & \texttt{NeuralNetFastAI\_r11\_BAG\_L1} & best & $0.079$ & $\mathbf{0.063}$ \\
    4 & 77 & \texttt{NeuralNetFastAI\_r103\_BAG\_L1} & best & $0.077$ & $\mathbf{0.064}$ \\
    5 & 83 & \texttt{NeuralNetTorch\_r30\_BAG\_L1} & best & $0.081$ & $\mathbf{0.065}$ \\
    6 & 74 & \texttt{NeuralNetFastAI\_r145\_BAG\_L1} & best & $0.076$ & $\mathbf{0.077}$ \\
    7 & 82 & \texttt{NeuralNetTorch\_r86\_BAG\_L1} & best & $0.081$ & $\mathbf{0.077}$ \\
    8 & 73 & \texttt{NeuralNetFastAI\_r191\_BAG\_L1} & best & $0.075$ & $\mathbf{0.080}$ \\
    9 & 76 & \texttt{NeuralNetFastAI\_BAG\_L1} & best & $0.077$ & $\mathbf{0.084}$ \\
    10 & 68 & \texttt{NeuralNetTorch\_r79\_BAG\_L1} & best & $0.072$ & $\mathbf{0.089}$ \\
    \midrule
    \multicolumn{6}{l}{\emph{\textsc{opt1}}} \\
    1 & 73 & \texttt{NeuralNetFastAI\_r102\_BAG\_L1} & best & $0.085$ & $\mathbf{0.075}$ \\
    2 & 80 & \texttt{NeuralNetFastAI\_r191\_BAG\_L1} & best & $0.089$ & $\mathbf{0.085}$ \\
    3 & 77 & \texttt{NeuralNetFastAI\_BAG\_L1} & best & $0.087$ & $\mathbf{0.086}$ \\
    4 & 78 & \texttt{NeuralNetFastAI\_r145\_BAG\_L1} & best & $0.088$ & $\mathbf{0.086}$ \\
    5 & 76 & \texttt{NeuralNetFastAI\_r103\_BAG\_L1} & best & $0.087$ & $\mathbf{0.090}$ \\
    6 & 85 & \texttt{NeuralNetTorch\_r30\_BAG\_L1} & best & $0.107$ & $\mathbf{0.102}$ \\
    7 & 64 & \texttt{NeuralNetTorch\_BAG\_L1} & best & $0.078$ & $\mathbf{0.106}$ \\
    8 & 82 & \texttt{NeuralNetFastAI\_r11\_BAG\_L1} & best & $0.092$ & $\mathbf{0.106}$ \\
    9 & 79 & \texttt{NeuralNetTorch\_r14\_BAG\_L1} & best & $0.089$ & $\mathbf{0.110}$ \\
    10 & 71 & \texttt{NeuralNetTorch\_r22\_BAG\_L1} & best & $0.083$ & $\mathbf{0.113}$ \\
    \midrule
    \multicolumn{6}{l}{\emph{\textsc{opt2}}} \\
    1 & 69 & \texttt{NeuralNetFastAI\_r102\_BAG\_L1} & best & $0.081$ & $\mathbf{0.073}$ \\
    2 & 79 & \texttt{NeuralNetFastAI\_r191\_BAG\_L1} & best & $0.089$ & $\mathbf{0.079}$ \\
    3 & 71 & \texttt{NeuralNetFastAI\_BAG\_L1} & best & $0.085$ & $\mathbf{0.081}$ \\
    4 & 80 & \texttt{NeuralNetFastAI\_r103\_BAG\_L1} & best & $0.089$ & $\mathbf{0.081}$ \\
    5 & 75 & \texttt{NeuralNetFastAI\_r145\_BAG\_L1} & best & $0.088$ & $\mathbf{0.086}$ \\
    6 & 83 & \texttt{NeuralNetTorch\_r30\_BAG\_L1} & best & $0.103$ & $\mathbf{0.094}$ \\
    7 & 78 & \texttt{NeuralNetFastAI\_r11\_BAG\_L1} & best & $0.089$ & $\mathbf{0.096}$ \\
    8 & 84 & \texttt{NeuralNetFastAI\_r143\_BAG\_L1} & best & $0.104$ & $\mathbf{0.099}$ \\
    9 & 8  & \texttt{NeuralNetTorch\_BAG\_L1} & best & $0.074$ & $\mathbf{0.103}$ \\
    10 & 44 & \texttt{NeuralNetTorch\_r22\_BAG\_L1} & best & $0.077$ & $\mathbf{0.110}$ \\
    \bottomrule
  \end{tabular}
\end{table}

\clearpage
\subsection{NeuralNet family stays closest to the Global-vs-\textsc{EX\_EX} diagonal}
\label{app:scatter_global_exex}

\begin{figure}[!htbp]
  \centering
  \begin{subfigure}{\linewidth}
    \centering
    \includegraphics[width=\linewidth]{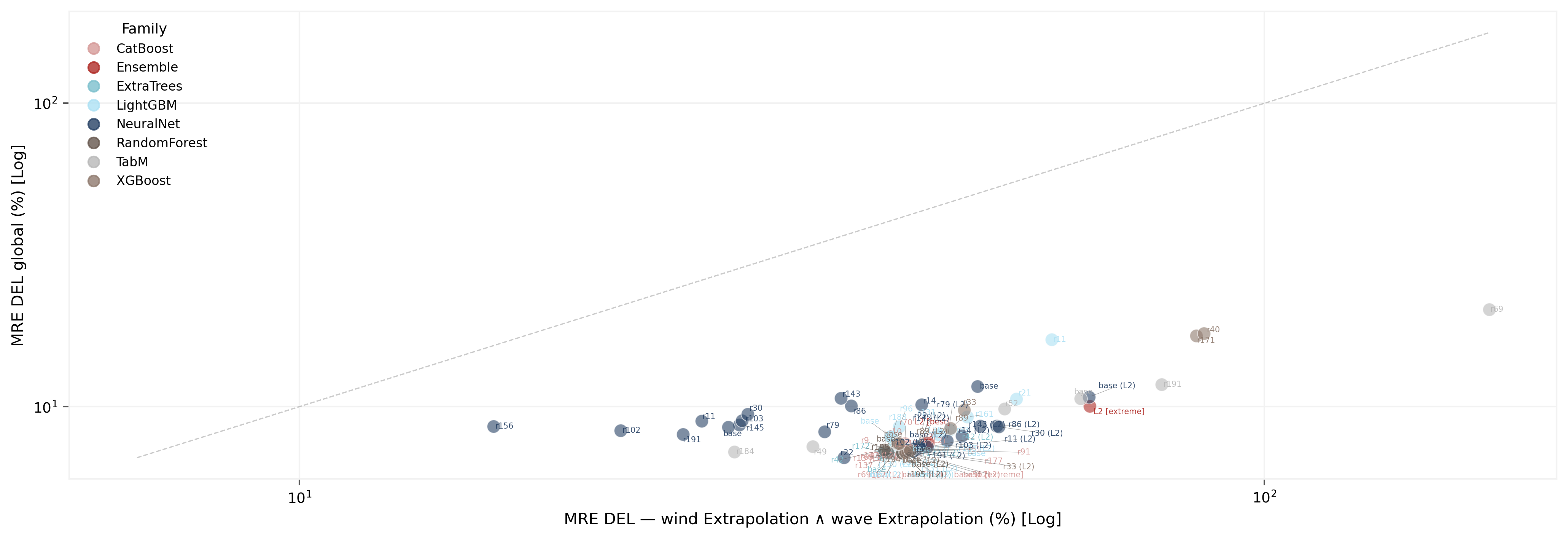}
    \caption{\textsc{ref}.}
    \label{fig:app_scatter_global_exex_ref}
  \end{subfigure}\\[6pt]
  \begin{subfigure}{\linewidth}
    \centering
    \includegraphics[width=\linewidth]{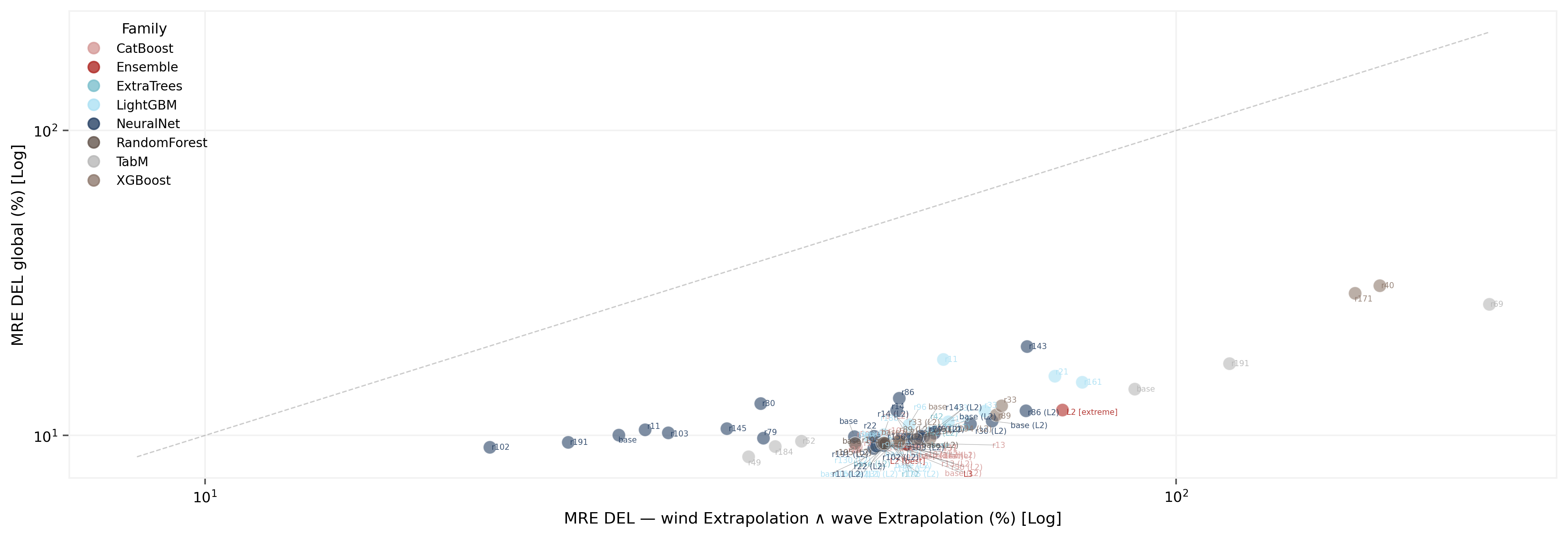}
    \caption{\textsc{opt1}.}
    \label{fig:app_scatter_global_exex_opt1}
  \end{subfigure}\\[6pt]
  \begin{subfigure}{\linewidth}
    \centering
    \includegraphics[width=\linewidth]{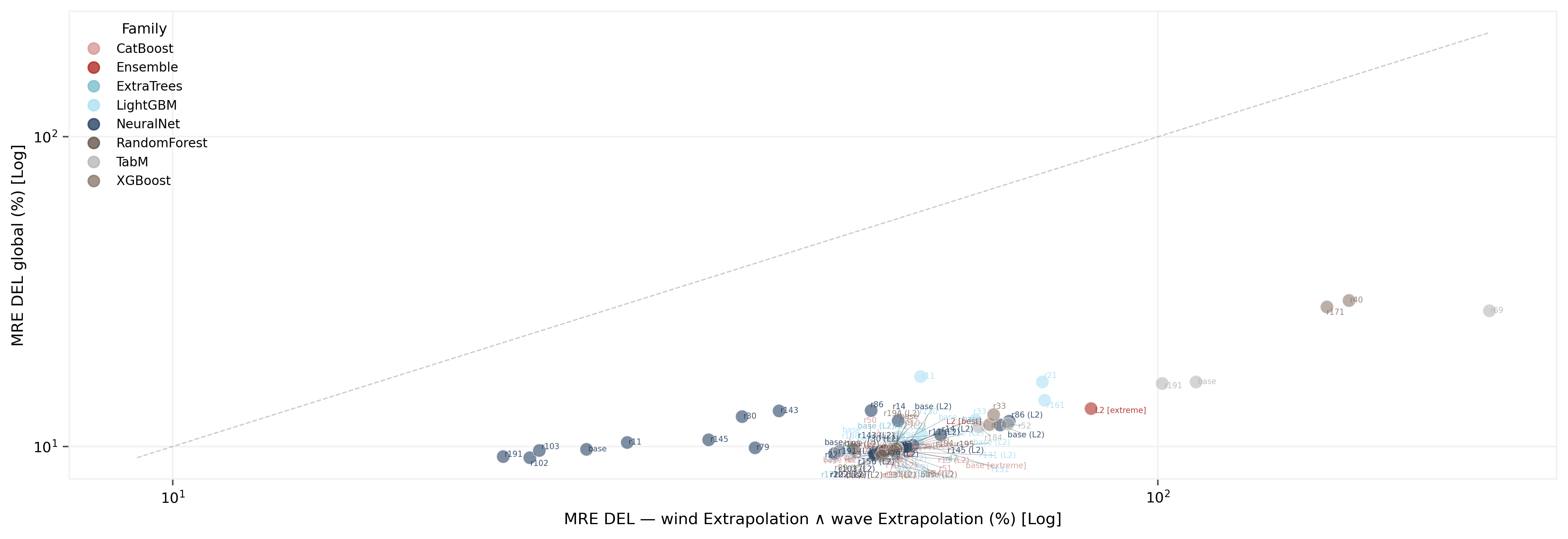}
    \caption{\textsc{opt2}.}
    \label{fig:app_scatter_global_exex_opt2}
  \end{subfigure}
  \caption{\textbf{Per-model MRE DEL: Global (y) vs.\ \textsc{EX\_EX} (x), three towers (E2).} All $\sim$95 surrogates per tower, colored by family. Points below the diagonal have higher \textsc{EX\_EX} error than Global error; models closer to the diagonal degrade less. \texttt{NeuralNet} variants stay closest to the diagonal, while \texttt{TabM} departs farthest below.}
  \label{fig:app_scatter_global_exex_3towers}
\end{figure}

\autoref{fig:app_scatter_global_exex_3towers} shows the per-model MRE DEL on Global (y-axis) vs.\ \textsc{EX\_EX} (x-axis) for the three towers. Each point is one of the $\sim$95 surrogates; color marks the family. Points below the diagonal have higher \textsc{EX\_EX} error than Global error; models closer to the diagonal degrade less. \texttt{NeuralNet} variants stay closest to the diagonal, while \texttt{TabM} departs farthest below, consistent with the family-level pattern reported in the main text.

\subsection{Wind Extrapolation dominates over wave at the family level}
\label{app:bar_family_regime}

\begin{figure}[!htbp]
  \centering
  \begin{subfigure}{\linewidth}
    \centering
    \includegraphics[width=\linewidth]{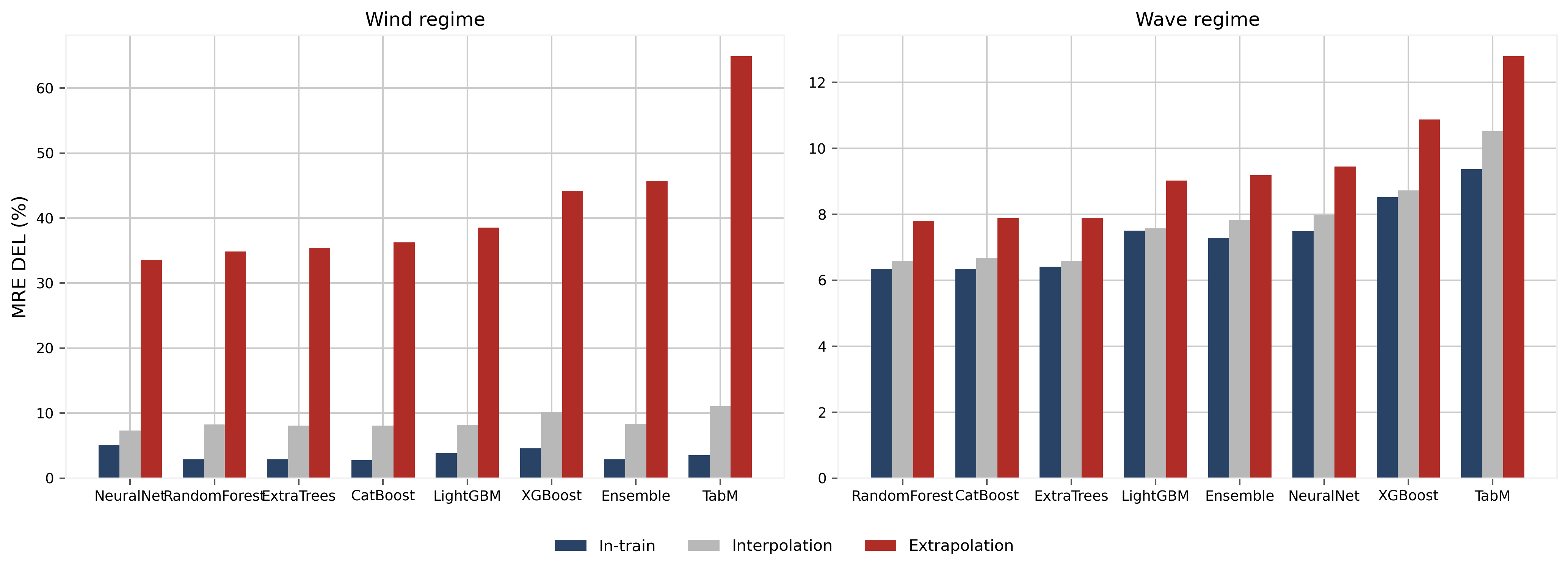}
    \caption{\textsc{ref}.}
    \label{fig:app_bar_family_regime_ref}
  \end{subfigure}\\[6pt]
  \begin{subfigure}{\linewidth}
    \centering
    \includegraphics[width=\linewidth]{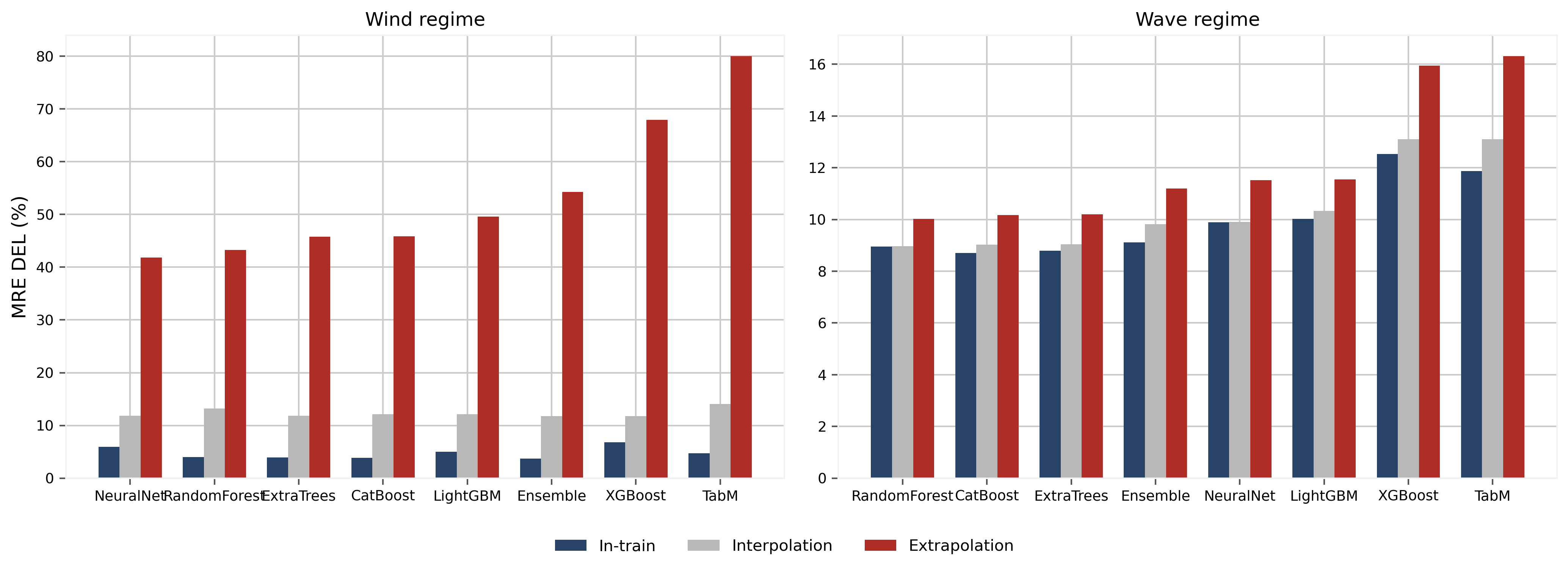}
    \caption{\textsc{opt1}.}
    \label{fig:app_bar_family_regime_opt1}
  \end{subfigure}\\[6pt]
  \begin{subfigure}{\linewidth}
    \centering
    \includegraphics[width=\linewidth]{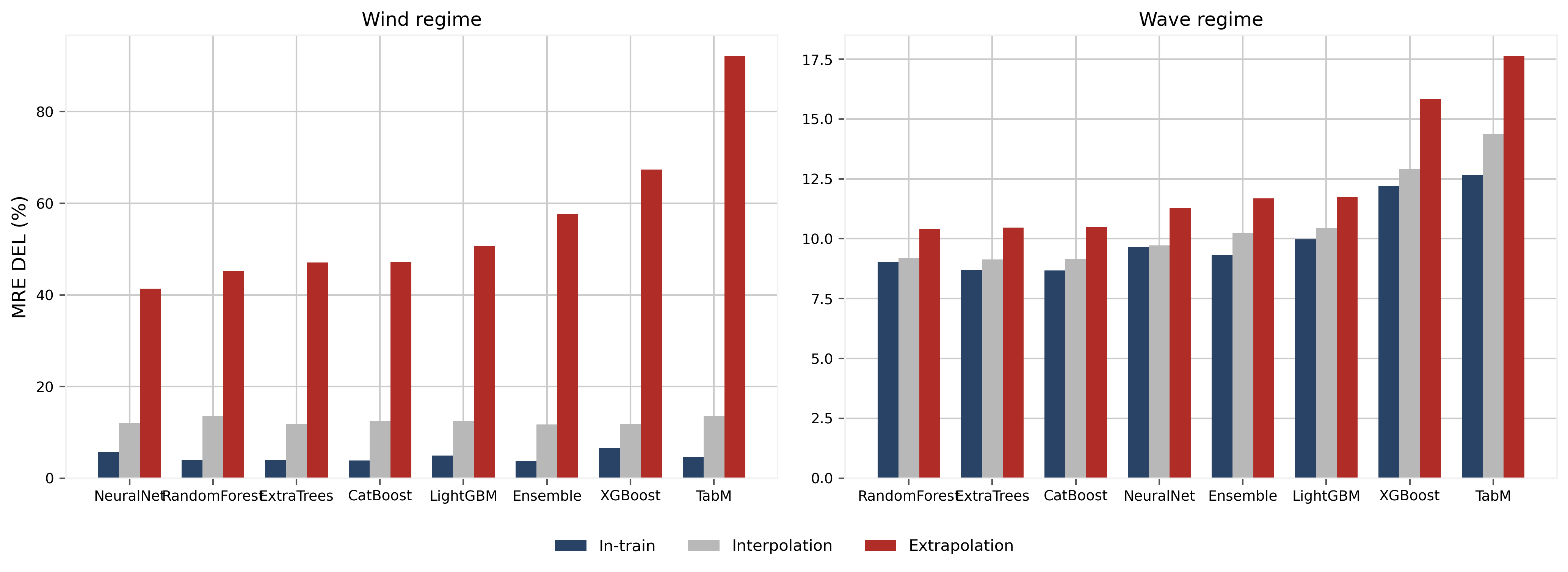}
    \caption{\textsc{opt2}.}
    \label{fig:app_bar_family_regime_opt2}
  \end{subfigure}
  \caption{\textbf{Family-aggregated MRE DEL by regime, three towers (E2).} Left sub-panel of each row: wind regime; right sub-panel: wave regime. The wind axis drives the bulk of the family-level extrapolation error; the wave axis is comparatively flat.}
  \label{fig:app_bar_family_regime_3towers}
\end{figure}

\autoref{fig:app_bar_family_regime_3towers} shows the family-aggregated MRE DEL split by wind regime (left sub-panel) and wave regime (right sub-panel) for the three towers; \texttt{NeuralNet} attains the lowest wind \textsc{EX} MRE DEL and \texttt{TabM} the highest on every tower, confirming the per-surrogate signature in \appref{app:scatter_global_exex}.

\subsection{Top-10 surrogates on Section~1 (base) and Section~30 (top) under E2, per tower}
\label{app:top10_section1}
\label{app:top10_section30}

\autoref{tab:app_top10_section1} and \autoref{tab:app_top10_section30} report the top-10 surrogates ranked by Section~1 (base) and Section~30 (top) Rel L\textsuperscript{2} DEL respectively. At the base, \texttt{NeuralNetFastAI} BAG\_L2 and \texttt{NeuralNetTorch} variants dominate on \textsc{ref}, while \texttt{WeightedEnsemble\_L2} retains rank~1 on \textsc{opt1}/\textsc{opt2}. At the top, \texttt{RandomForest} variants dominate \textsc{ref}, and \texttt{NeuralNetTorch\_r22\_BAG\_L1} consistently wins on \textsc{opt1}/\textsc{opt2}.

\begin{table}[!htbp]
  \caption{\textbf{Top-10 Section~1 (E2) per tower, sorted by Rel L\textsuperscript{2} DEL on Section~1.} Error metric on the DEL target.}
  \label{tab:app_top10_section1}
  \centering
  \scriptsize
  \setlength{\tabcolsep}{4pt}
  \begin{tabular}{c c l l r r}
    \toprule
    \makecell{Rank\\Sec.\,1} & \makecell{Rank\\Global} & Model & Preset & Rel L\textsuperscript{2} Global & Rel L\textsuperscript{2} Sec.\,1 \\
    \midrule

    \multicolumn{6}{l}{\emph{\textsc{ref}}} \\
    1 & 5 & \texttt{NeuralNetFastAI\_r11\_BAG\_L2} & best & $0.0638$ & $\mathbf{0.0601}$ \\
    2 & 2 & \texttt{NeuralNetTorch\_r22\_BAG\_L1} & best & $0.0635$ & $\mathbf{0.0601}$ \\
    3 & 4 & \texttt{NeuralNetFastAI\_r145\_BAG\_L2} & best & $0.0637$ & $\mathbf{0.0602}$ \\
    4 & 3 & \texttt{LightGBMLarge\_BAG\_L1} & best & $0.0636$ & $\mathbf{0.0603}$ \\
    5 & 16 & \texttt{CatBoost\_r13\_BAG\_L2} & best & $0.0641$ & $\mathbf{0.0603}$ \\
    6 & 11 & \texttt{CatBoost\_r177\_BAG\_L2} & best & $0.0641$ & $\mathbf{0.0604}$ \\
    7 & 1 & \texttt{WeightedEnsemble\_L2} & best & $0.0634$ & $\mathbf{0.0604}$ \\
    8 & 6 & \texttt{NeuralNetFastAI\_r191\_BAG\_L2} & best & $0.0639$ & $\mathbf{0.0605}$ \\
    9 & 20 & \texttt{RandomForest\_r195\_BAG\_L1} & best & $0.0642$ & $\mathbf{0.0605}$ \\
    10 & 27 & \texttt{NeuralNetTorch\_r79\_BAG\_L2} & best & $0.0643$ & $\mathbf{0.0606}$ \\
    \midrule
    \multicolumn{6}{l}{\emph{\textsc{opt1}}} \\
    1 & 1 & \texttt{WeightedEnsemble\_L2} & best & $0.0692$ & $\mathbf{0.0761}$ \\
    2 & 7 & \texttt{LightGBM\_r196\_BAG\_L2} & best & $0.0719$ & $\mathbf{0.0764}$ \\
    3 & 4 & \texttt{CatBoost\_r69\_BAG\_L1} & best & $0.0716$ & $\mathbf{0.0775}$ \\
    4 & 8 & \texttt{CatBoost\_r137\_BAG\_L1} & best & $0.0721$ & $\mathbf{0.0776}$ \\
    5 & 5 & \texttt{CatBoost\_BAG\_L1} & extreme & $0.0716$ & $\mathbf{0.0778}$ \\
    6 & 2 & \texttt{CatBoost\_r13\_BAG\_L1} & best & $0.0712$ & $\mathbf{0.0778}$ \\
    7 & 6 & \texttt{CatBoost\_BAG\_L1} & best & $0.0717$ & $\mathbf{0.0778}$ \\
    8 & 3 & \texttt{WeightedEnsemble\_L3} & best & $0.0712$ & $\mathbf{0.0780}$ \\
    9 & 10 & \texttt{CatBoost\_r177\_BAG\_L1} & best & $0.0724$ & $\mathbf{0.0785}$ \\
    10 & 29 & \texttt{XGBoost\_BAG\_L1} & best & $0.0744$ & $\mathbf{0.0788}$ \\
    \midrule
    \multicolumn{6}{l}{\emph{\textsc{opt2}}} \\
    1 & 1 & \texttt{WeightedEnsemble\_L2} & best & $0.0704$ & $\mathbf{0.0785}$ \\
    2 & 4 & \texttt{CatBoost\_r137\_BAG\_L1} & best & $0.0727$ & $\mathbf{0.0793}$ \\
    3 & 20 & \texttt{XGBoost\_r194\_BAG\_L1} & best & $0.0757$ & $\mathbf{0.0796}$ \\
    4 & 3 & \texttt{CatBoost\_r13\_BAG\_L1} & best & $0.0719$ & $\mathbf{0.0798}$ \\
    5 & 2 & \texttt{WeightedEnsemble\_L3} & best & $0.0719$ & $\mathbf{0.0798}$ \\
    6 & 6 & \texttt{CatBoost\_r69\_BAG\_L1} & best & $0.0730$ & $\mathbf{0.0802}$ \\
    7 & 5 & \texttt{CatBoost\_r51\_BAG\_L1} & extreme & $0.0727$ & $\mathbf{0.0803}$ \\
    8 & 7 & \texttt{CatBoost\_r177\_BAG\_L1} & best & $0.0735$ & $\mathbf{0.0811}$ \\
    9 & 30 & \texttt{XGBoost\_BAG\_L1} & best & $0.0763$ & $\mathbf{0.0811}$ \\
    10 & 9 & \texttt{CatBoost\_BAG\_L1} & best & $0.0738$ & $\mathbf{0.0812}$ \\
    \bottomrule
  \end{tabular}
\end{table}

\begin{table}[!htbp]
  \caption{\textbf{Top-10 Section~30 (E2) per tower, sorted by Rel L\textsuperscript{2} DEL on Section~30.} Error metric on the DEL target.}
  \label{tab:app_top10_section30}
  \centering
  \scriptsize
  \setlength{\tabcolsep}{4pt}
  \begin{tabular}{c c l l r r}
    \toprule
    \makecell{Rank\\Sec.\,30} & \makecell{Rank\\Global} & Model & Preset & Rel L\textsuperscript{2} Global & Rel L\textsuperscript{2} Sec.\,30 \\
    \midrule

    \multicolumn{6}{l}{\emph{\textsc{ref}}} \\
    1 & 60 & \texttt{RandomForestMSE\_BAG\_L1} & best & $0.0670$ & $\mathbf{0.0456}$ \\
    2 & 20 & \texttt{RandomForest\_r195\_BAG\_L1} & best & $0.0642$ & $\mathbf{0.0462}$ \\
    3 & 34 & \texttt{RandomForestMSE\_BAG\_L2} & best & $0.0646$ & $\mathbf{0.0466}$ \\
    4 & 26 & \texttt{RandomForest\_r195\_BAG\_L2} & best & $0.0643$ & $\mathbf{0.0468}$ \\
    5 & 19 & \texttt{LightGBM\_BAG\_L2} & best & $0.0642$ & $\mathbf{0.0471}$ \\
    6 & 18 & \texttt{LightGBM\_r131\_BAG\_L2} & best & $0.0642$ & $\mathbf{0.0473}$ \\
    7 & 7 & \texttt{LightGBM\_r130\_BAG\_L2} & best & $0.0640$ & $\mathbf{0.0474}$ \\
    8 & 11 & \texttt{CatBoost\_r177\_BAG\_L2} & best & $0.0641$ & $\mathbf{0.0476}$ \\
    9 & 15 & \texttt{LightGBMLarge\_BAG\_L2} & best & $0.0641$ & $\mathbf{0.0477}$ \\
    10 & 17 & \texttt{LightGBM\_r161\_BAG\_L2} & best & $0.0642$ & $\mathbf{0.0477}$ \\
    \midrule
    \multicolumn{6}{l}{\emph{\textsc{opt1}}} \\
    1 & 71 & \texttt{NeuralNetTorch\_r22\_BAG\_L1} & best & $0.0832$ & $\mathbf{0.0523}$ \\
    2 & 1 & \texttt{WeightedEnsemble\_L2} & best & $0.0692$ & $\mathbf{0.0531}$ \\
    3 & 3 & \texttt{WeightedEnsemble\_L3} & best & $0.0712$ & $\mathbf{0.0554}$ \\
    4 & 2 & \texttt{CatBoost\_r13\_BAG\_L1} & best & $0.0712$ & $\mathbf{0.0559}$ \\
    5 & 33 & \texttt{LightGBM\_r130\_BAG\_L2} & best & $0.0749$ & $\mathbf{0.0571}$ \\
    6 & 5 & \texttt{CatBoost\_BAG\_L1} & extreme & $0.0716$ & $\mathbf{0.0574}$ \\
    7 & 6 & \texttt{CatBoost\_BAG\_L1} & best & $0.0717$ & $\mathbf{0.0575}$ \\
    8 & 48 & \texttt{RandomForest\_r195\_BAG\_L1} & best & $0.0757$ & $\mathbf{0.0577}$ \\
    9 & 4 & \texttt{CatBoost\_r69\_BAG\_L1} & best & $0.0716$ & $\mathbf{0.0578}$ \\
    10 & 10 & \texttt{CatBoost\_r177\_BAG\_L1} & best & $0.0724$ & $\mathbf{0.0580}$ \\
    \midrule
    \multicolumn{6}{l}{\emph{\textsc{opt2}}} \\
    1 & 44 & \texttt{NeuralNetTorch\_r22\_BAG\_L1} & best & $0.0769$ & $\mathbf{0.0547}$ \\
    2 & 1 & \texttt{WeightedEnsemble\_L2} & best & $0.0704$ & $\mathbf{0.0580}$ \\
    3 & 12 & \texttt{LightGBM\_r96\_BAG\_L2} & best & $0.0750$ & $\mathbf{0.0593}$ \\
    4 & 2 & \texttt{WeightedEnsemble\_L3} & best & $0.0719$ & $\mathbf{0.0597}$ \\
    5 & 34 & \texttt{LightGBM\_BAG\_L2} & best & $0.0765$ & $\mathbf{0.0603}$ \\
    6 & 3 & \texttt{CatBoost\_r13\_BAG\_L1} & best & $0.0719$ & $\mathbf{0.0606}$ \\
    7 & 8 & \texttt{NeuralNetTorch\_BAG\_L1} & best & $0.0737$ & $\mathbf{0.0607}$ \\
    8 & 39 & \texttt{LightGBM\_r130\_BAG\_L2} & best & $0.0767$ & $\mathbf{0.0614}$ \\
    9 & 17 & \texttt{LightGBM\_r188\_BAG\_L2} & best & $0.0755$ & $\mathbf{0.0615}$ \\
    10 & 7 & \texttt{CatBoost\_r177\_BAG\_L1} & best & $0.0735$ & $\mathbf{0.0615}$ \\
    \bottomrule
  \end{tabular}
\end{table}


\clearpage
\section{E3: how cross-tower transfer behaves across folds and sections}
\label{app:cross_tower_appendix}

\subsection{Cross-tower transfer is asymmetric across folds}
\label{app:scatter_e3}

\begin{figure}[!htbp]
  \centering
  \begin{subfigure}{\linewidth}
    \centering
    \includegraphics[width=\linewidth]{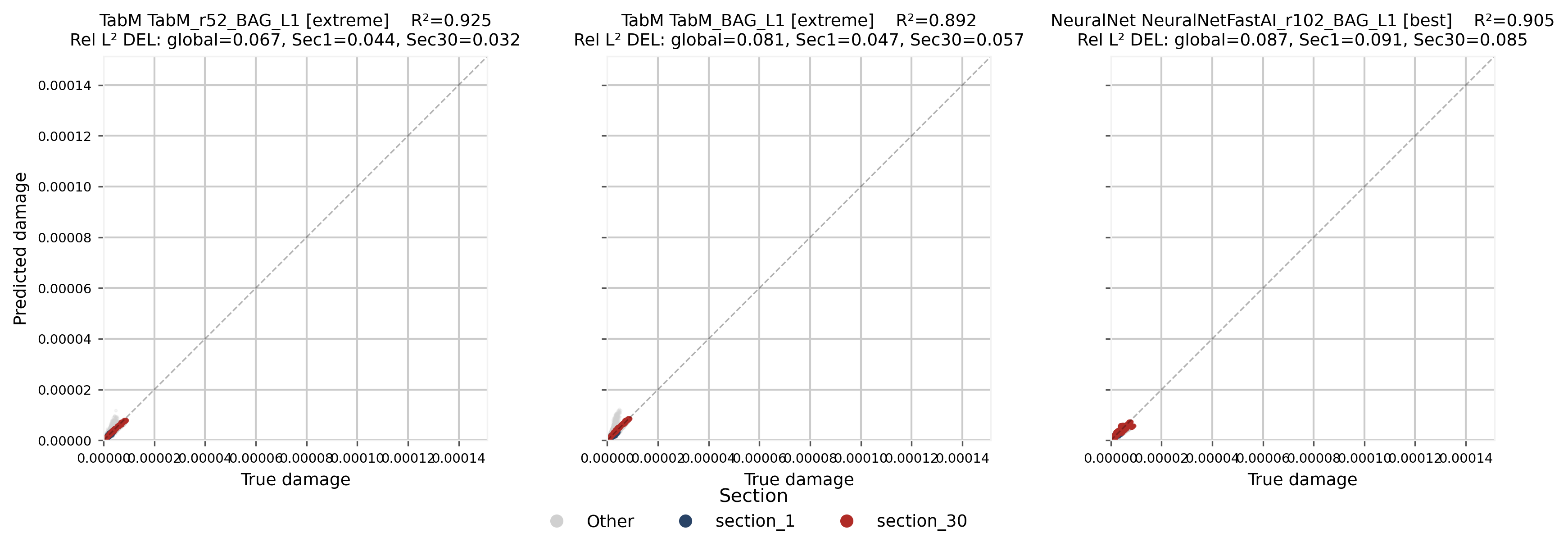}
    \caption{\textsc{ref+opt1}\,$\to$\,\textsc{opt2}.}
    \label{fig:app_scatter_e3_a}
  \end{subfigure}\\[6pt]
  \begin{subfigure}{\linewidth}
    \centering
    \includegraphics[width=\linewidth]{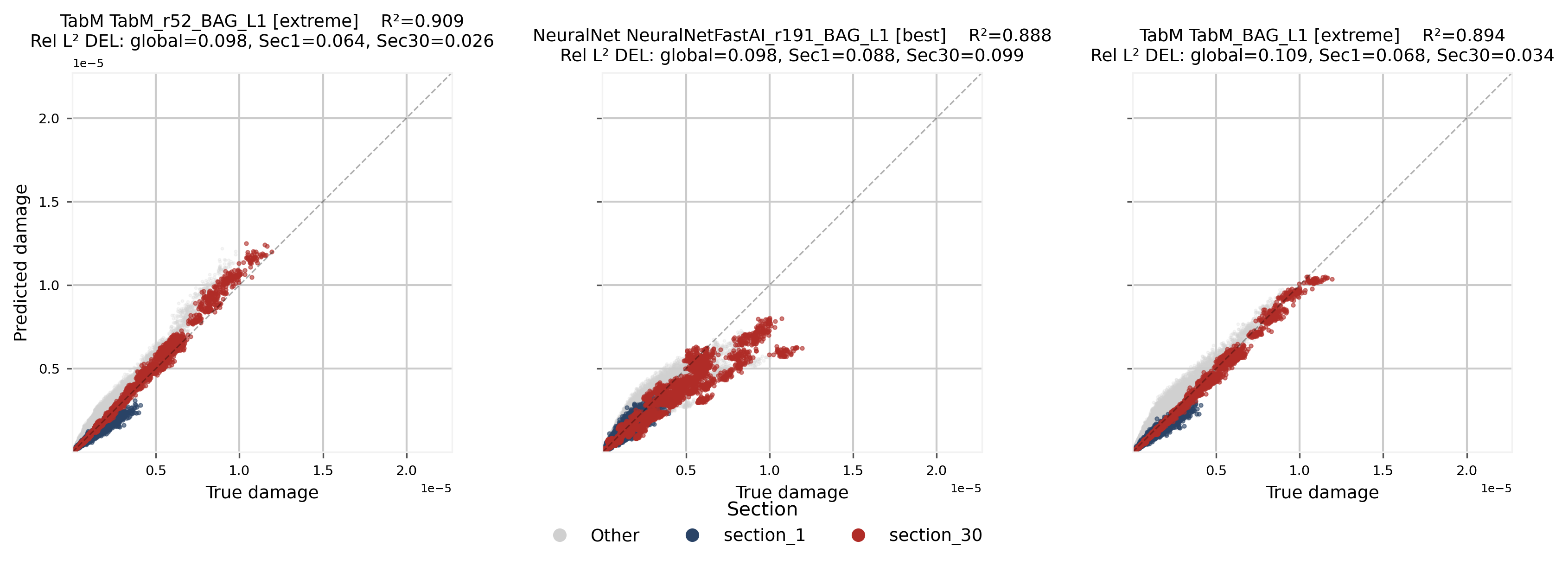}
    \caption{\textsc{ref+opt2}\,$\to$\,\textsc{opt1}.}
    \label{fig:app_scatter_e3_b}
  \end{subfigure}\\[6pt]
  \begin{subfigure}{\linewidth}
    \centering
    \includegraphics[width=\linewidth]{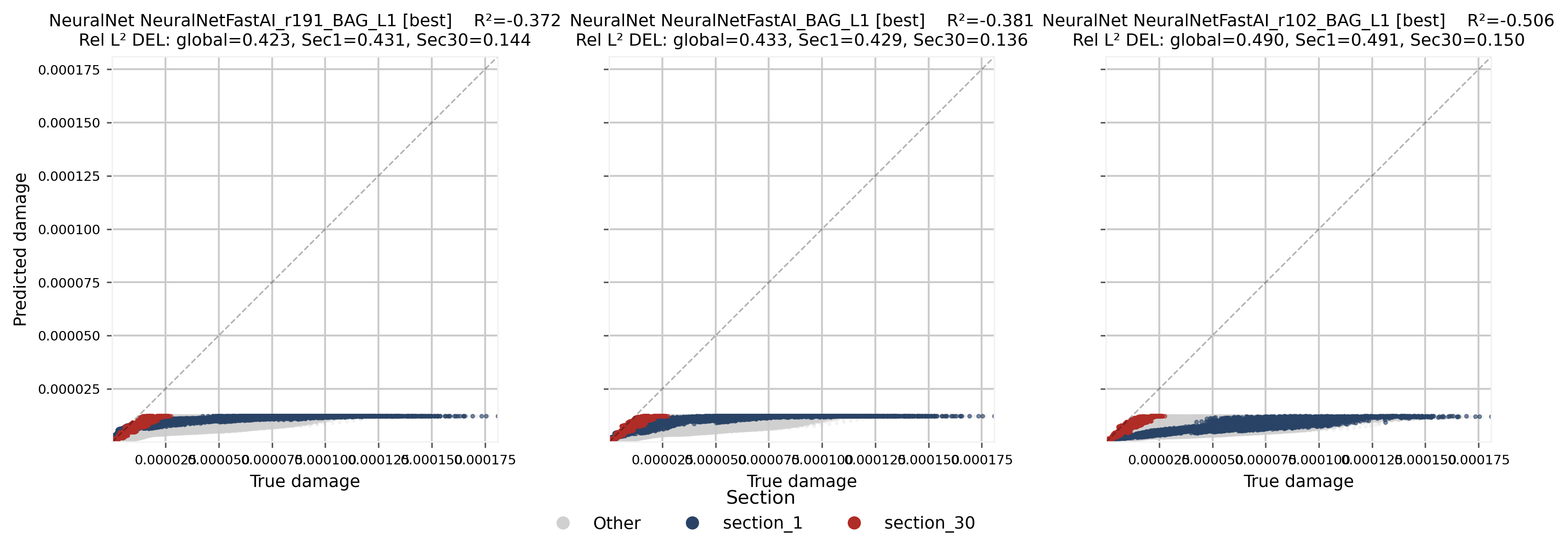}
    \caption{\textsc{opt1+opt2}\,$\to$\,\textsc{ref}.}
    \label{fig:app_scatter_e3_c}
  \end{subfigure}
  \caption{\textbf{Predicted-vs-true damage scatter for the top-3 Global surrogates per cross-tower fold (E3), sorted by Rel L\textsuperscript{2} DEL.} The bottom fold under-predicts the held-out tower; the top two folds are close to the diagonal.}
  \label{fig:app_scatter_e3}
\end{figure}

\autoref{fig:app_scatter_e3} shows the predicted-vs-true damage scatter for the top-3 Global surrogates per fold (sorted by Rel L\textsuperscript{2} DEL). The folds that include \textsc{ref} in training (top two rows) sit on the diagonal; the fold that holds \textsc{ref} out (bottom row) under-predicts, consistent with \textsc{ref}'s wider damage profile and most-distinct geometry (\autoref{fig:tower_geom_damage}).

\clearpage
\subsection{Top-10 surrogates on Section~1 (base) and Section~30 (top) under E3, per fold}
\label{app:cross_tower_sections}

\autoref{tab:app_e3_section1_top10} and \autoref{tab:app_e3_section30_top10} report the top-10 cross-tower surrogates ranked by Section~1 (base) and Section~30 (top) Rel L\textsuperscript{2} DEL respectively. \texttt{TabM\_r52\_BAG\_L1} (extreme) dominates Section~30 on all three folds; the Section~1 winners are \texttt{NeuralNet} variants on all folds: stacked BAG\_L2 variants on the easier folds (with \texttt{LightGBM} variants close behind), and a BAG\_L1 \texttt{NeuralNet} variant on \textsc{opt1+opt2}\,$\to$\,\textsc{ref}.

\begin{table}[!htbp]
  \caption{\textbf{Top-10 Section~1 (E3) per fold, sorted by Rel L\textsuperscript{2} DEL on Section~1.} Error metric on the DEL target.}
  \label{tab:app_e3_section1_top10}
  \centering
  \scriptsize
  \setlength{\tabcolsep}{4pt}
  \begin{tabular}{c c l l r r}
    \toprule
    \makecell{Rank\\Sec.\,1} & \makecell{Rank\\Global} & Model & Preset & Rel L\textsuperscript{2} Global & Rel L\textsuperscript{2} Sec.\,1 \\
    \midrule

    \multicolumn{6}{l}{\emph{\textsc{ref}\,+\,\textsc{opt1}\,$\to$\,\textsc{opt2}}} \\
    1 & 56 & \texttt{NeuralNetTorch\_r22\_BAG\_L2} & best & $0.6167$ & $\mathbf{0.0298}$ \\
    2 & 43 & \texttt{NeuralNetFastAI\_r102\_BAG\_L2} & best & $0.4541$ & $\mathbf{0.0302}$ \\
    3 & 31 & \texttt{LightGBM\_BAG\_L1} & best & $0.4318$ & $\mathbf{0.0303}$ \\
    4 & 32 & \texttt{XGBoost\_r33\_BAG\_L2} & best & $0.4357$ & $\mathbf{0.0307}$ \\
    5 & 37 & \texttt{XGBoost\_BAG\_L2} & best & $0.4422$ & $\mathbf{0.0309}$ \\
    6 & 28 & \texttt{CatBoost\_BAG\_L2} & best & $0.3976$ & $\mathbf{0.0310}$ \\
    7 & 41 & \texttt{LightGBMLarge\_BAG\_L1} & best & $0.4461$ & $\mathbf{0.0310}$ \\
    8 & 26 & \texttt{LightGBMXT\_BAG\_L2} & best & $0.3914$ & $\mathbf{0.0311}$ \\
    9 & 25 & \texttt{CatBoost\_r177\_BAG\_L2} & best & $0.3842$ & $\mathbf{0.0311}$ \\
    10 & 33 & \texttt{LightGBM\_r131\_BAG\_L2} & best & $0.4364$ & $\mathbf{0.0312}$ \\
    \midrule
    \multicolumn{6}{l}{\emph{\textsc{ref}\,+\,\textsc{opt2}\,$\to$\,\textsc{opt1}}} \\
    1 & 42 & \texttt{NeuralNetFastAI\_BAG\_L2} & best & $0.2251$ & $\mathbf{0.0305}$ \\
    2 & 47 & \texttt{LightGBMLarge\_BAG\_L1} & best & $0.2403$ & $\mathbf{0.0306}$ \\
    3 & 38 & \texttt{LightGBMLarge\_BAG\_L2} & best & $0.2132$ & $\mathbf{0.0309}$ \\
    4 & 41 & \texttt{LightGBM\_BAG\_L2} & best & $0.2219$ & $\mathbf{0.0310}$ \\
    5 & 36 & \texttt{LightGBM\_r131\_BAG\_L2} & best & $0.2070$ & $\mathbf{0.0310}$ \\
    6 & 32 & \texttt{CatBoost\_r177\_BAG\_L2} & best & $0.1967$ & $\mathbf{0.0310}$ \\
    7 & 26 & \texttt{ExtraTrees\_r42\_BAG\_L2} & best & $0.1741$ & $\mathbf{0.0311}$ \\
    8 & 31 & \texttt{CatBoost\_BAG\_L2} & best & $0.1966$ & $\mathbf{0.0311}$ \\
    9 & 27 & \texttt{ExtraTreesMSE\_BAG\_L2} & best & $0.1816$ & $\mathbf{0.0311}$ \\
    10 & 15 & \texttt{LightGBM\_BAG\_L1} & best & $0.1355$ & $\mathbf{0.0311}$ \\
    \midrule
    \multicolumn{6}{l}{\emph{\textsc{opt1}\,+\,\textsc{opt2}\,$\to$\,\textsc{ref}}} \\
    1 & 2 & \texttt{NeuralNetFastAI\_BAG\_L1} & best & $0.4332$ & $\mathbf{0.4288}$ \\
    2 & 1 & \texttt{NeuralNetFastAI\_r191\_BAG\_L1} & best & $0.4228$ & $\mathbf{0.4311}$ \\
    3 & 3 & \texttt{NeuralNetFastAI\_r102\_BAG\_L1} & best & $0.4903$ & $\mathbf{0.4915}$ \\
    4 & 4 & \texttt{NeuralNetTorch\_BAG\_L1} & best & $0.5384$ & $\mathbf{0.5631}$ \\
    5 & 6 & \texttt{NeuralNetFastAI\_r145\_BAG\_L1} & best & $0.5604$ & $\mathbf{0.5872}$ \\
    6 & 7 & \texttt{NeuralNetTorch\_r22\_BAG\_L1} & best & $0.5687$ & $\mathbf{0.5892}$ \\
    7 & 5 & \texttt{NeuralNetTorch\_r79\_BAG\_L1} & best & $0.5585$ & $\mathbf{0.5980}$ \\
    8 & 8 & \texttt{LightGBMXT\_BAG\_L1} & best & $0.5927$ & $\mathbf{0.6510}$ \\
    9 & 12 & \texttt{WeightedEnsemble\_L3} & best & $0.6026$ & $\mathbf{0.6510}$ \\
    10 & 13 & \texttt{WeightedEnsemble\_L2} & best & $0.6026$ & $\mathbf{0.6510}$ \\
    \bottomrule
  \end{tabular}
\end{table}

\begin{table}[!htbp]
  \caption{\textbf{Top-10 Section~30 (E3) per fold, sorted by Rel L\textsuperscript{2} DEL on Section~30.} Error metric on the DEL target.}
  \label{tab:app_e3_section30_top10}
  \centering
  \scriptsize
  \setlength{\tabcolsep}{4pt}
  \begin{tabular}{c c l l r r}
    \toprule
    \makecell{Rank\\Sec.\,30} & \makecell{Rank\\Global} & Model & Preset & Rel L\textsuperscript{2} Global & Rel L\textsuperscript{2} Sec.\,30 \\
    \midrule

    \multicolumn{6}{l}{\emph{\textsc{ref}\,+\,\textsc{opt1}\,$\to$\,\textsc{opt2}}} \\
    1 & 1 & \texttt{TabM\_r52\_BAG\_L1} & extreme & $0.0670$ & $\mathbf{0.0323}$ \\
    2 & 13 & \texttt{WeightedEnsemble\_L2} & extreme & $0.1142$ & $\mathbf{0.0469}$ \\
    3 & 2 & \texttt{TabM\_BAG\_L1} & extreme & $0.0808$ & $\mathbf{0.0573}$ \\
    4 & 31 & \texttt{LightGBM\_BAG\_L1} & best & $0.4318$ & $\mathbf{0.0623}$ \\
    5 & 44 & \texttt{CatBoost\_r9\_BAG\_L1} & best & $0.4729$ & $\mathbf{0.0627}$ \\
    6 & 7 & \texttt{NeuralNetTorch\_r22\_BAG\_L1} & best & $0.0989$ & $\mathbf{0.0676}$ \\
    7 & 18 & \texttt{LightGBM\_r188\_BAG\_L1} & best & $0.1757$ & $\mathbf{0.0753}$ \\
    8 & 17 & \texttt{LightGBMXT\_BAG\_L1} & best & $0.1632$ & $\mathbf{0.0780}$ \\
    9 & 3 & \texttt{NeuralNetFastAI\_r102\_BAG\_L1} & best & $0.0870$ & $\mathbf{0.0853}$ \\
    10 & 8 & \texttt{ExtraTreesMSE\_BAG\_L1} & best & $0.0993$ & $\mathbf{0.0864}$ \\
    \midrule
    \multicolumn{6}{l}{\emph{\textsc{ref}\,+\,\textsc{opt2}\,$\to$\,\textsc{opt1}}} \\
    1 & 1 & \texttt{TabM\_r52\_BAG\_L1} & extreme & $0.0975$ & $\mathbf{0.0259}$ \\
    2 & 3 & \texttt{TabM\_BAG\_L1} & extreme & $0.1094$ & $\mathbf{0.0344}$ \\
    3 & 13 & \texttt{WeightedEnsemble\_L2} & extreme & $0.1268$ & $\mathbf{0.0491}$ \\
    4 & 11 & \texttt{TabM\_r184\_BAG\_L1} & extreme & $0.1266$ & $\mathbf{0.0781}$ \\
    5 & 9 & \texttt{XGBoost\_BAG\_L1} & best & $0.1259$ & $\mathbf{0.0795}$ \\
    6 & 5 & \texttt{NeuralNetFastAI\_BAG\_L1} & best & $0.1131$ & $\mathbf{0.0890}$ \\
    7 & 4 & \texttt{NeuralNetFastAI\_r102\_BAG\_L1} & best & $0.1098$ & $\mathbf{0.0944}$ \\
    8 & 14 & \texttt{NeuralNetTorch\_r22\_BAG\_L1} & best & $0.1344$ & $\mathbf{0.0956}$ \\
    9 & 2 & \texttt{NeuralNetFastAI\_r191\_BAG\_L1} & best & $0.0985$ & $\mathbf{0.0987}$ \\
    10 & 20 & \texttt{TabM\_r191\_BAG\_L1} & extreme & $0.1504$ & $\mathbf{0.1000}$ \\
    \midrule
    \multicolumn{6}{l}{\emph{\textsc{opt1}\,+\,\textsc{opt2}\,$\to$\,\textsc{ref}}} \\
    1 & 10 & \texttt{TabM\_r52\_BAG\_L1} & extreme & $0.5992$ & $\mathbf{0.0587}$ \\
    2 & 17 & \texttt{WeightedEnsemble\_L2} & extreme & $0.6290$ & $\mathbf{0.0801}$ \\
    3 & 47 & \texttt{TabM\_r184\_BAG\_L1} & extreme & $0.6548$ & $\mathbf{0.1050}$ \\
    4 & 5 & \texttt{NeuralNetTorch\_r79\_BAG\_L1} & best & $0.5585$ & $\mathbf{0.1304}$ \\
    5 & 2 & \texttt{NeuralNetFastAI\_BAG\_L1} & best & $0.4332$ & $\mathbf{0.1365}$ \\
    6 & 1 & \texttt{NeuralNetFastAI\_r191\_BAG\_L1} & best & $0.4228$ & $\mathbf{0.1443}$ \\
    7 & 3 & \texttt{NeuralNetFastAI\_r102\_BAG\_L1} & best & $0.4903$ & $\mathbf{0.1502}$ \\
    8 & 15 & \texttt{TabM\_r191\_BAG\_L1} & extreme & $0.6155$ & $\mathbf{0.1564}$ \\
    9 & 4 & \texttt{NeuralNetTorch\_BAG\_L1} & best & $0.5384$ & $\mathbf{0.1705}$ \\
    10 & 16 & \texttt{TabM\_BAG\_L1} & extreme & $0.6210$ & $\mathbf{0.1732}$ \\
    \bottomrule
  \end{tabular}
\end{table}


\clearpage
{
\small
\label{app:references}
\bibliographystyle{unsrtnat}
\bibliography{references}

@misc{ribeiro2026float,
      title={FLOAT: Fatigue-Aware Design Optimization of Floating Offshore Wind Turbine Towers},
      author={Jo{\~a}o Alves Ribeiro and Francisco Pimenta and Bruno Alves Ribeiro and S{\'e}rgio M. O. Tavares and Faez Ahmed},
      year={2026},
      eprint={2601.01657},
      archivePrefix={arXiv},
      primaryClass={cs.CE},
      note={arXiv preprint \href{https://arxiv.org/abs/2601.01657}{arXiv:2601.01657}. Code and data: \url{https://joao97ribeiro.github.io/FLOAT/}},
}

@article{ribeiro2025owt_review,
title = {{Offshore wind turbine tower design and optimization: A review and AI-driven future directions}},
journal = {Applied Energy},
volume = {397},
pages = {126294},
year = {2025},
issn = {0306-2619},
doi = {10.1016/j.apenergy.2025.126294},
author = {João {Alves Ribeiro} and Bruno {Alves Ribeiro} and Francisco Pimenta and Sérgio {M.O. Tavares} and Jie Zhang and Faez Ahmed},
}

@techreport{iea22mw,
  author       = {Zahle, Frederik and Barlas, Thanasis and Lonbaek, Kenneth and Bortolotti, Pietro and Zalkind, Daniel and Wang, Lu and Labuschagne, Casper and Sethuraman, Latha and Barter, Garrett},
  title        = {Definition of the IEA Wind 22-Megawatt Offshore Reference Wind Turbine},
  institution  = {National Renewable Energy Laboratory \& Technical University of Denmark},
  address      = {Golden, CO, USA \& Lyngby, DNK},
  publisher    = {Technical University of Denmark},
  year         = {2024},
  month        = {04},
  note         = {\doi{10.11581/DTU.00000317}. Code and data: \url{https://github.com/IEAWindSystems/IEA-22-280-RWT}}
}

@techreport{iec61400_3_2,
  type = {Standard},
  title        = {Wind Energy Generation Systems - Part 3-2: Design Requirements for Floating Offshore Wind Turbines},
  year         = {2019},
  Author       = {{IEC 61400-3-2:2019}},
  institution  = {International Electrotechnical Commission},
  url          = {https://webstore.iec.ch/en/publication/29244},
  address = {Geneva, CH}
}

@article{mueller2018database,
title = {Validation of Uncertainty in IEC Damage Calculations Based on Measurements from Alpha Ventus},
journal = {Energy Procedia},
volume = {94},
pages = {133-145},
year = {2016},
note = {13th Deep Sea Offshore Wind R\&D Conference, EERA DeepWind'2016},
issn = {1876-6102},
doi = {10.1016/j.egypro.2016.09.208},
author = {Kolja Müller and Po Wen Cheng},
}

@article{dimitrov2018kriging,
AUTHOR = {Dimitrov, N. and Kelly, M. C. and Vignaroli, A. and Berg, J.},
TITLE = {From wind to loads: wind turbine site-specific load estimation with surrogate models trained on high-fidelity load databases},
JOURNAL = {Wind Energy Science},
VOLUME = {3},
YEAR = {2018},
NUMBER = {2},
PAGES = {767--790},
DOI = {10.5194/wes-3-767-2018}
}

@article{slot2020bayesian,
  title = {Surrogate model uncertainty in wind turbine reliability assessment},
journal = {Renewable Energy},
volume = {151},
pages = {1150-1162},
year = {2020},
issn = {0960-1481},
doi = {10.1016/j.renene.2019.11.101},
author = {René M.M. Slot and John D. Sørensen and Bruno Sudret and Lasse Svenningsen and Morten L. Thøgersen},
}

@article{singh2024fcnn,
  AUTHOR = {Haghi, R. and Crawford, C.},
TITLE = {Data-driven surrogate model for wind turbine damage equivalent load},
JOURNAL = {Wind Energy Science},
VOLUME = {9},
YEAR = {2024},
NUMBER = {11},
PAGES = {2039--2062},
DOI = {10.5194/wes-9-2039-2024}
}

@article{singh2024mdn,
  AUTHOR = {Singh, D. and Dwight, R. and Vir\'e, A.},
TITLE = {Probabilistic surrogate modeling of damage equivalent loads on onshore and offshore wind turbines using mixture density networks},
JOURNAL = {Wind Energy Science},
VOLUME = {9},
YEAR = {2024},
NUMBER = {10},
PAGES = {1885--1904},
DOI = {10.5194/wes-9-1885-2024}
}

@article{singh2025fowt,
  AUTHOR = {Singh, D. and Haugen, E. and Laugesen, K. and Dwight, R. P. and Vir\'e, A.},
TITLE = {Data-driven probabilistic surrogate model for floating wind turbine lifetime damage equivalent load prediction},
JOURNAL = {Wind Energy Science},
VOLUME = {10},
YEAR = {2025},
NUMBER = {12},
PAGES = {2865--2888},
DOI = {10.5194/wes-10-2865-2025}
}

@misc{papi2024oc4,
  author = {Papi, Francesco and Bianchini, Alessandro},
  title = {50 {Years} of {Load} {Simulation} of the {NREL} 5{MW} {OC4} {Floating} {Wind} {Turbine}},
  year = {2024},
  publisher = {Zenodo},
  note = {\doi{10.5281/zenodo.10514143}}
}

@misc{iwt75mw,
  author = {Requate, Niklas and Meyer, Tobias},
  title = {Database of {Short} {Term} {Damage} {Equivalent} {Loads} ({DEL}) of {IWT}7.5{MW} wind turbine depending on wind, {TI}, yaw and derating},
  year = {2023},
  publisher = {Zenodo},
  note = {\doi{10.5281/zenodo.8385296}}
}

@misc{lifes50_d23,
  author = {Papi, Francesco and De Luna, Robert Behrens and Saverin, Joseph and Marten, David and Compbreau, Cyril and Mirra, Gerardo and Troise, Giancarlo and Bianchini, Alessandro},
  title = {Deliverable 2.3 {Design} {Load} {Case} {Database} for {Code-to-Code} {Comparison}},
  year = {2023},
  publisher = {Zenodo},
  note = {\doi{10.5281/zenodo.8383686}}
}

@article{hu2023multisection,
title = {AK-MDAmax: Maximum fatigue damage assessment of wind turbine towers considering multi-location with an active learning approach},
journal = {Renewable Energy},
volume = {215},
pages = {118977},
year = {2023},
issn = {0960-1481},
doi = {10.1016/j.renene.2023.118977},
author = {Chao Ren and Yihan Xing},
}

@inproceedings{muller2018rbf,
    author = {Müller, Kolja and Cheng, Po Wen},
    title = {A Surrogate Modeling Approach for Fatigue Damage Assessment of Floating Wind Turbines},
    booktitle = {Proceedings of the ASME 2018 37th International Conference on Ocean, Offshore and Arctic Engineering (OMAE)},
    series = {Volume 10: Ocean Renewable Energy},
    pages = {V010T09A065},
    year = {2018},
    month = {06},
    doi = {10.1115/OMAE2018-78219},
}

@article{wei2024vinecopula,
  AUTHOR = {Yuan, Xinyu and Huang, Qian and Song, Dongran and Xia, E and Xiao, Zhao and Yang, Jian and Dong, Mi and Wei, Renyong and Evgeny, Solomin and Joo, Young-Hoon},
TITLE = {Fatigue Load Modeling of Floating Wind Turbines Based on Vine Copula Theory and Machine Learning},
JOURNAL = {Journal of Marine Science and Engineering},
VOLUME = {12},
YEAR = {2024},
NUMBER = {8},
ARTICLE-NUMBER = {1275},
ISSN = {2077-1312},
DOI = {10.3390/jmse12081275}
}

@article{zhao2024fowt,
  title = {Fatigue reliability analysis of floating offshore wind turbines under the random environmental conditions based on surrogate model},
journal = {Ocean Engineering},
volume = {314},
pages = {119686},
year = {2024},
issn = {0029-8018},
doi = {10.1016/j.oceaneng.2024.119686},
author = {Guanhua Zhao and Sheng Dong and Yuliang Zhao},
}

@article{schmidt2025kriging,
AUTHOR = {Schmidt, F. and H\"ubler, C. and Rolfes, R.},
TITLE = {Kriging meta-models for damage equivalent load assessment of idling offshore wind turbines},
JOURNAL = {Wind Energy Science},
VOLUME = {10},
YEAR = {2025},
NUMBER = {12},
PAGES = {3069--3089},
DOI = {10.5194/wes-10-3069-2025}
}

@article{avendano2021gp,
title = {Virtual fatigue diagnostics of wake-affected wind turbine via Gaussian Process Regression},
journal = {Renewable Energy},
volume = {170},
pages = {539-561},
year = {2021},
issn = {0960-1481},
doi = {10.1016/j.renene.2021.02.003},
author = {Luis David Avendaño-Valencia and Imad Abdallah and Eleni Chatzi},
}

@article{liu2024fowt,
title = {On long-term fatigue damage estimation for a floating offshore wind turbine using a surrogate model},
journal = {Renewable Energy},
volume = {225},
pages = {120238},
year = {2024},
issn = {0960-1481},
doi = {10.1016/j.renene.2024.120238},
author = {Ding Peng Liu and Giulio Ferri and Taemin Heo and Enzo Marino and Lance Manuel},
}

@inproceedings{drivAerNetPP,
    author    = {Elrefaie, Mohamed and Morar, Florin and Dai, Angela and Ahmed, Faez},
    booktitle = {Advances in Neural Information Processing Systems},
    editor    = {A. Globerson and L. Mackey and D. Belgrave and A. Fan and U. Paquet and J. Tomczak and C. Zhang},
    pages     = {499--536},
    publisher = {Curran Associates, Inc.},
    title     = {DrivAerNet++: A Large-Scale Multimodal Car Dataset with Computational Fluid Dynamics Simulations and Deep Learning Benchmarks},
    url       = {https://neurips.cc/virtual/2024/poster/97609},
    volume    = {37},
    year      = {2024}
}

@misc{carbench,
      title={CarBench: A Comprehensive Benchmark for Neural Surrogates on High-Fidelity 3D Car Aerodynamics},
      author={Mohamed Elrefaie and Dule Shu and Matt Klenk and Faez Ahmed},
      year={2025},
      eprint={2512.07847},
      archivePrefix={arXiv},
      primaryClass={cs.LG},
      note={arXiv preprint \href{https://arxiv.org/abs/2512.07847}{arXiv:2512.07847}},
}

@inproceedings{pdebench,
title={{PDEBench}: An Extensive Benchmark for Scientific Machine Learning},
author={Makoto Takamoto and Timothy Praditia and Raphael Leiteritz and Dan MacKinlay and Francesco Alesiani and Dirk Pfl{\"u}ger and Mathias Niepert},
booktitle={Thirty-sixth Conference on Neural Information Processing Systems Datasets and Benchmarks Track},
year={2022},
url={https://proceedings.neurips.cc/paper_files/paper/2022/hash/0a9747136d411fb83f0cf81820d44afb-Abstract-Datasets_and_Benchmarks.html}
}

@inproceedings{airfrans,
title={Airf{RANS}: High Fidelity Computational Fluid Dynamics Dataset for Approximating Reynolds-Averaged Navier{\textendash}Stokes Solutions},
author={Florent Bonnet and Jocelyn Ahmed Mazari and Paola Cinnella and Patrick Gallinari},
booktitle={Thirty-sixth Conference on Neural Information Processing Systems Datasets and Benchmarks Track},
year={2022},
url={https://proceedings.neurips.cc/paper_files/paper/2022/hash/94ab7b23a345f93333eac8748a66c763-Abstract-Datasets_and_Benchmarks.html}
}

@inproceedings{thewell,
title={The Well: a Large-Scale Collection of Diverse Physics Simulations for Machine Learning},
author={Ruben Ohana and Michael McCabe and Lucas Thibaut Meyer and Rudy Morel and Fruzsina Julia Agocs and Miguel Beneitez and Marsha Berger and Blakesley Burkhart and Stuart B. Dalziel and Drummond Buschman Fielding and Daniel Fortunato and Jared A. Goldberg and Keiya Hirashima and Yan-Fei Jiang and Rich Kerswell and Suryanarayana Maddu and Jonah M. Miller and Payel Mukhopadhyay and Stefan S. Nixon and Jeff Shen and Romain Watteaux and Bruno R{\'e}galdo-Saint Blancard and Fran{\c{c}}ois Rozet and Liam Holden Parker and Miles Cranmer and Shirley Ho},
booktitle={The Thirty-eight Conference on Neural Information Processing Systems Datasets and Benchmarks Track},
year={2024},
url={https://neurips.cc/virtual/2024/poster/97882}
}

@article{weatherbench2,
author = {Rasp, Stephan and Hoyer, Stephan and Merose, Alexander and Langmore, Ian and Battaglia, Peter and Russell, Tyler and Sanchez-Gonzalez, Alvaro and Yang, Vivian and Carver, Rob and Agrawal, Shreya and Chantry, Matthew and Ben Bouallegue, Zied and Dueben, Peter and Bromberg, Carla and Sisk, Jared and Barrington, Luke and Bell, Aaron and Sha, Fei},
title = {{WeatherBench} 2: A Benchmark for the Next Generation of Data-Driven Global Weather Models},
journal = {Journal of Advances in Modeling Earth Systems},
volume = {16},
number = {6},
pages = {e2023MS004019},
year = {2024},
doi = {10.1029/2023MS004019},
}

@article{oc20,
  author = {Chanussot, Lowik and Das, Abhishek and Goyal, Siddharth and Lavril, Thibaut and Shuaibi, Muhammed and Riviere, Morgane and Tran, Kevin and Heras-Domingo, Javier and Ho, Caleb and Hu, Weihua and Palizhati, Aini and Sriram, Anuroop and Wood, Brandon and Yoon, Junwoong and Parikh, Devi and Zitnick, C. Lawrence and Ulissi, Zachary},
  title = {Open {Catalyst} 2020 ({OC20}) Dataset and Community Challenges},
  journal = {ACS Catalysis},
  volume = {11},
  number = {10},
  pages = {6059--6072},
  year = {2021},
  doi = {10.1021/acscatal.0c04525}
}

@inproceedings{lips,
title={{LIPS} - Learning Industrial Physical Simulation benchmark suite},
author={Milad Leyli-abadi and Antoine Marot and J{\'e}r{\^o}me Picault and David Danan and Mouadh Yagoubi and Benjamin Donnot and Seif-Eddine Attoui and Pavel Dimitrov and Asma Farjallah and Clement Etienam},
booktitle={Thirty-sixth Conference on Neural Information Processing Systems Datasets and Benchmarks Track},
year={2022},
url={https://proceedings.neurips.cc/paper_files/paper/2022/hash/b3ac9866f6333beaa7d38926101b7e1c-Abstract-Datasets_and_Benchmarks.html}
}

@inproceedings{pinnacle,
title={{PINN}acle: A Comprehensive Benchmark of Physics-Informed Neural Networks for Solving {PDE}s},
author={Zhongkai Hao and Jiachen Yao and Chang Su and Hang Su and Ziao Wang and Fanzhi Lu and Zeyu Xia and Yichi Zhang and Songming Liu and Lu Lu and Jun Zhu},
booktitle={The Thirty-eight Conference on Neural Information Processing Systems Datasets and Benchmarks Track},
year={2024},
url={https://proceedings.neurips.cc/paper_files/paper/2024/hash/8c63299fb2820ef41cb05e2ff11836f5-Abstract-Datasets_and_Benchmarks_Track.html}
}

@inproceedings{openfwi,
title={Open{FWI}: Large-scale Multi-structural Benchmark Datasets for Full Waveform Inversion},
author={Chengyuan Deng and Shihang Feng and Hanchen Wang and Xitong Zhang and Peng Jin and Yinan Feng and Qili Zeng and Yinpeng Chen and Youzuo Lin},
booktitle={Thirty-sixth Conference on Neural Information Processing Systems Datasets and Benchmarks Track},
year={2022},
url={https://proceedings.neurips.cc/paper_files/paper/2022/hash/27d3ef263c7cb8d542c4f9815a49b69b-Abstract-Datasets_and_Benchmarks.html}
}

@article{matbench,
  author = {Dunn, Alexander and Wang, Qi and Ganose, Alex and Dopp, Daniel and Jain, Anubhav},
  title = {Benchmarking materials property prediction methods: the {MatBench} test set and {Automatminer} reference algorithm},
  journal = {npj Computational Materials},
  volume = {6},
  number = {1},
  pages = {138},
  year = {2020},
  doi = {10.1038/s41524-020-00406-3}
}

@inproceedings{gorishniy2024tabm,
title={TabM: Advancing tabular deep learning with parameter-efficient ensembling},
author={Yury Gorishniy and Akim Kotelnikov and Artem Babenko},
booktitle={The Thirteenth International Conference on Learning Representations},
year={2025},
url={https://iclr.cc/virtual/2025/poster/29590}
}

@article{hollmann2024tabpfn,
  author = {Hollmann, Noah and M{\"u}ller, Samuel and Purucker, Lennart and Krishnakumar, Arjun and K{\"o}rfer, Max and Hoo, Shi Bin and Schirrmeister, Robin Tibor and Hutter, Frank},
  title = {Accurate predictions on small data with a tabular foundation model},
  journal = {Nature},
  volume = {637},
  number = {8045},
  pages = {319--326},
  year = {2025},
  doi = {10.1038/s41586-024-08328-6}
}

@misc{qu2025tabicl,
title={Tab{ICL}: A Tabular Foundation Model for In-Context Learning on Large Data},
author={Jingang QU and David Holzm{\"u}ller and Ga{\"e}l Varoquaux and Marine Le Morvan},
booktitle={Forty-second International Conference on Machine Learning},
year={2025},
url={https://icml.cc/virtual/2025/poster/46681}
}

@misc{mehta2025mitra,
title={Mitra: Mixed Synthetic Priors for Enhancing Tabular Foundation Models},
author={Xiyuan Zhang and Danielle C. Maddix and Junming Yin and Nick Erickson and Abdul Fatir Ansari and Boran Han and Shuai Zhang and Leman Akoglu and Christos Faloutsos and Michael W. Mahoney and Cuixiong Hu and Huzefa Rangwala and George Karypis and Bernie Wang},
booktitle={The Thirty-ninth Annual Conference on Neural Information Processing Systems},
year={2025},
url={https://neurips.cc/virtual/2025/loc/san-diego/poster/115625}
}

@misc{erickson2020autogluon,
      title={AutoGluon-Tabular: Robust and Accurate AutoML for Structured Data},
      author={Nick Erickson and Jonas Mueller and Alexander Shirkov and Hang Zhang and Pedro Larroy and Mu Li and Alexander Smola},
      year={2020},
      eprint={2003.06505},
      archivePrefix={arXiv},
      primaryClass={stat.ML},
      note={arXiv preprint \href{https://arxiv.org/abs/2003.06505}{arXiv:2003.06505}},
}

@inproceedings{caruana2004ensembleselection,
author = {Caruana, Rich and Niculescu-Mizil, Alexandru and Crew, Geoff and Ksikes, Alex},
title = {Ensemble selection from libraries of models},
year = {2004},
isbn = {1581138385},
publisher = {Association for Computing Machinery},
address = {New York, NY, USA},
doi = {10.1145/1015330.1015432},
booktitle = {Proceedings of the Twenty-First International Conference on Machine Learning},
pages = {18},
location = {Banff, Alberta, Canada},
series = {ICML '04}
}

@inproceedings{chen2016xgboost,
author = {Chen, Tianqi and Guestrin, Carlos},
title = {XGBoost: A Scalable Tree Boosting System},
year = {2016},
isbn = {9781450342322},
publisher = {Association for Computing Machinery},
address = {New York, NY, USA},
doi = {10.1145/2939672.2939785},
booktitle = {Proceedings of the 22nd ACM SIGKDD International Conference on Knowledge Discovery and Data Mining},
pages = {785–794},
numpages = {10},
location = {San Francisco, California, USA},
series = {KDD '16}
}

@inproceedings{ke2017lightgbm,
  author = {Ke, Guolin and Meng, Qi and Finley, Thomas and Wang, Taifeng and Chen, Wei and Ma, Weidong and Ye, Qiwei and Liu, Tie-Yan},
  title = {{LightGBM}: A Highly Efficient Gradient Boosting Decision Tree},
  booktitle = {Advances in Neural Information Processing Systems (NeurIPS)},
  volume = {30},
  pages = {3146--3154},
  year = {2017},
   url = {https://papers.nips.cc/paper_files/paper/2017/hash/6449f44a102fde848669bdd9eb6b76fa-Abstract.html},
}

@inproceedings{prokhorenkova2018catboost,
  author = {Prokhorenkova, Liudmila and Gusev, Gleb and Vorobev, Aleksandr and Dorogush, Anna Veronika and Gulin, Andrey},
  title = {{CatBoost}: Unbiased Boosting with Categorical Features},
  booktitle = {Advances in Neural Information Processing Systems (NeurIPS)},
  volume = {31},
  pages = {6638--6648},
   url = {https://papers.nips.cc/paper_files/paper/2018/hash/14491b756b3a51daac41c24863285549-Abstract.html},

  year = {2018}
}

@article{howard2020fastai,
AUTHOR = {Howard, Jeremy and Gugger, Sylvain},
TITLE = {Fastai: A Layered API for Deep Learning},
JOURNAL = {Information},
VOLUME = {11},
YEAR = {2020},
NUMBER = {2},
ARTICLE-NUMBER = {108},
ISSN = {2078-2489},
DOI = {10.3390/info11020108}
}

@article{geurts2006extratrees,
  author = {Geurts, Pierre and Ernst, Damien and Wehenkel, Louis},
  title = {Extremely randomized trees},
  journal = {Machine Learning},
  volume = {63},
  number = {1},
  pages = {3--42},
  year = {2006},
  doi = {10.1007/s10994-006-6226-1}
}

@article{breiman2001randomforest,
  author = {Breiman, Leo},
  title = {Random Forests},
  journal = {Machine Learning},
  volume = {45},
  number = {1},
  pages = {5--32},
  year = {2001},
  doi = {10.1023/A:1010933404324}
}

@article{efron1979bootstrap,
  author = {Efron, B.},
  title = {Bootstrap Methods: Another Look at the Jackknife},
  journal = {The Annals of Statistics},
  volume = {7},
  number = {1},
  pages = {1--26},
  year = {1979},
  doi = {10.1214/aos/1176344552}
}

@article{gebru2021datasheets,
author = {Gebru, Timnit and Morgenstern, Jamie and Vecchione, Briana and Vaughan, Jennifer Wortman and Wallach, Hanna and III, Hal Daum\'{e} and Crawford, Kate},
title = {Datasheets for datasets},
year = {2021},
issue_date = {December 2021},
publisher = {Association for Computing Machinery},
address = {New York, NY, USA},
volume = {64},
number = {12},
issn = {0001-0782},
doi = {10.1145/3458723},
journal = {Commun. ACM},
month = nov,
pages = {86–92},
numpages = {7}
}

@techreport{dnv_rp_c203,
type = {Standard},
  title        = {Fatigue Design of Offshore Steel Structures},
  institution = {DNV},
  year         = {2024},
  Author       = {{DNV-RP-C203}},
  url          = {https://www.dnv.com/oilgas/download/dnv-rp-c203-fatigue-design-of-offshore-steel-structures/},
  address   = {H{\o}vik, NO}
}

@article{matsuishi1968fatigue,
    title = {Fatigue of metals subjected to varying stress},
    year = {1968},
    journal = {Proceedings of the Kyushu Branch of Japan Society of Mechanical Engineering},
    author = {Matsuishi, M and Endo, T},
    pages = {37--40},
    url = {https://www.semanticscholar.org/paper/467c88ec1feaa61400ab05fbe8b9f69046e59260}
}

@article{downing1982rainflow,
title = {Simple rainflow counting algorithms},
journal = {International Journal of Fatigue},
volume = {4},
number = {1},
pages = {31-40},
year = {1982},
issn = {0142-1123},
doi = {10.1016/0142-1123(82)90018-4},
author = {S.D. Downing and D.F. Socie}
}

@inproceedings{openfast,
author = {Jonkman, Jason},
title = {The New Modularization Framework for the {FAST} Wind Turbine {CAE} Tool},
booktitle = {51st AIAA Aerospace Sciences Meeting including the New Horizons Forum and Aerospace Exposition},
doi = {10.2514/6.2013-202},
pages = {},
year = "2013",
  note = {Software: NREL (2024), {OpenFAST}: Open-source wind turbine simulation tool, v3.5, \url{https://github.com/OpenFAST/openfast}}
}

@inproceedings{ribeiro2023simustruct,
title={{SimuStruct:} Simulated Structural Plate with Holes Dataset with Machine Learning Applications},
author={Ribeiro, Bruno Alves and Ribeiro, Jo{\~a}o Alves and Ahmed, Faez and Penedones, Hugo and Belinha, Jorge and Sarmento, Lu{\'i}s and Bessa, Miguel Anibal and Tavares, S{\'e}rgio},
booktitle={Workshop on ``Machine Learning for Materials'' ICLR 2023},
year={2023},
pages={1--10},
url={https://iclr.cc/virtual/2023/14174}
}

@article{ribeiro2021stressstrain,
author = {Ribeiro, Jo{\~a}o Alves and Tavares, S{\'e}rgio M. O. and Parente, Marco},
title = {Stress--strain evaluation of structural parts using artificial neural networks},
journal = {Proceedings of the Institution of Mechanical Engineers, Part L: Journal of Materials: Design and Applications},
volume = {235},
number = {6},
pages = {1271--1286},
year = {2021},
doi = {10.1177/1464420721992445}
}

@article{mlp_ribeiro,
author = {Ribeiro, Jo{\~a}o Alves and Gomes, Lu{\'i}s and Tavares, S{\'e}rgio M. O.},
title = {Artificial Neural Networks Applied in Mechanical Structural Design},
journal = {Journal of Computation and Artificial Intelligence in Mechanics and Biomechanics},
volume = {1},
number = {1},
pages = {14--21},
year = {2021},
month = {04},
doi = {10.5281/zenodo.4669797}
}

@article{designs8020029,
author = {Tavares, S{\'e}rgio M. O. and Ribeiro, Jo{\~a}o Alves and Ribeiro, Bruno Alves and de Castro, Paulo M. S. T.},
title = {Aircraft Structural Design and Life-Cycle Assessment through Digital Twins},
journal = {Designs},
volume = {8},
number = {2},
pages = {29},
year = {2024},
doi = {10.3390/designs8020029}
}

@article{edelsbrunner1983shape,
author = {Edelsbrunner, H. and Kirkpatrick, D. and Seidel, R.},
title = {On the shape of a set of points in the plane},
year = {1983},
publisher = {IEEE Press},
volume = {29},
number = {4},
issn = {0018-9448},
doi = {10.1109/TIT.1983.1056714},
journal = {IEEE Trans. Inf. Theor.},
month = sep,
pages = {551–559},
numpages = {9}
}
}


\end{document}